\documentclass[twoside,11pt]{article}

\usepackage{graphicx}
\usepackage{amsmath}
\usepackage{comment}
\usepackage{lmodern}
\usepackage{url}
\usepackage{appendix}
\usepackage{amsmath,setspace,paralist}
\usepackage{epstopdf}
\usepackage[normalem]{ulem}
\usepackage{multirow}
\usepackage{float}

\usepackage{bbm}
\usepackage{subcaption}
\usepackage{rotating}

\def\m{\mathcal}
\def\mb{\mathbb}
\newcommand{\+}[1]{\ensuremath{\boldsymbol {#1}}}
\newcommand{\ra}{\rightarrow}
\newcommand{\be}{\begin{equs}}
	\newcommand{\ee}{\end{equs}}
\newcommand{\bpm}{\begin{pmatrix}}
	\newcommand{\epm}{\end{pmatrix}}
\DeclareMathOperator{\E}{\mathbf E}

\newcommand{\ind}{\mathbbm 1}
\newcommand{\indm}{\mathbbm I}

\DeclareMathOperator*{\argmin}{argmin}

\usepackage{enumitem}
\newcommand \rd{\mathrm{d}}
\usepackage{accents}
\usepackage{bibunits}
\usepackage{tikz}
\usepackage{animate}
\usetikzlibrary{shapes,bayesnet}

\usepackage{jmlr2e}
\usepackage{lastpage}
\jmlrheading{25}{2024}{1-\pageref{LastPage}}{5/22; Revised
8/24}{9/24}{22-0514}{Peng Zhao, Anirban Bhattacharya, Debdeep Pati and Bani K. Mallick}
\ShortHeadings{Structured Optimal Variational Inference}{Zhao et al.}

\begin{document}



\title{Structured Optimal Variational Inference for Dynamic Latent Space Models}

\author{\name Peng Zhao \email pzhao@udel.edu \\
\addr Department of Applied Economics and Statistics \\
       University of Delaware\\
       Newark, DE 19716, USA       
\AND \name Anirban Bhattacharya \email anirbanb@stat.tamu.edu 
    \AND \name Debdeep Pati \email debdeep@stat.tamu.edu 
   \AND\name  Bani K. Mallick \email bmallick@stat.tamu.edu\\
       \addr Department of Statistics \\
       Texas A\&M University \\
       College Station, TX 77843, USA}

\editor{Ji Zhu}

\maketitle

\begin{abstract}
		We consider a latent space model for dynamic networks, where our objective is to estimate the pairwise inner products plus the intercept of the latent positions. To balance posterior inference and computational scalability, we consider a structured mean-field variational inference framework, where the time-dependent properties of the dynamic networks are exploited to facilitate computation and inference. Additionally, an easy-to-implement block coordinate ascent algorithm is developed with message-passing type updates in each block, whereas the complexity per iteration is linear with the number of nodes and time points. To certify the optimality, we demonstrate that the variational risk of the proposed variational inference approach attains the minimax optimal rate with only a logarithm factor under certain conditions. To this end,  we first derive the minimax lower bound, which might be of independent interest. In addition, we show that the posterior under commonly adopted Gaussian random walk priors can achieve the minimax lower bound with only a logarithm factor. To the best of our knowledge, this is the first such a throughout theoretical analysis of Bayesian dynamic latent space models. Simulations and real data analysis demonstrate the efficacy of our methodology and the efficiency of our algorithm. 
\end{abstract}%


	\section{Introduction}
	Statistical analysis of network-valued data is rapidly gaining popularity in modern scientific research, with applications in diverse domains such as social, biological, and computer sciences to name a few. While there is now established literature on static networks (see, e.g., the survey articles by \citealp{goldenberg2010survey}, \citealp{snijders2011statistical} and \citealp{newman2018networks}), the literature studying dynamic networks, that is, networks evolving over time, continues to show rapid growth; see \cite{xing2010state,yang2011detecting,xu2014dynamic,hoff2015multilinear,sewell2015latent,matias2017statistical,durante2017nonparametric,durante2018bayesian,pensky2019dynamic} for a flavor.

	The latent class model proposed in \cite{hoff2002latent}; see also \cite{handcock2007model,NIPS2007_766ebcd5,krivitsky2009representing,ma2020universal}; constitutes an important class of static network models and has been widely used in visualization \citep{sewell2015latent}, edge prediction \citep{durante2017bayesian} and clustering \citep{ma2020universal}. Latent space models represent each node $i$ by a latent Euclidean
	vector $\+x_i$, with the likelihood of an edge $Y_{ij}$ between nodes $i$ and $j$ entirely characterized through some distance or discrepancy $d(\+x_i,\+x_j)$ between the respective latent coordinates. Dynamic extensions of latent space models \citep{sarkar2005dynamic,sewell2015latent,friel2016interlocking,sewell2017latent,liu2022variational,loyal2023eigenmodel} are also available, which assume a Markovian evolution of the latent positions. We focus on statistical and computational aspects of variational inference in such dynamic latent space models in this article.

	To set some preliminary notation, consider a network of $n$ individuals observed over $T$ time points. For $1 \le i \ne j \le n$, let $Y_{ijt}$ denote the observed data corresponding to an edge between nodes $i$ and $j$ at time $t$. For example, $Y_{ijt} \in \{0, 1\}$ may denote the absence/presence of an edge, or $Y_{ijt} \in \mathbb{R}$ could indicate a measure of association between nodes $i$ and $j$. Let $\+Y_t = (Y_{ijt}) \in \mathbb{R}^{n \times n}$ denote the $n \times n$ network matrix at time $t$ (with only the off-diagonal part relevant), and let $\m Y = \{\+Y_t\}_{t=1}^T$ denote the observed data. We formulate our latent space model using the commonly used negative inner product $d(\+x_{it},\+x_{jt})=-\+x_{it}'\+x_{jt}$ as the discrepancy measure \citep{durante2017bayesian,ma2020universal}, where $\+x_{it} \in \mathbb{R}^d$ denotes the latent Euclidean position of node $i$ at time $t$ and $\+x'$ denotes the transpose of a vector $\+x$. 
	The observed data likelihood then takes the form
	\begin{align}\label{eq:latent_space}
	P(\m Y\mid \m X, \beta) = \prod_{t=1}^T \prod_{1\leq i \ne j \leq n} P(Y_{ijt}\mid \beta, \+x_{it},\+x_{jt}),
	\end{align}
	where $P(Y_{ijt}\mid \beta, \+x_{it},\+x_{jt})$ is decided by $\beta +\+x_{it}'\+x_{jt}$, and $\m X = \{\+X_t\}_{t=1}^T$, with $\+X_t = [\+x_{1t},...,\+x_{nt}]' \in \mathbb{R}^{n \times d}$ the matrix of the latent positions at time $t$ and $d$ is defined as the dimension of the latent space. To model the evolution of the latent positions, assume a Markov process  
	\begin{align}\label{eq:prior}
	\begin{aligned}
	\+x_{i1} &\sim \m N(\+0,\sigma_0^2\mb I_d),\quad i=1,...,n. \\
	\+x_{i(t+1)}\mid \+x_{it} &\sim \m N( \+x_{it},\tau^2\mb I_d), \quad  i=1,...,n,\,t=1,...,T-1 ,
	\end{aligned}
	\end{align}
	where $\mb I_d$ is a $d\times d$ identity matrix.

To alleviate computational inefficiencies of sampling-based posterior inference, posterior approximations based variational inference have been developed where the variational posteriors of latent positions across all times are either directly \citep{liu2022variational} or implicitly \citep{sewell2017latent}  assumed to be independent. { In the paper by  \cite{sewell2017latent}, the variational family is presented in a joint form of $q(\m X)$ in their parametrization. However, in Section 2.2 of the supplementary document, the derivation only needs a fully factorized structure of the latent positions, and their algorithm for variational inference only obtains the marginal distributions $q(\+ x_{it})$ instead of joint ones. } In dynamic models, where there is already {\em a priori} dependence between the latent states over time, assuming such an independent structure is restrictive and can lead to inconsistent estimation \citep{wang2004lack}.

	In this article, we consider a more flexible \textit{structured} mean-field (SMF) variational family, which only assumes a nodewise factorization.
	An efficient block coordinate ascent algorithm targeting the optimal SMF solution is developed, which scales linearly in the network size and retains the $O(nT)$ per-iteration computational cost of mean field (MF) by carefully constructing message-passing (MP) updates within each block to exploit the specific nature of the temporal dependence. Moreover, we empirically demonstrate that our algorithm achieves faster convergence across a wide range of simulated and real data examples. We also exhibit the mean of the optimal SMF solution to retain the same convergence rate as the exact posterior mean, providing strong support for its statistical accuracy. Overall, SMF achieves an optimal balance between the statistical accuracy of the exact posterior and the computational convenience of MF, retaining the best of both worlds.  \cite{loyal2023eigenmodel} developed a similar SMF variational approach, and a coordinate ascent variational inference (CAVI) algorithm was introduced for a latent space model aimed at dynamic multilayer networks.  Their study emphasized the algorithmic and computational perspectives, while one of our key emphasis is on deriving theoretical risk bounds for the proposed variational inference method below.

	To adaptively learn the initial and transition standard derivations, we adopt  priors
	\begin{equation}\label{eq:prior_sd}
	\sigma_0^2 \sim \mbox{Inverse-Gamma}\left(a_{\sigma_0},b_{\sigma_0}\right), \quad \tau^2 \sim \mbox{Gamma}\left(c_\tau,d_\tau\right),
	\end{equation}
	and incorporate them into our SMF framework. Although an inverse-gamma prior on the transition variance $\tau^2$ (e.g., \citealp{sewell2015latent}) leads to simple conjugate updates, it is now well-documented that an inverse-gamma prior on a lower-level variance parameter in Bayesian hierarchical models has undesirable properties when a strong shrinkage effect towards the prior mean is desired \citep{gelman2006prior,gustafson2006conservative,polson2012half}. In contrast, adopting a Gamma prior~\eqref{eq:prior_sd} on $\tau^2$ places sufficient mass near the origin, which aids our subsequent theoretical analysis and also retains closed-form updates in the form of Generalized inverse Gaussian distributions~\citep{jorgensen2012statistical}. 
	
	From a theoretical perspective, statistical analysis of variational posteriors has received major attention recently \citep{pati2018statistical,wang2019frequentist,alquier2020concentration,yang2020alpha,zhang2020convergence}. In particular, motivated by the recent development of Bayesian oracle inequalities for $\alpha$-R\'{e}nyi divergence risks \citep{bhattacharya2019bayesian}, \cite{alquier2020concentration} and \cite{yang2020alpha} proposed a theoretical framework, named $\alpha$-Variational Bayes ($\alpha$-VB), to analyze the variational risk of tempered or fractional posteriors in terms of $\alpha$-R\'{e}nyi divergences. Under the $\alpha$-VB framework, statistical optimality of variational estimators can be guaranteed by sufficient prior concentration around the true parameter and appropriate control on the Kullback–Leibler (KL) divergence between a specific variational distribution and the prior. We adopt and extend their framework to derive Bayes risk bounds under the variational posterior towards the recovery of the latent positions in an appropriate metric. A novel ingredient of our theory is the ability to provide statistical analysis for SMF variational family $q(\m X,\tau,\sigma_0)=q(\m X)q(\tau)q(\sigma_0)$ given hierarchically specified prior distributions of the form $p(\m X,\tau,\sigma_0)=p(\m X \mid \tau,\sigma_0)p(\tau)p(\sigma_0)$. 
	
The proof technique of  \cite{alquier2020concentration} and \cite{yang2020alpha}, where a specific variational candidate is constructed by truncating the prior to a small neighborhood around the true parameters, has become common for providing statistical guarantees of variational estimates. However, this technique cannot be directly applied to MF variational families endowed with a hierarchical prior specification, as the truncated distribution may not be a candidate in MF variational family due to the dependence through the global prior in the hierarchy. Previous literature \citep{liu2022variational} avoids this issue by treating the upper-level parameters of the hierarchical prior as fixed constants, thus losing the adaptivity. \cite{bai2020nearly} developed statistical guarantees for full MF variational distribution in the context of regression with global local hierarchical priors. While their algorithm used a full MF family, their theoretical results are proven assuming a dependence between the upper and lower-level parameters of the hierarchy leading to a richer family rather than a fully factorized MF. On the other hand, our theoretical results and algorithm are both developed using the same structured mean-field family where the global parameter in the hierarchy is assumed to be independent of the remaining parameters. 
	
		In addition, we exhibit the optimality of our proposed variational estimator by showing its rate of convergence to be optimal up to a logarithmic term. En route, we identify an appropriate parameter space for the latent positions and derive information-theoretic lower bounds. To the best of our knowledge, this is the first derivation of a minimax lower bound for dynamic latent space models. In fact, the only other work we are aware of that studies minimax rates for dynamic network models is \cite{pensky2019dynamic} in the context of dynamic stochastic block models.

	Finally, the computational and theoretical framework developed here can be safely adapted to the case where different nodes are equipped with different initials and transitions to capture nodewise differences:
	\begin{align}\label{eq:prior_nodewise}
	\begin{aligned}
	\+x_{i1} &\sim \m N(\+0,\sigma_{0i}^2 \mb I_d), \quad &\+x_{i(t+1)}\mid \+x_{it} \sim \m N( \+x_{it},\tau_i^2 \mb I_d),   \\
	\sigma_{0i}^2 &\sim \mbox{Inverse-Gamma}\left(a_{\sigma_0},b_{\sigma_0}\right), \quad &\tau_i^2 \sim \mbox{Gamma}\left(c_\tau,d_\tau\right),
	\end{aligned}
	\end{align}
	for $i=1,...,n\,;\, t=1,...,T-1$. 
	Due to space constraints, we present the computation and theoretical results for such nodewise adaptive priors~\eqref{eq:prior_nodewise} in Section~\ref{sec:node} of the supplementary material.

In summary, the contributions of our paper can be summarized as follows:
\begin{enumerate}
    \item  {Through the use of an SMF variational family, we proposed a CAVI algorithm, which offers an improvement over MF variational inference with minimal additional computational cost. Although our work was developed concurrently with \cite{loyal2023eigenmodel}, which also adopts an SMF variational family, our approach is distinct in that it utilizes message passing rather than a variational Kalman smoother as in \cite{loyal2023eigenmodel};}
    \item A detailed theoretical analysis of the lower bound for the squared error loss associated with sufficiently smooth latent variables is presented, which is a first practice for dynamic latent space models to our best knowledge. In addition, we develop contraction rates of the posterior and its variational approximation of the proposed SMF procedure. To our best knowledge, such an analysis is the first result in the literature on Bayesian dynamic latent space models;
    \item This technique for analyzing MF variational families for hierarchical prior distributions contributes to filling a gap in the recent literature regarding the analysis of variational inference for hierarchical prior distributions.%
  \end{enumerate}

      	\textbf{Notation.} For a vector $\+x$, we use $\|\+x\|_2$, $\|\+x\|_1$, $\|\+x\|_\infty$ to represent its $\ell_2$, $\ell_1$ and $\ell_\infty$ norms and $\+x'$ as its transpose.  For a matrix $\+A$, let $\|\+A\|_F$ be its Frobenius norm.  We use $\indm$  and $\+1$  to denote the identity matrix and vector with all ones. Suppose $P$ and $Q$ are probability measures on a common probability space with a dominating measure $\mu$,
and let $p = dP/d\mu, q = dQ/d\mu$. We use $D_{KL} \left\{ p \mid \mid q \right\} = \int p \log (p/q) d\mu $ to denote the KL divergence between the density $p$ and $q$. In addition, we use $D_{\alpha} \left\{ x\mid \mid x_0 \right\} = \log \int p_x^{\alpha} p_{x_0}^{1-\alpha} d\mu $ to denote the  R\'{e}nyi divergence of order $\alpha$ between the density $p_x$ and $p_{x_0}$.  Given sequences $a_n$ and $b_n$, we denote $a_n = O(b_n)$ or
	$a_n \lesssim b_n$ if there exists a constant $C>0$ such that $a_n \leq C b_n$ for all large enough $n$. Similarly, we
	define $a_n \gtrsim b_n$. In addition, let $a_n=o(b_n)$ to be $\lim_{n \rightarrow \infty} a_n/b_n = 0$. Let $P_X$ denote a probability distribution with parameter $X$, and $p_X$ denote the corresponding density function. Denote $\E_{x} $ as the expectation taken with respect to a variable $x$. Let $\mathcal{N}(\mu,\sigma)$ be the normal distribution with mean $\mu$ and variance $\sigma$ while $N(x;\mu,\sigma)$ be the normal density function of value $x$ with mean $\mu$ and variance $\sigma$. For any subset of $B$ of $\Theta$, we use $\Pi(B)$ to denote the probability of prior distribution taken on the set $B$.

	\section{Posterior Convergence of Dynamic Latent Space Models}\label{sec:3}
{Our main objective is to theoretically analyze the proposed SMF variational inference scheme. To this end, we first need to determine the appropriate parameter space to ensure that the variational posterior can converge at a near minimax rate. Since variational inference is an approximation of posterior inference, we will first establish that the $\alpha$-fractional posterior (a variant of posterior) can achieve the minimax optimal rate for some specific parameter space. This will provide us with the necessary tools to prove the properties of the variational posterior. One of the sufficient conditions for optimal concentration of the $\alpha$-variational posterior is that the $\alpha$-posterior itself be well behaved \citep{yang2020alpha}, which in turn is guaranteed by \cite{bhattacharya2019bayesian} through the optimal prior mass condition. In addition, the $\alpha$-fractional posterior can indicate the optimal rate at which the upper-level parameters, such as scales, need to concentrate.}
 
	In this section, we first identify a suitable parameter space \eqref{truth:PWD} for the unknown latent positions and obtain an information-theoretic lower bound to the rate of recovery (relative to a loss function defined subsequently) for said parameter space in Theorem \ref{thm:lower_minimax}. Such minimax lower-bound results for dynamic networks are scarce, and therefore this may be of independent interest. 
	Next, under mild conditions on the evolution of the latent positions, we show in Theorem \ref{cor:NWD} that the rate of contraction of the fractional posterior matches the lower bound. We expand the posterior convergence result to include a prior on the transition variance.

\subsection{Modeling Framework}\label{sec:intro}
	We first state our assumptions on the data-generating process. Assume data is generated according to \eqref{eq:latent_space} with true latent position $\m X^* = \{\+X_t^*\}_{t=1}^T$. {We assume that $\beta^*=0$ for simplicity in theoretical analysis. Therefore, the network is assumed to be dense.
As $\beta$ in Equation~\eqref{eq:latent_space} is unknown and approaches negative infinity for sparse networks, further research is needed to develop theoretical results for sparse networks in the Bayesian latent space model.} We consider Gaussian or Bernoulli distributions for the observed links, respectively, 
	\begin{equation}\label{eq:dynamic}
	\begin{aligned}
	Y_{ijt} &\stackrel{ind.}\sim \mathcal{N}(\+x_{it}^{*'}\+x_{jt}^{*}, \sigma^2), \quad 1\le i \ne j \le n, \ t=1,...,T, \quad \mbox{or} \\
	Y_{ijt} &\stackrel{ind.}\sim  \mbox{Bernoulli}\left[1/\{1+\exp(-\+x_{it}^{*'}\+x_{jt}^{*})\}\right], \quad 1\le i \ne j \le n, \ t=1,...,T,
	\end{aligned}
	\end{equation} 
	where $\+x_{it}^* \in \mathbb{R}^d$ is the $i$th row of $\+X_t^*$ and designates the true latent coordinate of individual $i$ at time $t$. The Gaussian likelihood can be considered a natural Bayesian alternative for estimating low-rank latent positions through singular value decomposition (SVD). It is worth noting that the optimization objective of SVD is associated with the best low-rank approximation in terms of the \textit{Frobenius norm}, which implies that a Gaussian-likelihood type of Bayesian alternative can be used. On the other hand, the Bernoulli likelihood, which is a natural way to model binary responses, is widely used in the network and dynamic literature; see, for instance, \cite{hoff2002latent, sewell2015latent, ma2020universal,zhang2020flexible}.

\textbf{Fractional posterior:} 
We  adopt the expanded framework of a fractional posterior \citep{walker2001bayesian}, where the usual likelihood $P(\m Y\mid \m X)$ is raised to a power $\alpha \in (0, 1)$ to form a pseudo-likelihood $P_\alpha(\m Y \mid \m X):= \left[P(\m Y\mid \m X)\right]^\alpha$, which then leads to a fractional posterior $P_\alpha(\m X, \tau,\sigma_0\mid \m Y) \propto P_\alpha(\m Y \mid \m X) \, p(\m X \mid \tau, \sigma_0)p(\tau)p(\sigma_0) $. Denote the $\epsilon$ ball for KL divergence neighborhood centered at $\m X^*$ as
	\begin{equation}\label{con:prior_mass}
         \begin{aligned}
		B_{n,T}(\m X^*;\epsilon) := \left\{\m X\in \Theta: \int p_{\m X^*}\log(\frac{p_{\m X^*}}{p_{\m X}}) d\mu \leq n(n-1)T\epsilon^2, \right. \\
  \left.\int p_{\m X^*}\log^2(\frac{p_{\m X^*}}{p_{\m X}}) d\mu \leq n(n-1)T\epsilon^2\right\},
  \end{aligned}
		\end{equation}
		where $\mu$ is the Lebesgue measure and $\Theta$ is the parameter space of $\m X$. Consider a subset $B$ of the parameter space $\Theta$, we use   $\Pi_\alpha(B \mid \m Y)= \int_B [P(\m Y \mid \m X)]^\alpha P(X ) d\m X/ \{\int_\Theta [P(\m Y \mid \m X)]^\alpha P(X ) d\m X\}$ to denote the $\alpha$-fractional posterior. Then
our technique of analyzing fractional posterior is the following Lemma, adapted from Theorem 3.1 in \cite{bhattacharya2019bayesian}:
\begin{lemma}[Contraction of fractional posterior distributions]
    Fix $\alpha \in (0,1)$. Assume $\epsilon_{n,T}$ satisfies $n^2T \epsilon_{n,T}^2 \geq 2$ and 
    $$
\Pi\left(B_{n,T}\left(\m X^*, \epsilon_{n,T} \right)\right) \geq e^{-n^2T \epsilon_{n,T}^2} .
$$
Then, for any $D \geq 2$ and $t>0$,
$$
\Pi_{\alpha}\left(\frac{1}{n(n-1)T} D_{\alpha}\left(\m X, \m X^*\right) \geq \frac{D+3 t}{1-\alpha} \epsilon_{n,T}^2 \mid \m Y\right) \leq e^{-t n^2 T \epsilon_{n,T}^2}
$$
holds with ${P}_{\m X^*}$ probability at least $1-2 /\left\{(D-1+t)^2 n^2T \epsilon_{n,T}^2\right\}$.
\end{lemma}

In contrast to the theory of original posterior distributions, which requires additional conditions (for details, refer to \citealp{ghosal2000convergence}), the prior mass condition~\eqref{con:prior_mass} is sufficient to ensure optimal concentration of fractional posterior. This is advantageous in the theoretical analysis of fractional posteriors because verifying the other conditions for complex parameter spaces can be a challenging exercise. On the other hand, the fractional power $\alpha$ only appears as a multiple factor and will not affect the main rate with respect to $n$ and $T$.  In related literature, the $\alpha$-variational posterior has been considered instead of the original one to facilitate theoretical analysis (see, for example, \citealp{linero2018bayesian,martin2020empirical,jeong2021posterior,liu2022variational}). They have also concluded that the concentration rates do not vary based on the choice of $\alpha$. When $\alpha=1/2$, the Hellinger divergence $h(p,q)$ is commonly used, and it is related to the $D_{1/2}(p,q)=-2\log(1-h^2(p,q))$, as discussed in Section 2.2 of \cite{bhattacharya2019bayesian}.

	\subsection{Lower Bounds to the Risk}\label{sec:lb_risk}
We first examine the optimal lower bound under a suitable parameter space. To capture a smooth evolution of the latent coordinates over time, we assume the following parameter space for the latent position matrices:
	\begin{equation}\label{truth:PWD}
	\mbox{PWD}(L) :=\left\{\m X:\sum_{t=2}^{T} \sum_{i=1}^{n} \|\+x_{it}-\+x_{i(t-1)}\|_2 \leq L\right\}.
	\end{equation}
	Here, PWD abbreviates point-wise dependence. The quantity $L$; which may depend on $n$ or $T$; provides an aggregate quantification of the overall `smoothness' in the evolution of the latent coordinates. 
	
	Given an estimator $\hat{\m X}$ of $\m X^*$, we consider the squared loss 
	to formulate the minimax lower bound. Observe that the latent positions are only identifiable up to rotation, and thus the loss function above is formulated in terms of the Gram matrix corresponding to the latent position matrix, which is rotation invariant. 
	
	\begin{theorem}[Minimax lower bound]\label{thm:lower_minimax}
		Suppose the data generating process follows Equation~\eqref{eq:dynamic}.
		For $\m X \in \mathrm{PWD}(L)$, with $n-d+1\geq16$, $n \geq 2d$, $T\geq 4$, and $d$ fixed, we have:
		\begin{equation*}
		\inf_{\hat{\m X}}\sup_{ \m X \in \mathrm{PWD}(L)} \E_{\m X} \left[\frac{1}{n(n-1)T} \sum_{t=1}^{T}\sum_{i \ne j=1}^n \left( \hat{\+x}'_{it}\hat{\+x}_{jt} -\+x_{it}^{'}\+x_{jt} \right)^2\right] \gtrsim \min \left\{\frac{L^{\frac{2}{3}}}{n^{\frac{4}{3}} T^{\frac{2}{3}}},\frac{1}{n}\right\}+\frac{1}{nT}.
		\end{equation*}
	\end{theorem}
	While there is a sizable literature on minimax lower bounds for various static network models \citep{abbe2015community,gao2015rate,gao2016optimal,zhang2016minimax,klopp2017oracle}, similar results for dynamic networks are scarce. To the best of our knowledge, only \cite{pensky2019dynamic} conducted such an analysis for dynamic stochastic block models, and there are no such results for latent space models. We prove the lower bound using a construction of a subset of low-rank latent states in Equation~\eqref{eq:construction} in the appendix, which is adapted from the general construction of rank-one estimation of low-rank decomposed models \citep{vu2012minimax,birnbaum2013minimax} to account for the network structure. We believe that such a construction can be used to analyze other latent space models for networks.

	Theorem \ref{thm:lower_minimax} characterizes the dependence of the lower bound on the number of time points $T$, the size of the network $n$, and the smoothness parameter $L$. We assume the latent dimension $d$ to be a fixed constant in our calculations and refrain from making the dependence of the lower bound on $d$ explicit. For fixed $n, T$, the term $L^{2/3}n^{-4/3}T^{-2/3}$ is an increasing function of $L$, implying that smoother transitions lead to better rates. However, the rate cannot be faster than $1/(nT)$ even if $L$ is arbitrarily small because under the extreme situation where all the latent positions $\+X_1,...,\+X_T$ are the same, we still need to estimate a matrix of latent positions $\+X_1$ with $O(n)$ parameters given $O(n^2T)$ observations. On the other hand, if $L$ is large enough so that $T\sqrt{n}/L=o(1)$, the lower bound is $1/n$, which is equivalent to estimating each network separately ignoring the dependence. Finally, if $n$ is fixed, the lower bound as a function of $L$ and $T$ reduces to $O(L^{2/3}T^{-2/3})$, which is the minimax rate for total variation denoising \citep{donoho1998minimax,mammen1997locally}. 
	
	\subsection{Convergence Rates of Fractional Posterior and Variational Risk}
	In this subsection, we show that under mild additional conditions, the minimax lower bound can be matched by the fractional posterior under the Gaussian random walk prior~\eqref{eq:prior}. First, we impose an identifiability condition in terms of a norm restriction:
	\begin{equation}\label{bounded}
	\|\+x_{it}^*\|_2 \leq C,\, \forall\,  i=1,...,n, t=1,...,T, \mbox{ for some constant $C>0$}.       
	\end{equation}
	Condition~\eqref{bounded} requires that all the latent positions are norm-bounded by a constant, which is mild and reasonable considering the loss is in the inner product form. Under the above condition~\eqref{bounded}, all the probabilities induced by the inner product $p_{x^*_{it},x^*_{jt}}:= 1/\{1+\exp(-\+x_{it}^{*'}\+x^*_{jt})\}$ are bounded away from $0$ and $1$ for the Bernoulli likelihood. Such an assumption is common for logistic models. 
	
	We additionally assume a homogeneity condition where we require that 
	there exists a constant $C_0>0$, such that 
	\begin{equation}\label{eq:PWD_condition}
	{\|\+x^*_{it}-\+x^*_{i(t-1)}\|_2} \leq  C_0 L/(nT), \, \forall \,  i=1,...,n, t=1,...,T.
	\end{equation}
	If the true transitions satisfy \eqref{eq:PWD_condition}, it is immediate they lie in the PWD class defined in \eqref{truth:PWD}. The homogeneity condition is compatible with random generating processes in the literature \citep{sewell2015latent}) such as a Gaussian random walk with bounded transition variance. Indeed, as long as $X^*_{ijt}-X^*_{ij(t-1)}$ for  all $i,j,t$ are sub-Gaussian random variables centered at zero and sub-Gaussian norm bounded by $\tau^{*}$,  using a concentration inequality for the maximal of sub-Gaussian random variables (Lemma~\ref{lem:sub_gaussian}), we have 
	\begin{equation}
	P\left(\max_{i,j,t} |X^*_{ijt}-X^*_{ij(t-1)}| \geq \sqrt{2\tau^{*2} \{\log (nTd) +t }\}\right) \leq 2e^{-t}.
	\end{equation}
	Therefore, with probability $1-2/(nTd)$, the homogeneity condition \eqref{eq:PWD_condition} holds when $\tau^* \leq C_0 L/(4nT \log(nTd))$. Similar conditions amounting to smooth transitions of the edge probabilities in a dynamic stochastic block model can also be found in \cite{pensky2019dynamic}. 

{ When considering binary likelihood and aiming for the convergence of the Frobenius norm of the difference between inner products, we also have a technical assumption for the prior: consider the event $ B_p = \{  \|\+x_{it}\|_2 \leq C_4 , \forall  i \ne j=1,...,n, t=1,...,T \} $ for a constant $C_4>\max \|\+x^*_{it}\|_2$. 
\begin{equation}\label{eq:prior_prob}
   \mbox{We consider the prior restricted on event }  B_p(\m X), \,\, \tilde \Pi := \Pi(\cdot \cap B_p(\m X))/\Pi(B_p(\m X)),  
\end{equation}
to replace the original prior such that $\tilde \Pi(B_p^c)=0$.
Without ambiguity, we still use $\Pi(\cdot)$ to denote $\tilde \Pi()$. {However, the assumption is used for technical simplicity in the proof of the Theorem,  ensuring the connecting probabilities are controlled at a specific rate as in the literature (e.g., \citealp{ma2020universal,zhang2022directed,zhang2022joint})}, while not used in the algorithm. In particular, adopting such a restriction will only result in a negligible difference between using the original prior. A similar phenomenon is also reported in Remark 2 in \cite{ma2020universal}.}

	Under the above conditions, we have the following theorem:
	\begin{theorem}[Fractional posterior convergence with the fixed hyperparameters]\label{cor:NWD}
		Suppose the true data generating process satisfies Equation~\eqref{eq:dynamic}, $\m X^* \in \mbox{PWD}(L)$  with $0 \leq L=o(Tn^2)$, and conditions~\eqref{bounded} and~\eqref{eq:PWD_condition} hold. Suppose $d$ is a known fixed constant. Let $\epsilon_{n,T}=L^{1/3}/(T^{1/3}n^{2/3})+\sqrt{\log(nT)/(nT)} $. Then, under the Gaussian random walk prior on $\m X$ defined in Equation~\eqref{eq:prior} and choosing  $\sigma_0$ as a fixed constant and $\tau^2 = c_1\{\epsilon_{n,T}L/(nT)+\log^2(nT)/(nT^2)\}$ for some constants $c_1>0$;	we have for $n,T \rightarrow \infty$, 
\begin{equation*}
    \Pi_{\alpha}\left(\frac{1}{n(n-1)T} D_{\alpha}\left(\m X, \m X^*\right) \geq  M\epsilon_{n,T}^2 \mid \m Y\right) \rightarrow 0.
\end{equation*}
  In addition, if condition~\eqref{eq:prior_prob} also holds, we also have
		\begin{equation}\label{result:NWD}
		\E\left[\Pi_{\alpha}\left\{ \frac{1}{n(n-1)T} \sum_{t=1}^{T}\sum_{i \ne j=1}^n \left( {\+x}'_{it}{\+x}_{jt} -\+x_{it}^{*'}\+x_{jt}^{*} \right)^2 \geq M\epsilon^2_{n,T} \mid \m Y\right\} \right] \rightarrow 0,
		\end{equation}
				with $P_{\m X^*}$ probability converging to one, where $M>0$ is a large enough constant.
	\end{theorem}
		
			Theorem~\ref{cor:NWD}  demonstrates that the minimax lower bound can be matched by the fractional posterior under specific choices of the hyperparameters $\sigma$ and $\tau$. In particular, the choice of $\tau$ ensues from an interplay between the smoothness of the Gaussian random walk prior and the truth. If $\tau$ is too small, the prior over-smoothes and fails to optimally capture the truth, while if $\tau$ is too large, then the prior under-smoothes, leading to overfitting. In particular, the smallest choice of $\tau^2$ is at the rate of $\log^2(nT)/(nT^2)$, which corresponds to the smallest error rate $\sqrt{\log(nT)/(nT)}$. 
	Moreover, Theorem~\ref{cor:NWD} implies that when the dependence is weak ($L$ is larger than $T\sqrt{n}$), applying Gaussian random walk priors with small transitions could damage the convergence rate of estimation accuracy.  Besides, the rate implies that as long as the number of networks $T$ is at least at the order of $L/\sqrt{n}$, the temporal dependence can be utilized to gain a rate no slower than the order of static network $\sqrt{1/n}$.   	The proof of Theorem~\ref{cor:NWD} is based on transforming the Gaussian random walks into initial estimations together with Brownian motions initialed at zero and traditional techniques of calculating the shifted small ball probability for Brownian motions (e.g., \citealp{van2008rates}).  
		
		\begin{theorem}[Fractional posterior convergence with hierarchical priors]\label{cor:NWD2}
		Suppose the true data generating process satisfies Equation~\eqref{eq:dynamic}, $\m X^* \in \mbox{PWD}(L)$  with $0 \leq L=o(Tn^2)$, and conditions~\eqref{bounded} and~\eqref{eq:PWD_condition} hold. Suppose $d$ is a known fixed constant. Let $\epsilon_{n,T}=L^{1/3}/(T^{1/3}n^{2/3})+\sqrt{\log(nT)/(nT)} $. Then, under the Gaussian random walk prior on $\m X$ defined in Equation~\eqref{eq:prior} and adopting priors~\eqref{eq:prior_sd} for $\sigma_0$ and $\tau$,	we have for $n,T \rightarrow \infty$,
  \begin{equation*}
    \Pi_{\alpha}\left(\frac{1}{n(n-1)T} D_{\alpha}\left(\m X, \m X^*\right) \geq  M\epsilon_{n,T}^2 \mid \m Y\right) \rightarrow 0.
\end{equation*}
  In addition, if condition~\eqref{eq:prior_prob} also holds, we also have
		\begin{equation}\label{result:NWD2}
		\E\left[\Pi_{\alpha}\left\{ \frac{1}{n(n-1)T} \sum_{t=1}^{T}\sum_{i \ne j=1}^n \left( {\+x}'_{it}{\+x}_{jt} -\+x_{it}^{*'}\+x_{jt}^{*} \right)^2 \geq M\epsilon^2_{n,T} \mid \m Y\right\} \right] \rightarrow 0,
		\end{equation}
				with $P_{\m X^*}$ probability converging to one, where $M>0$ is a large enough constant.
	\end{theorem}
		
			Theorem~\ref{cor:NWD2}, which is practically more relevant than Theorem~\ref{cor:NWD}, shows that the hierarchical prior on $\m X$ specified by $\m X \mid \sigma_0^2, \tau^2$ as in \eqref{eq:prior} and endowing the hyperparameters $\sigma^2$ and $\tau^2$ with priors as in \eqref{eq:prior_sd} leads to the same rate of contraction without knowledge of the smoothness parameter $L$. The Gamma prior on the transition variance $\tau^2$ places sufficient mass around the `optimal choice' in Theorem~\ref{cor:NWD}, which is a key ingredient in the proof of Theorem~\ref{cor:NWD2}. We comment that the current proof technique does not work with an inverse-gamma prior on $\tau^2$, with zero density at the origin.

	\section{Structured Mean-field in Latent Space Models}\label{sec:2}
 Variational approximations of fractional posteriors have also recently gained prominence \citep{bhattacharya2019bayesian,alquier2020concentration,yang2020alpha} --- from a computational point of view, minor changes are needed while Bayes risk bounds for purely fractional powers ($\alpha < 1$) require fewer conditions than the usual posterior ($\alpha=1$). Furthermore, as with the usual posterior, optimal convergence of the fractional posterior directly implies rate-optimal point estimators constructed from the fractional posterior.	Variational inference approximates the posterior distribution 
	$p(\m X, \beta, \tau,\sigma_0 \mid \m Y) \, \propto\, P(\m Y\mid \m X,\beta ) \, p(\m X \mid \tau,\sigma)p(\tau)p(\sigma_0) p(\beta)$ by its closest member in KL divergence from a pre-specified family of distributions $\Gamma$:
	\begin{align}\label{eq:KL_mini}
	\begin{aligned}
	\hat{q}(\m X, \beta,\tau,\sigma_0) &=\argmin_{q(\m X, \beta,\tau,\sigma_0) \in \Gamma} D_{KL}  \left\{ q(\m X, \beta,\tau,\sigma_0) \mid \mid p(\m X,\beta, \tau,\sigma_0\mid \m Y) \right\} \\
	&=\argmin_{q(\m X,\beta,\tau,\sigma_0) \in \Gamma} -\E_q \left\{\log \left(\frac{p(\m Y,\m X,\beta,\tau,\sigma_0)}{q(\m X,\beta,\tau,\sigma_0)}\right)\right\},
	\end{aligned} 
	\end{align}
	where the term $\E_q\{\log(p(\m X,\beta, \tau,\sigma_0\mid \m Y)/q(\m X, \beta,\tau,\sigma_0))\}$ is called evidence-lower bound (ELBO). 
	\subsection{The Structured Mean-field Family}

	For dynamic latent space models with fixed initial and transition scales $\sigma_0$ and $\tau$, the mean-field (MF) variational family \citep{liu2022variational} assumes the form
	\begin{align}\label{eq:full_mf}
	q(\m X,\beta)= \bigg[\prod_{t=1}^T\prod_{i=1}^nq(\+x_{it}) \bigg] \, q(\beta).
	\end{align} 
	The variational posterior under MF can be obtained through CAVI to maximize the ELBO (e.g., see \citealp{blei2017variational}): 
	\begin{align}\label{eq:mf}
	q^{(new)}(\beta) \propto \exp [\E_{-\beta}\{\log p(\m X,\beta,\m Y )\}]; \quad q^{(new)}(\+x_{it}) \propto \exp [\E_{-\+x_{it}}\{\log p(\m X,\beta,\m Y )\}],
	\end{align}  where $\E_{-\beta}$, $\E_{-\+x_{it}}$ are the expectations taken with respect to the densities $ q(\m X)$ and $\big[\prod_{t=1}^T\prod_{j \neq i} q(\+x_{jt})\big] q(\beta)$, respectively. On the other hand, \cite{sewell2017latent} adopted a likelihood function where the dependence across time of $\m X$ is captured by latent labels $\m Z$. According to their Equation $(22)$, they adopted a joint variational posterior for latent positions $q(\m X)$ with the assumption of independence among $\m X$  in the variational family leading to independence of $\m X$ across time $q(\m X)=\prod_{t=1}^T\prod_{i=1}^n q(\+x_{it})$. To see this more clearly, the algorithm in Section 2.2 of their supplementary material for the variational inference indicates to compute only all marginal distributions $q(\+ x_{it})$, rather than joint ones $q(\m X)$ in the variational posterior. Our proposed structured MF (SMF) variational family is instead given by
	\begin{equation}\label{eq:block_mf}
	q(\m X, \beta,\tau,\sigma_0)  = \bigg[\prod_{i=1}^{n} q_i(\+x_{i\cdot}) \bigg] \, q(\beta)q(\tau)q(\sigma_0),
	\end{equation}
	where $\+x_{i\cdot} = [\+x_{i1}',\+x_{i2}',...,\+x_{iT}']'$.
	Compared to MF, SMF does not enforce additional independence across time points $q_{it,i(t+1)}(\+x_{it},\+x_{i(t+1})=q_{it}(\+x_{it})q_{i(t+1)}(\+x_{i(t+1)})$ for $i=1,...,n$, $t=1,...,T-1$.  Figure \ref{fig:graph_representation} offers a visual comparison of the dependence structures among MF, SMF, and posterior predictives.
	
	\begin{figure}
		\begin{subfigure}{0.3\textwidth}
			\centering
			\begin{tikzpicture}[main/.style = {draw, rectangle}] 
			\node[main] (1)  {$\+x_{it}$};
			\node[main] (2) [right  =of 1] {$\+x_{jt}$};
			\node[obs] (3) [above right =of 1] {$Y_{ijt}$};
			\plate {plate1} {(1)(2)(3)} {$t=1,...,T;i \ne j=1,...,n$};
			\edge [-] {1,2}{3}
			\end{tikzpicture} 
		\end{subfigure}
		\hfill
		\begin{subfigure}{0.3\textwidth}
			\centering
			\begin{tikzpicture}[main/.style = {draw,rectangle}] 
			\node[main] (1) at (0.5,0) {$\+x_{it}$};
			\node[main] (2) [right  =of 1] {$\+x_{jt}$};
			\node[obs] (3) [above right =of 1] {$Y_{ijt}$};
			\node[main] (4) [below  =of 1] {$\+x_{i(t+1)}$};
			\node[main] (5) [right  =of 4] {$\+x_{j(t+1)}$};
			\node[obs] (6) [below right =of 4] {$Y_{ij(t+1)}$};
			\plate {plate1} {(1)(2)(3)(4)(5)(6)} {$t=1,...,T-1;i \ne j=1,...,n$};
			\edge [-] {1,2}{3}
			\edge [-] {4,5}{6}
			\edge [-] {1}{4}
			\edge [-] {2}{5}
			\end{tikzpicture} 
		\end{subfigure}
		\hfill
		\begin{subfigure}{0.3\textwidth}
			\begin{tikzpicture}[main/.style = {draw, rectangle}] 
			\node[main] (1) at (0.5,0) {$\+x_{it}$};
			\node[main] (2) [right  =of 1] {$\+x_{jt}$};
			\node[obs] (3) [above right =of 1] {$Y_{ijt}$};
			\node[main] (4) [below  =of 1] {$\+x_{i(t+1)}$};
			\node[main] (5) [right  =of 4] {$\+x_{j(t+1)}$};
			\node[obs] (6) [below right =of 4] {$Y_{ij(t+1)}$};
			\plate {plate1} {(1)(2)(3)(4)(5)(6)} {$t=1,...,T-1;i \ne j=1,...,n$};
			\edge [-] {1,2}{3}
			\edge [-] {4,5}{6}
			\edge [-] {1}{4}
			\edge [-] {2}{5}
			\edge [-] {1}{2}
			\edge [-] {4}{5}
			\end{tikzpicture} 
		\end{subfigure}
		\caption{ Graph representations for MF, SMF and exact posterior predictive distribution for latent space model given a fixed $\beta,\tau,\sigma_0$. Conditional on $\m Y$, the graph structure formed by latent positions are $nT$ isolated nodes for MF, $n$ separated chains with length $T$ for SMF and a graph with many loops for posterior.}\label{fig:graph_representation}
	\end{figure}

	\subsection{Computation for SMF}

	Utilizing the structure of the likelihood and prior, we have
	\begin{align}\label{eq:lsm}
	\begin{aligned}
	&p_\alpha(\m Y, \m X, \beta,\tau,\sigma_0) \propto P_\alpha(\m Y \mid \m X , \beta)p(\m X \mid \tau,\sigma_0)p(\tau)p(\sigma_0)p(\beta) \\
	&=  \prod_{t=1}^T \prod_{1\leq i \ne j \leq n} P_\alpha(Y_{ijt}\mid \+x_{it},\+x_{jt},\beta)  \prod_{i=1}^{n} \left\{\prod_{t=1}^{T-1}p(\+x_{i(t+1)}\mid \+x_{it} , \tau) p(\+x_{i1} \mid \sigma_0)\right\} \\
 &\times p(\beta)p(\tau)p(\sigma_0),
	\end{aligned}
	\end{align}
	with $p(\+x_{i(t+1)}\mid \+x_{it}, \tau) \propto \exp(-\|\+x_{i(t+1)}- \+x_{it}\|_2^2/(2\tau^2))$ for $t=1,...,T-1$, where $\|\+x\|_2$ represents its $\ell_2$ norm of a vector $\+x$. Based on the variational family~\eqref{eq:block_mf}, the CAVI updating of $q(\beta)$ $q(\tau)$,$q(\sigma_0)$ and $q_i(\+x_{i\cdot}),\, i=1,...,n$ are performed in an alternating fashion. The update of $\beta$ is standard and deferred to the supplemental material. We discuss the updating of the variance components in Section~\ref{sec:scale}, and at present focus on the update of $q_i$. Specifically, suppose $q(\beta)$, $q(\tau)$,$q(\sigma_0)$ and $q_j(\+x_{j\cdot}), \, j \ne i$  are fixed at their current values and the target is to update $q_i(\+x_{i\cdot})$. The CAVI scheme gives
	\begin{equation}\label{eq:variational_xit}
	\begin{aligned}
	q_i(\+x_{i\cdot}) \propto \exp \left[ \E_{-\+x_{i\cdot}}\left\{{\sum_{t=1}^{T}\sum_{j\neq i, j=1}^{n}  \{\log P_\alpha(Y_{ijt}\mid \+x_{it},\+x_{jt},\beta)\}} \right.\right. \\
	\left.\left.+ { \sum_{t=1}^{T-1}\log p(\+x_{i(t+1)}\mid \+x_{it},\tau) + \log p(\+x_{i1} \mid \sigma_0)} \right\} \right],
	\end{aligned}
	\end{equation}
	where $\E_{-\+x_{it}}$ is the expectation taken with respect to the density $\big[\prod_{j \neq i} q_j(\+x_{j\cdot})\big] q(\beta)q(\tau)q(\sigma_0)$.

	Substituting the prior and likelihood~\eqref{eq:lsm} into  Equation~\eqref{eq:variational_xit}, it follows that $q_i(\+x_{i\cdot})$ assumes the form:
	\begin{equation}\label{eq:Markov}
	q_i(\+x_{i\cdot}) = q_{i1}(\+x_{i1})\prod_{t=1}^{T-1}q(\+x_{i(t+1)} \mid q(\+x_{it})) = \prod_{t=1}^{T-1}\frac{q_{it,i(t+1)}(\+x_{it},\+x_{i(t+1)})}{q_{it}(\+x_{it})q_{i(t+1)}(\+x_{i(t+1)})}\prod_{t=1}^T q_{it}(\+x_{it}),
	\end{equation}
	which implies that the graph of random variable $\+x_{i\cdot}$ is structured by a chain from $\+x_{i1}$ to $\+x_{iT}$. It is important to notice that the structure~\eqref{eq:Markov} is not imposed by our variational family~\eqref{eq:block_mf}, rather a natural consequence of the Markov property of the prior and conditional independence of the likelihood in Equation~\eqref{eq:lsm}. Given the above structure~\eqref{eq:Markov}, computing the building blocks, i.e., the unary marginals $\{q_{it}\}$ and binary marginals $\{q_{it, i(t+1)}\}$, can be conducted in an efficient manner using message-passing \citep{pearl1982reverend}.  To that end, we first define the following quantities:
	\begin{align}\label{eq:potential}
	\begin{aligned}
	&\phi_{i1}(\+x_{i1}) = \exp\{-\mu_{1/\tau^2}\|\+x_{i1}\|_2^2/2-\mu_{1/\sigma_0^2}\|\+x_{i1}\|_2^2/2\}\prod_{j\ne i}\exp [\E_{q(\beta)q(\+x_{j1})} \{ \log P_\alpha(Y_{ij1}\mid \+x_{i1},\+x_{j1},\beta)\} ], \\
	&\phi_{it}(\+x_{it}) =\exp\{-\mu_{1/\tau^2}\|\+x_{it}\|_2^2/2\}\prod_{j\ne i}\exp [\E_{q(\beta)q(\+x_{jt})} \{ \log P_\alpha(Y_{ij1}\mid \+x_{it},\+x_{jt},\beta)\} ],   \,\forall t\in \{2,...,T\}\\
	&\psi_{it,i(t+1)}(\+x_{it},\+x_{i(t+1)}) =\exp(\mu_{1/\tau^2} \+x_{i(t+1)}' \+x_{it}), \,  \forall t\in \{1,...,T-1\},
	\end{aligned}
	\end{align}
	where $\mu_{1/\tau^2} = \E_{q(\tau)}(1/\tau^2)$ and $\mu_{1/\sigma_0^2} = \E_{q(\sigma_0)}(1/\sigma_0^2)$.
	For the ease of notation, we also denote 
	$\psi_{i0,i1}(\+x_{i0},\+x_{i1})=1$ and $\psi_{iT,i(T+1)}(\+x_{iT},\+x_{i(T+1)})=1.$

	\begin{proposition}\label{prop:message_updating}
	    The quantities appearing in the right-hand side of Equation~\eqref{eq:Markov} are given by,
		\begin{align}\label{eq:7}
		\begin{aligned}
		q_{it}(\+x_{it}) &\,\,\propto\,\,\phi_{it}(\+x_{it}) m_{i(t+1),it}(\+x_{it}) m_{i(t-1),it}(\+x_{it}), \\
		q_{it,i(t+1)}(\+x_{it},\+x_{i(t+1)}) &\propto \phi_{it}(\+x_{it}) \phi_{i(t+1)}(\+x_{i(t+1)}) m_{i(t+2),i(t+1)}(\+x_{i(t+1)})     m_{i(t-1),it}(\+x_{it}),
		\end{aligned}
		\end{align} 
		where
		\begin{align*}
		m_{i(t+1),it}(\+x_{it})  \propto \int \phi_{i(t+1)}(\+x_{i(t+1)}) \psi_{it,i(t+1)}(\+x_{it},\+x_{i(t+1)}) m_{i(t+2),i(t+1)}(\+x_{i(t+1)}) d \+x_{i(t+1)}
		\end{align*} 
		and
		\begin{align*}
		m_{it,i(t+1)}(\+x_{i(t+1)})  \propto \int \phi_{it}(\+x_{it}) \psi_{it,i(t+1)}(\+x_{it},\+x_{i(t+1)})  m_{i(t-1),it}(\+x_{it}) d \+x_{it}
		\end{align*}
		respectively are backward and forward messages for $t=1,...,T-1$. 
	\end{proposition}
 	 In the message-passing literature, messages are computational items that can be reused from different marginalization queries, which are not necessary to be distributions (see \citealp{wainwright2008graphical} for more details). Proposition~\ref{prop:message_updating} provides the order of updatings to obtain  $q_i(\+x_{i\cdot})$: first, the initial backward/forward messages satisfy $m_{i(T+1),iT}(\+x_{iT})=m_{i0,i1}(\+x_{i1})=1$. Then the other backward messages are obtained in the backward order from $m_{iT,i(T-1)}(\+x_{i(T-1)})$ to $m_{i2,i1}(\+x_{i1})$ and forward messages in the forward order from $m_{i1,i2}(\+x_{i2})$ to $m_{i(T-1),iT}(\+x_{iT})$. All messages are calculated based on the graph potentials in Equation~\eqref{eq:potential}, which can be computed analytically in conditionally conjugate Gaussian models illustrated in the next two subsections.   Then updatings of all the unary and binary marginals are performed simultaneously according to Equation~\eqref{eq:7}. Then the update of distribution $q(\+x_{i \cdot})$ can also be obtained via property~\eqref{eq:Markov} thereafter.

	The alternate MP updatings lead to an efficient block coordinate ascent algorithm where the dynamic structure of the same node is employed through MP within each block. When updating each node, the time complexity for MP is $O(T)$, hence the overall complexity per cycle is $O(nT)$. 
	For linear state-space models, the established Kalman smoothing~\citep{Kalman1960new} is often employed to obtain marginals of latent states efficiently.    Our proposed algorithm is closely connected to Kalman smoothing. Specifically, we perform MP for a chain when updating each node, which is equivalent to Kalman smoothing for state-space models only up to updating rearrangements \citep{weiss2010belief}. Similar to the variational inference literature that uses Kalman smoothing in linear state-space models to replace MP \citep{chiappa2007unified},  our proposed algorithm can also be rewritten as blockwisely implementing Kalman smoothing; see also \cite{loyal2023eigenmodel} for a parallel work in a hierarchical network model using the variational Kalman smoothing approach. We stick to the message-passing version of the proposed algorithm throughout the paper. {Note that both the SMF and MF variational inference optimization problems are non-convex and can have many local minima. In order to make sure that the algorithms converge to a good optimum, multiple random starts can be used. Furthermore, SMF has an advantage over MF in that its optimization landscape may contain fewer local minima. }

	\subsection{Gaussian Likelihood}
	
	We detail the steps of the SMF algorithm for a Gaussian likelihood:
	\begin{align*}
	P_\alpha(\m Y\mid \m X,\beta) = \prod_{t=1}^T \prod_{1\leq i \ne j \leq n}  \frac{1}{\sqrt{2 \pi} \sigma} \exp \left[-\alpha\frac{\{Y_{ijt}-\beta -\+x_{it}'\+x_{jt}\}^2}{2\sigma^2}\right].
	\end{align*}
	where $\sigma$ is assumed to be known. 
	Suppose at current step, the variational distribution for node $\+x_{it}$ follows the normal distribution $ \m N({\+\mu}_{it},{\+\Sigma}_{it})$, and the MF updating of $q(\beta)$ has been already performed so that $\E_{q(\beta)}(\beta) = \mu_{\beta}$ (provided in the supplementary material). 
	Since $ \phi_{it}(\+x_{it})$ and $\psi_{it,i(t+1)}(\+x_{it},\+x_{i(t+1)}) $ are proportional to Gaussian densities for $\+x_{it}$, the MP updating can be implemented in the framework of Gaussian belief propagation networks. Given node $i$, suppose  $\phi_{it}(\+x_{it})$ is proportional to $ N(\+x_{it}; \tilde{\+\mu}_{it},\tilde{\+\Sigma}_{it})$, which is the density function of a $\m N(\tilde{\+\mu}_{it},\tilde{\+\Sigma}_{it})$ distribution evaluated at $\+x_{it}$. Denoting $m_{it,i(t+1)}(\+x_{i(t+1)}) \propto  N(\+x_{i(t+1)} ; \+\mu_{it \rightarrow i(t+1)}, \+\Sigma_{it \rightarrow i(t+1)})$ and  $m_{it,i(t-1)}(\+x_{i(t-1)})\propto  N(\+x_{i(t+1)}; \+\mu_{it \rightarrow i(t-1)}, \+\Sigma_{it \rightarrow i(t-1)})$, and based on calculations of Gaussian conjugate and marginalization using the Schur complement, we have the forward updating steps: 
	
	$$\+\mu^{(new)}_{it \ra i(t+1)} \leftarrow -\tau^{2} \left[ \tilde{\+\Sigma}_{it}^{-1} \tilde{\+\mu}_{it}+\+\Sigma_{i(t-1) \rightarrow it}^{-1} \+\mu_{i(t-1) \rightarrow it} + \alpha\sum_{j \ne i}  (Y_{ijt}-\mu_{\beta}) \+\mu_{jt}/\sigma^2 \right]; $$
	
	$$ \+\Sigma^{(new)}_{it,i(t+1)} \leftarrow  - \tau^{4}  \left[\tilde{\+\Sigma}_{it}^{-1}+  \+\Sigma_{i(t-1) \rightarrow it}^{-1} +\alpha \sum_{j \ne i}(\+\mu_{jt}\+\mu_{jt}'+\+\Sigma_{jt})/\sigma^2\right]^{-1} . $$
	
	Similarly, for the backward updating, we have

	$$\+\mu^{(new)}_{it \ra i(t-1)} \leftarrow -\tau^{2} \left[ \tilde{\+\Sigma}_{it}^{-1} \tilde{\+\mu}_{it}+\+\Sigma_{i(t+1) \rightarrow it}^{-1} \+\mu_{i(t+1) \rightarrow it} + \alpha\sum_{j \ne i}  (Y_{ijt}-\mu_{\beta}) \+\mu_{jt}/\sigma^2 \right];$$   
	
	$$ \+\Sigma^{(new)}_{it,i(t-1)} \leftarrow  - \tau^{4}  \left[\tilde{\+\Sigma}_{it}^{-1}+  \+\Sigma_{i(t+1) \rightarrow it}^{-1} +\alpha \sum_{j \ne i}(\+\mu_{jt}\+\mu_{jt}'+\+\Sigma_{jt})/\sigma^2\right]^{-1} . $$

	\subsection{Bernoulli Likelihood}\label{sec:2.5}
	
	Next, we consider a Bernoulli likelihood
	$$P_\alpha(Y_{ijt}\mid \beta, \+x_{it},\+x_{jt})=\exp[\alpha Y_{ijt}(\beta+\+x_{it}'\+x_{jt})-\alpha\log \{1+\exp(\beta+\+x_{it}'\+x_{jt})\}],$$
	where a larger value in $-\+x_{it}'\+x_{jt}$ results in a smaller probability that nodes $i$ and $j$ are connected at time $t$. We adopt the tangent transform approach proposed by \cite{jaakkola2000bayesian} in the present context to obtain closed-form updates that are otherwise unavailable. The tangent-transform can be viewed as MF variational inference under Pólya--gamma data augmentation \citep{durante2019conditionally}. Statistical analysis of the tangent-transform for logistic regression was presented in \cite{ghosh2022statistical}.

	By introducing $\Xi=\{\xi_{ijt}: i,j=1,...,n,t=1,...,T\}$ with $A(\xi_{ijt})=-\mbox{tanh}(\xi_{ijt}/2)/(4 \xi_{ijt})$ and $C(\xi_{ijt})=\xi_{ijt}/2-\log(1+\exp(\xi_{ijt}))+\xi_{ijt} \mbox{tanh}(\xi_{ijt}/2)/(4 \xi_{ijt})$ for any $\xi_{ijt}$, we have the following lower bound on $P_\alpha (Y_{ijt}\mid \+x_{it},\+x_{jt},\beta)$:
	\begin{equation*}
	\begin{aligned}
	\underline{P}_\alpha(Y_{ijt}\mid \beta, \+x_{it},\+x_{jt} ;\xi_{ijt}) = \exp \left[\alpha A(\xi_{ijt})(\beta+\+x_{it}'\+x_{jt})^2+\alpha \left(Y_{ijt}-\frac{1}{2} \right)(\beta+\+x_{it}'\+x_{jt})+\alpha C(\xi_{ijt}) \right].
	\end{aligned}
	\end{equation*}
	By replacing $P_\alpha(Y_{ijt}\mid  \+x_{it},\+x_{jt},\beta)$ with its lower bound $\underline{P}_\alpha(Y_{ijt}\mid \+x_{it},\+x_{jt} ;\xi_{ijt},\beta)$, we can update the posterior distribution of $\m X$ in the Gaussian conjugate framework given the rest densities. 	After updating all the blocks, $\xi_{ijt}$ is optimized based on EM algorithm and the property of $A(\xi_{ijt})$ according to \cite{jaakkola2000bayesian}: $\xi^{(new)2}_{ijt} =\E_{q(\beta,\m X)}\{(\beta+\+x_{it}'\+x_{jt})^2\}$.

	In summary, for Gaussian or Bernoulli likelihood, the SMF framework allows all updatings in the Gaussian conjugate paradigm by only assuming independence between different nodes in the variational family.

	\subsection{Updatings of Scales}\label{sec:scale}
	The updating of scales can also be performed in closed form. Note that with the Gamma$(c_\tau,d_\tau)$ prior for $\tau$, by the CAVI algorithm, we have
	\begin{align}\label{eq:tau_update}
	q(\tau^2) \propto \exp\left[\E_{q(\m X)} \left\{-\sum_{t=2}^{T} \frac{\|\+X_{t}-\+X_{t-1}\|_F^2}{2\tau^2}\right\}-\frac{n(T-1)d+c_\tau-1}{2} \log(\tau^2)-d_\tau\tau^2\right].
	\end{align}
	Equation~\eqref{eq:tau_update} implies that the new update of $\tau^2$ under CAVI follows a Generalized inverse Gaussian distribution \citep{jorgensen2012statistical} with parameter $a=2d_\tau,b=\E_{q(\m X)} \{\sum_{t=2}^{T} \|\+X_{t}-\+X_{t-1}\|_F^2/2\},p=1/2-n(T-1)d/2-c_\tau/2$, where $\|\cdot\|_F$ is the Frobenius norm. Then the moment required in updating $\m X$ in Equation~\eqref{eq:potential} can be obtained:  $\E_{q(\tau)}(1/\tau^2)=\sqrt{a}K_{p+1}(\sqrt{ab})/\left\{\sqrt{b} K_p(\sqrt{ab})\right\}-2p/b$, where $K_p(\cdot)$ is the modified Bessel function of the second kind.  When $p$ is large, overflow could happen in directly calculating the value of $K_p(\cdot)$. To address this issue, expansions of $K_p(\cdot)$ can be performed in the logarithm scale, which is implemented in R package \textit{Bessel} \citep{Maechler2019Bessel}.
	
	For the initial variance $\sigma_0$ with prior~\eqref{eq:prior_sd}, the inverse-Gamma conjugate updating can be performed:
	\begin{align}\label{eq:sigma_0_update}
	q(\sigma_0^2) \propto \exp\left[\E_{q(\m X)} \left(- \frac{\|\+X_1\|_F^2}{2\sigma_0^2}\right)-\left(\frac{nd}{2}+a_{\sigma_0}+1 \right) \log(\sigma_0^2)-\frac{b_{\sigma_0}}{\sigma_0^2}\right].
	\end{align}
	Hence we have $\sigma_0^{(new)2}\sim$ Inverse-Gamma$((nd+a_{\sigma_0})/2,\{\E_{q(\m X)}(\|\+X_1\|_F^2)+2b_{\sigma_0}\}/2)$, which implies $\mu_{1/\sigma_0^2} = \E_{q(\sigma_0)}(1/\sigma_0^2)= (nd+a_{\sigma_0})/\{\E_{q(\m X)}(\|\+X_1\|_F^2)+2b_{\sigma_0}\}$.
	
	The choice of the priors~\eqref{eq:prior_sd} of the scales leads to both the closed-form updating algorithm in CAVI and the optimal convergence rate detailed in the next section. Finally, it is important to notice that the above computational framework can be safely extended to nodewise adaptive priors defined in Equation~\eqref{eq:prior_nodewise}, whose details are in Section~\ref{sec:node} in the supplementary material. 

\subsection{Theoretical Results for SMF}

To show the theoretical result of the global optimizer of the proposed SMF algorithm. First, we need the following Lemma, which is adapted from Lemma 3.3 from \cite{yang2020alpha} to prove the convergence of the $\alpha$-fractional variational posterior:
 \begin{lemma}[Variational risk bound of the $\alpha$-fractional variational posterior]\label{lem:variational}
     With $P_{\mathcal{X}^*}$ probability at least $1-\xi$ that for any probability measure $q_{\mathcal{X}} \ll p_{\mathcal{X}}$, we have 
     \begin{equation*}
         \begin{aligned}
& \int\left\{D_\alpha\left(X \| \mathcal{X}^*\right)\right\} \hat{q}_{\mathcal{X}}(\mathcal{X}) d \mathcal{X}\\
& =\frac{\alpha}{n(n-1)T(1-\alpha)}\left[-\int_{\Theta} \log \frac{p\left(Y \mid \mathcal{X}\right)}{p\left(Y \mid \mathcal{X}^*\right)} q_\mathcal{X}(\mathcal{X}) d\mathcal{X}+\frac{D\left(q_\mathcal{X}\| p_\mathcal{X}\right)}{\alpha}+\frac{\log (1 / \zeta)}{\alpha}\right],
\end{aligned}
     \end{equation*}
 \end{lemma}
 where $\hat{q}_{\mathcal{X}}(\mathcal{X})$ is the global minimizer of the KL divergence under the $\alpha$-fractional framework.

 Finally, we show in Theorems \ref{thm:varia} and~\ref{thm:varia2} below that the Bayes risk bound from Theorem \ref{cor:NWD} and \ref{cor:NWD2} is retained under the optimal SMF solution $\hat{q}$ by using Lemma~\ref{lem:variational}. As an important upshot, the point estimate obtained from the variational solution retains the same convergence rate as the fractional posterior. 
	\begin{theorem}[Variational risk bound for marginal VB families]\label{thm:varia}
		Suppose the true data generating process satisfies Equation~\eqref{eq:dynamic}, $\m X^* \in \mbox{PWD}(L)$  with $0 \leq L=o(Tn^2)$ and conditions~\eqref{bounded} and~\eqref{eq:PWD_condition} hold. Suppose $d$ is a known fixed constant. Let $\epsilon_{n,T}=L^{1/3}/(T^{1/3}n^{2/3})+\sqrt{\log(nT)/nT} $. Then if we apply the priors defined in Equation~\eqref{eq:prior},  and either the following $(a)$ or $(b)$ holds:
		\begin{itemize}
			\item [(a).] choosing  $\sigma_0$ as a fixed constant and $\tau^2 = c_1\{\epsilon_{n,T}L/(nT)+\log^2(nT)/(nT^2)\}$ for some constants $c_1>0 $ and obtaining the optimal variational distribution $\hat{q}(\m X)$ under the SMF family $q(\m X)=\prod_{i=1}^{n}q_i(\+x_{i\cdot})$;
			\item [(b).] adopting  priors~\eqref{eq:prior_sd} for $\sigma_0$ and $\tau$  and obtaining the optimal variational distribution $\hat{q}(\m X)$ under the SMF family $q(\m X)=\prod_{i=1}^{n}q_i(\+x_{i\cdot})$;
		\end{itemize}
		we have with $P_{\m X^*}$ probability tending to one as $n,T \rightarrow \infty$,
  \begin{equation*}
    \int \frac{1}{n(n-1) T} D_{\alpha}\left(\m X, \m X^*\right) \hat{q}(\m X)d\m X \lesssim \epsilon_{n,T}^2 .
\end{equation*}
  In addition, if condition~\eqref{eq:prior_prob} also holds, we also have
		\begin{equation}\label{result:variational_risk}
		\E_{\hat{q}(\m X)} \left[ \frac{1}{n(n-1)T} \sum_{t=1}^{T}\sum_{i \ne j=1}^n \left( \+x'_{it}\+x_{jt} -\+x_{it}^{*'}\+x_{jt}^{*} \right)^2\right]  \lesssim \epsilon^2_{n,T} .
		\end{equation}
	\end{theorem}
	
	Theorem~\ref{thm:varia} (a) and (b) correspond to the strategies adopted in \cite{liu2022variational} and \cite{bai2020nearly} respectively.  Theorem~\ref{thm:varia} (a) requires the tuning of hyperparameters, which loses the adaptive property of posterior under the adopted hierarchical prior as in Theorem~\ref{cor:NWD2}. In Theorem~\ref{thm:varia} (b), the variational inference is performed within a marginal family, resulting in a richer family than additional independence between $\m X$ and $\tau,\sigma_0$ as in Equation~\eqref{eq:block_mf}.  The optimization with respect to the marginal VI family, however, does not have an analytical expression and will therefore require Monte Carlo approximations (see discussions in Appendix C in \citealp{bai2020nearly}), which is not inconvenient as the algorithm proposed towards VI family~\eqref{eq:block_mf}. Nevertheless, our following Theorem~\ref{thm:varia2} shows that the gap between computation and theory can be bridged.

		\begin{theorem}[Variational risk bound for SMF]\label{thm:varia2}
		Suppose the true data generating process satisfies Equation~\eqref{eq:dynamic}, $\m X^* \in \mbox{PWD}(L)$  with $0 \leq L=o(Tn^2)$ and conditions~\eqref{bounded} and~\eqref{eq:PWD_condition} hold. Suppose $d$ is a known fixed constant. Let $\epsilon_{n,T}=L^{1/3}/(T^{1/3}n^{2/3})+\sqrt{\log(nT)/nT} $. Then if we apply the priors defined in Equation~\eqref{eq:prior},  and adopt priors~\eqref{eq:prior_sd} for $\sigma_0$ and $\tau$  and obtaining the optimal variational distribution $\hat{q}(\m X)$ under SMF  family~\eqref{eq:block_mf},
		we have with $P_{\m X^*}$ probability tending to one as $n,T \rightarrow \infty$,
   \begin{equation*}
    \int \frac{1}{n(n-1)T} D_{\alpha}\left(\m X, \m X^*\right) \hat{q}(\m X)d\m X \lesssim \epsilon_{n,T}^2 .
\end{equation*}
  In addition, if condition~\eqref{eq:prior_prob} also holds, we also have
		\begin{equation}\label{result:variational_risk2}
		\E_{\hat{q}(\m X)} \left[ \frac{1}{n(n-1)T} \sum_{t=1}^{T}\sum_{i \ne j=1}^n \left( \+x'_{it}\+x_{jt} -\+x_{it}^{*'}\+x_{jt}^{*} \right)^2\right]  \lesssim \epsilon^2_{n,T} .
		\end{equation}
	\end{theorem}

	Theorem~\ref{thm:varia2} is the main theorem in this paper that corresponds to the proposed algorithm. It implies that the independence in the SMF family between $\m X$ and $\tau,\sigma_0$ does not bring any damage to the convergence rate of the optimal variational estimator. Therefore, the proposed VI algorithm enjoys the same adaptive property as the posterior under the adopted hierarchical prior without any loss.  The proof strategy of Theorem \ref{thm:varia2} has a key distinction with Theorem~\ref{thm:varia}: the mismatch between the hierarchical prior $p(\m X,\tau,\sigma_0)=p(\m X \mid \tau,\sigma_0)p(\tau)p(\sigma_0)$ and independent variational family $q(\m X,\tau,\sigma_0)=q(\m X) q(\tau) q(\sigma_0)$ adds some complexity in the analysis. We construct a candidate in the variational family, which leads to the optimal rate. Specifically, for any chosen $\tau^*,\sigma_0^*$ that satisfy the condition in Theorem~\ref{thm:varia} (a),  construct $\bar{q}(\m X,\tau,\sigma_0)$ by restricting  $p(\+x_{i(t+1)} \mid \+x_{it} ,\tau^*)$ to a neighborhood of $\+x_{i(t+1)}^*$ for all $i$ and $t>1$ and restricting $p(x_{i1} \mid \sigma_0^*)$ to a neighborhood of $\+x_{i1}^*$ for all $i$. Also restrict $p(\tau)$, $p(\sigma_0)$ to a neighborhood of $\tau^*,\sigma_0^*$ (see Equation~\eqref{eq:variational_candidate} in the appendix). Observe that such a construction lies within the proposed SMF family~\eqref{eq:block_mf}. By an appropriate choice of the size of the neighborhood, we can achieve
	\begin{equation*}
	D_{KL}(\bar q(\m X,\tau,\sigma_0) \,||\, p(\m X,\tau,\sigma_0)) \lesssim Tn(n-1)\epsilon_{n,T}^2.
	\end{equation*}
	In addition, by using the decomposition
	\begin{align*}
	\begin{aligned}
	D_{KL}(\bar q(\m X,\tau,\sigma_0) \,||\, p(\m X,\tau,\sigma_0)) = D_{KL}(\bar q(\tau) || p(\tau)) + D_{KL}(\bar q(\sigma_0) || p(\sigma_0))\\
	+ \int \bar  q(\tau)q(\bar \sigma_0) \int \bar q( {\m X}) \log \frac{\bar  q( {\m X})}{p(\m X |\tau, \sigma_0)} d\m X d\tau d\sigma_0,
	\end{aligned}
	\end{align*}
	we are able to show that the selected candidate achieves optimal bounds for all three terms on the right-hand side.  We conjecture that this strategy is fairly general and can be applied to mean-field inference for other Bayesian hierarchical models as well.

	\section{Simulations and Real Data Analysis}\label{sec:4}
In this section, we will first provide simulation examples to illustrate our results. Then, we will present two real data examples: the Enron Email data set and the McFarland Classroom data set.
	\subsection{Simulation Experiments}
	We perform replicated simulation studies to compare SMF, MF and MCMC. Throughout all simulation and real data analyses, we fix the fractional power $\alpha=0.95$. We also fix the hyperparameters $a_{\sigma_0}=1/2,b_{\sigma_0}=1/2$ and $c_\tau=1, d_\tau=1/2$ whenever the prior~\eqref{eq:prior_sd} is used. Simulation results for Gaussian likelihood can be found in Section~\ref{sec:additional_simu} in the appendix.
 
	\textbf{Binary Networks:}
	$25$ replicated data sets are generated from \eqref{eq:latent_space} with $Y_{ijt}$ $\sim$ $\mbox{Bernoulli}$ $[1/\{1+\exp(-2+\+x_{it}'\+x_{jt})\}]$ for $i\neq j =1,...,n$ and $t=1,...,T$ with $d=2$.  The latent positions are initialized as $\+x_{i1} \sim 0.5 \mathcal{N}((1,0)',0.5^2 \indm) +0.5 \mathcal{N}((-1,0)',0.5^2 \indm)$ with subsequent draws from $\+x_{it} = \+x_{i(t-1)} + \+\epsilon_{t-1}$, where given any coordinate $j$ for a fixed node $i$, we have $[\epsilon_{ij1},...,\epsilon_{ijT}]' \sim \mathcal{N}(\+0,\tau^2 ((1-\rho) \indm +\rho \+1\+1'))$. The transition sd $\tau$ controls the magnitude of transition, and the auto-correlation $\rho$  controls the positive dependence. As a measure of discrepancy between the true and estimated probabilities, we use the sample Pearson correlation coefficient (PCC, which is also used in other literature, e.g, \citealp{sewell2017latent}): $\sum_{i=1}^n(x_i-\bar x)(y_i -\bar y)/\sqrt{\sum_{i=1}^n(x_i-\bar x)^2}/\sqrt{\sum_{i=1}^n(y_i-\bar y)}$ for two lists of probabilities $(x_1,...,x_n)$ and $(y_1,...,y_n)$.  The number of iterations until convergence is reported to investigate the computational efficiency.  The stopping criterion is taken to be the difference between training AUCs (area under the curve) in two consecutive cycles not exceeding $0.01$. To implement SMF and MF,  we assume the initial variance to be $\sigma_0=0.5$. The prior for parameter $\beta$ is set to {$\mathcal{N}(0,10)$}.

\begin{figure}
\centering
\includegraphics[width=0.8\linewidth]{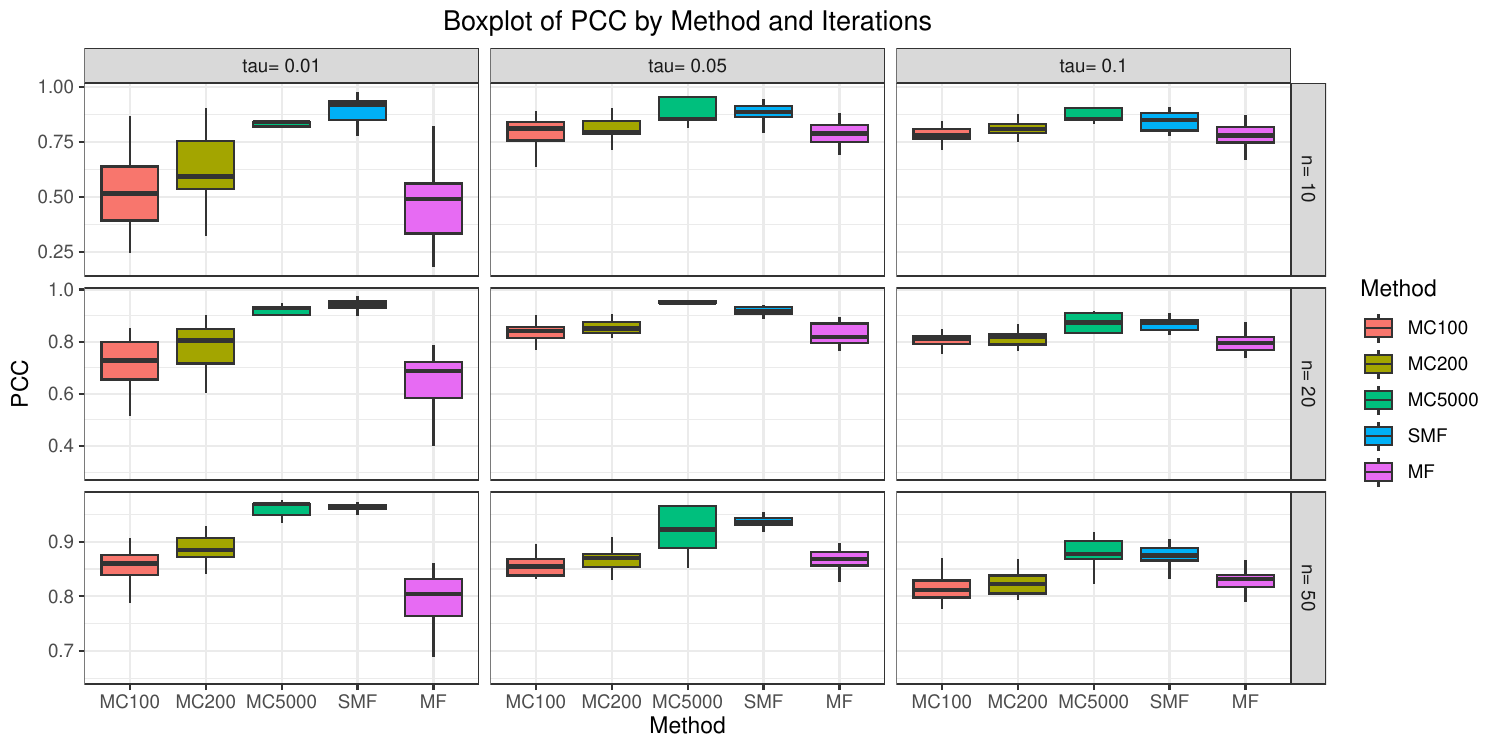}
\caption{ {Boxplots comparing the estimation accuracy of Pearson Correlation Coefficient (PCC) between the estimated and true connected probabilities for SMF, MF, and various numbers of MCMC iterations. A higher PCC indicates better estimation performance for the corresponding method.  MC100,  MC200 and MC 5000 represent posterior means obtained after 100,  200 and 5000 iterations of Gibbs samplers, respectively, with the first half of iterations discarded as burn-in. Among all cases, SMF achieves a similar level of estimation accuracy with MCMC with 5000 iterations.}}
\label{fig:MCMC_PCC}
\end{figure}

\begin{figure}
\centering
\includegraphics[width=0.8\linewidth]{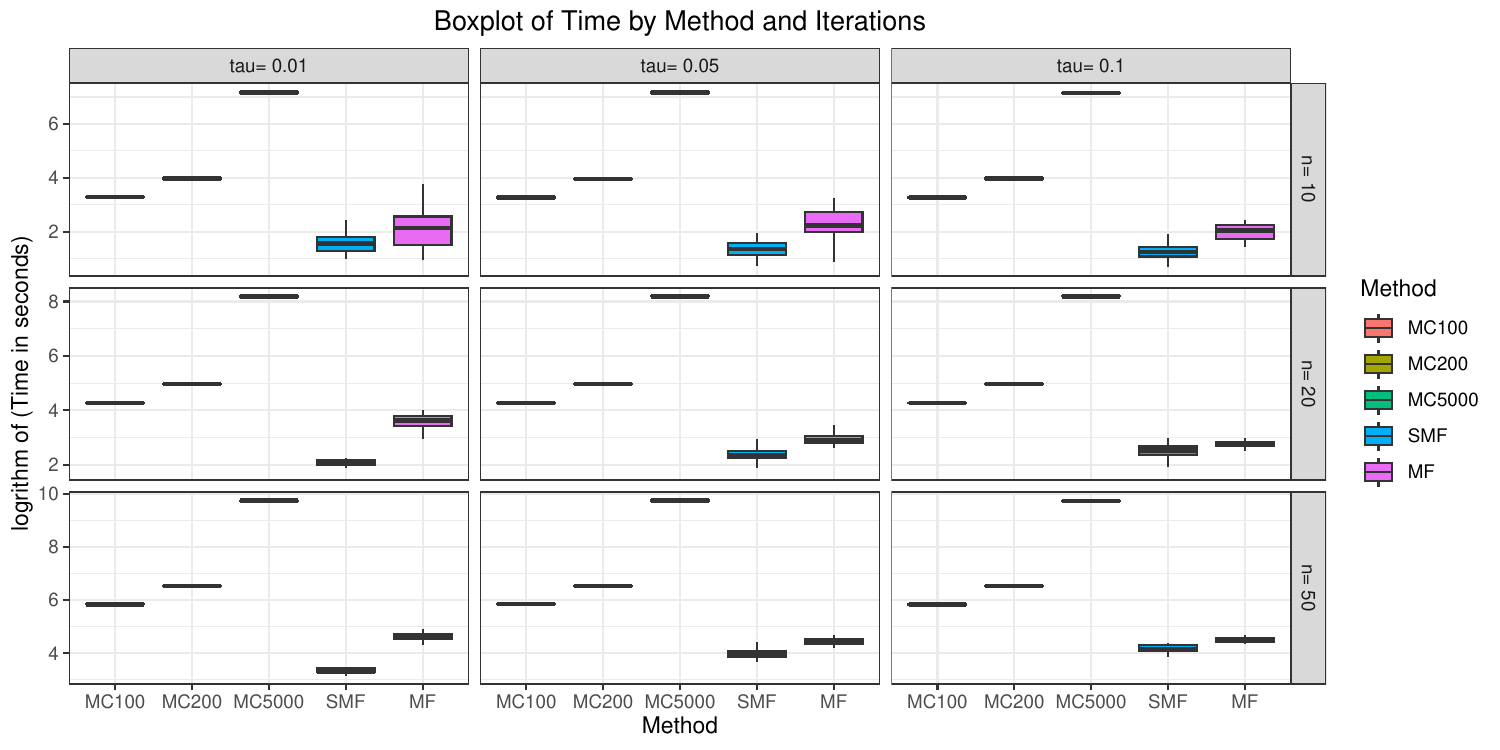}
\caption{ {Boxplots comparing the computation time for SMF, MF, and different numbers of MCMC iterations, as simulated in Figure~\ref{fig:MCMC_PCC}.  MC100,  MC200 and MC 5000 represent 100, 200 and 5000 iterations of the Gibbs samplers, respectively. Remarkably, SMF exhibits the shortest computation time while almost achieving the highest level of estimation accuracy in Figure~\ref{fig:MCMC_PCC} across all cases.}}
\label{fig:MCMC_Time}
\end{figure}

    We compared standard MCMC and SMF in terms of estimation accuracy and computation time for binary networks. We ran MCMC with 100, 200 and 5000 iterations using a Gibbs Sampler algorithm, where each coefficient was sampled from its full conditional distribution. For the MCMC chain, we discarded the first half of iterations as burn-in and used the sample means from the last half of iterations to calculate the estimator. We then compared this accuracy with SMF using the PCC with the true probabilities, as well as considering computation time. We set the transition smoothness $\tau=0.01, 0.05, 0.1$, sample size $n=10, 20, 50$, time point $T=100$, and correlation $\rho=0.5$. The simulations were repeated 25 times for each setting. The results are presented in boxplot comparisons shown in Figure~\ref{fig:MCMC_PCC} and Figure~\ref{fig:MCMC_Time}, which illustrate several noteworthy findings: First, the stronger the dependence across time, the better the performance of SMF in terms of higher PCC accuracy when $\tau$ decreases from 0.1 to 0.01. This is because the computation of SMF incorporates the dependence across time, resulting in improved performance. However, for MCMC, the weaker the dependence across time, the better the performance in terms of higher PCC accuracy when $\tau$ increases from 0.01 to 0.1. This is because the mixing of the Markov Chain is affected negatively by the dependence across time. Similarly, for MF, the weaker the dependence across time, the better the performance in terms of higher PCC accuracy when $\tau$ increases from 0.01 to 0.1, as weaker dependence will better fit the independence structure of MF. Increasing $\tau$ reduces the gap in estimation accuracy for MCMC among iterations 100,  200 and 5000, indicating faster mixing of the Markov Chains. Overall, SMF requires less computation time than MCMC and MF under the given settings while achieving almost the best estimation accuracy, which is similar to MCMC with 5000 iterations. This indicates that when the dependence across time is strong, SMF significantly improves computation efficiency.

	In Table~\ref{tab:3}, we report the median of PCC from 25 simulation experiments for binary networks with $n=100$ and $T=100$. The comparison is between the true and estimated probabilities for SMF with $\sigma_0 = 0.5$ and a known $\tau$, versus adaptive SMF using \eqref{eq:prior_sd}. It's interesting to note that learning the initial and transition variances adaptively using the prior~\eqref{eq:prior_sd} doesn't lead to any loss of accuracy compared to when these parameters are known as a priori.
	
	\begin{table}[ht]
		\centering
		\begin{tabular}{r|lllllllllllll}
			\hline
			$\tau$           &\multicolumn{3}{c}{0.01}   & \multicolumn{3}{c}{0.1} \\
			\hline
			$\rho$   & 0 & 0.4 & 0.8  & 0 & 0.4 & 0.8 \\ 
			\hline
			{Adaptive SMF}      & 0.885 & 0.888 & 0.892 & 0.880 & 0.910 & 0.914 \\ 
			
			{SMF}              & 0.887 & 0.881 & 0.898 & 0.897 & 0.912 & 0.904  \\ 
			\hline
			$\tau$  & \multicolumn{3}{c}{0.2} &\multicolumn{3}{c}{0.3}\\ 
			\hline
			$\rho$  & 0 & 0.4 & 0.8 & 0 & 0.4 & 0.8 \\
			\hline
			{Adaptive SMF}     & 0.907 & 0.915 & 0.918 & 0.908 & 0.919 & 0.923   \\ 
			
			{SMF}               & 0.895 & 0.918 & 0.915 & 0.904 & 0.919 & 0.931 \\
			\hline
		\end{tabular}
  \caption{Performance comparisons for binary networks between adaptive SMF and SMF with known initial and transition variances. The measure compared are the medians of Pearson correlation coefficient (PCC) between true and estimated probabilities of the repeated simulations.}\label{tab:3}
	\end{table}

\subsection{Enron Email}\label{sec:4.2}
	Using the Enron email data set \citep{klimt2004enron}, we compare our model with the latent space model with the same likelihood but with an inverse Gamma prior on the transition variance. Enron data consists of emails collected from 2359 employees of the Enron company. From all the emails, we examine a subset consisting of $n=184$ employees communicating among $T=44$ months from Nov. 1998 to June 2002 recorded in the \texttt{R} package \texttt{networkDynamic} \citep{butts2020networkDynamic}.  The networks depict the email communication status of employees over that period. The edges in the network are ones if one of the corresponding two employees sent at least one email to the other during that month. According to the data set, all networks are sparse and many edges remain unchanged over time. The aim of this study is to determine whether shrinkage on transitions induced by Gamma prior on transition variance can be beneficial for sparse dynamic networks. 
With the dynamic networks, we consider all the edges to be missed with probability $p=0.01,0.02,...,0.1$ independently, train the two latent space models without the missed data, and then make predictions based on the missed data. We use two criteria for comparison: the testing AUC score and the ratio of true positive detection over all missed edges, which is defined as the ratio of predictive probability greater than 0.5 when the true edge value is $1$ over all missed edges. Since all networks are extremely sparse and negative predictions are trivial, the second criterion above is meaningful. 
	The same SMF variational inference method is used in both latent space models. In both of the latent space models, we assign a latent dimension of $5$ (more results about $d=2,3,4$ are provided in Section~\ref{sec:additional_simu} in the appendix), the same initialization and stopping criteria. The variational mean of the latent positions is used to estimate the latent positions.

       \begin{figure}
              \centering
              \includegraphics[width=0.8\linewidth]{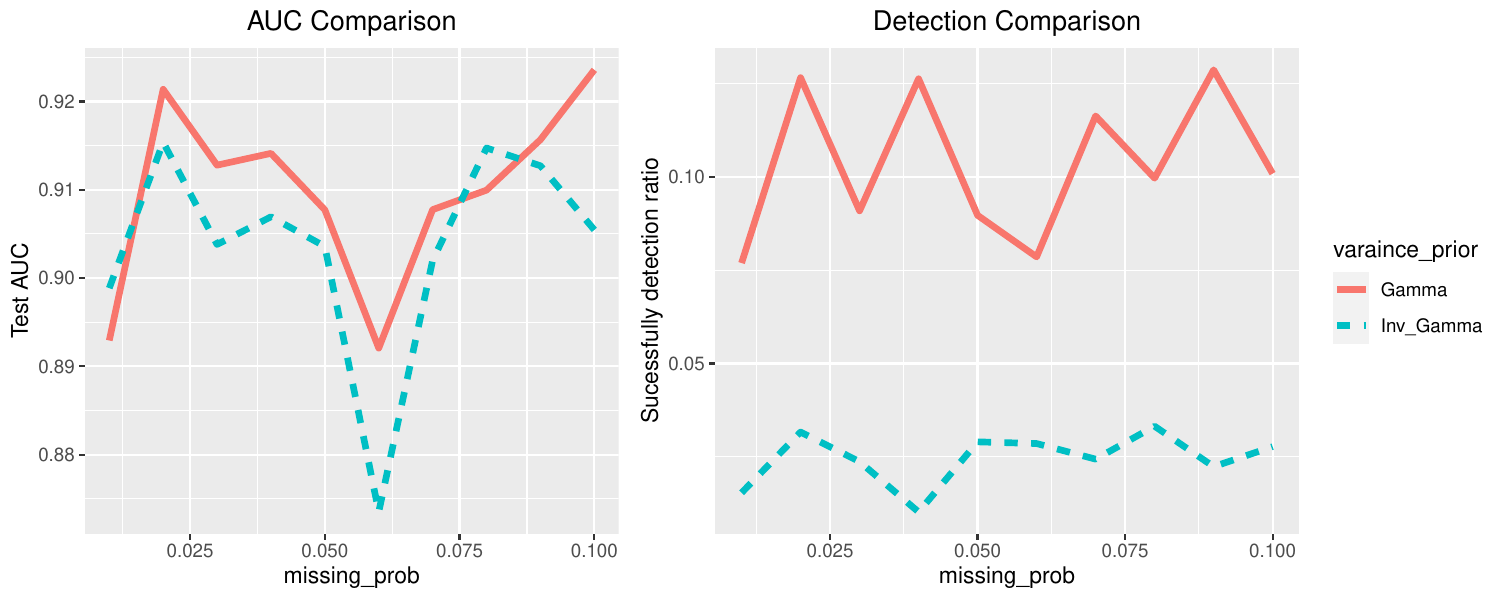}
               \caption{ Comparisons of latent space models between Gamma or Inverse Gamma priors on the Enron email testing data.}
	               \label{fig:enron}
	      \end{figure}
	
Figure~\ref{fig:enron} illustrates a performance comparison between the two approaches. The Gamma prior leads to a better fit based on the AUC comparison (left subfigure) and improves the detection of missed links (right subfigure). A Gamma prior shrinks the transitions more compared to an inverse-Gamma prior so that if two employees communicate at time $t$, the predictive probability for them to communicate at time $t+1$ is high. 
	
	\subsection{McFarland Classroom}
	McFarland’s streaming classroom data set provides interactions of conversation turns from streaming observations of a class observed by Daniel McFarland in 1996 \citep{mcfarland2001student}. The data set is available in the \texttt{R} package \texttt{networkDynamic} \citep{butts2020networkDynamic}. The class comprised of $2$ instructors and $18$ students. Of the $2$ instructors, one is the main instructor who lectured most of the time, while the other is an assistant. During the class, the instructors began by providing instructions to all students. Then, the students were divided into groups and assigned collaborative group work. The two instructors oversaw the activities across the groups to assist the students.  Here, we aim to compare MF and SMF via prediction accuracy and visualize the dynamic evolution of the latent positions.

\begin{figure}
	\centering
		\includegraphics[width=0.6\linewidth]{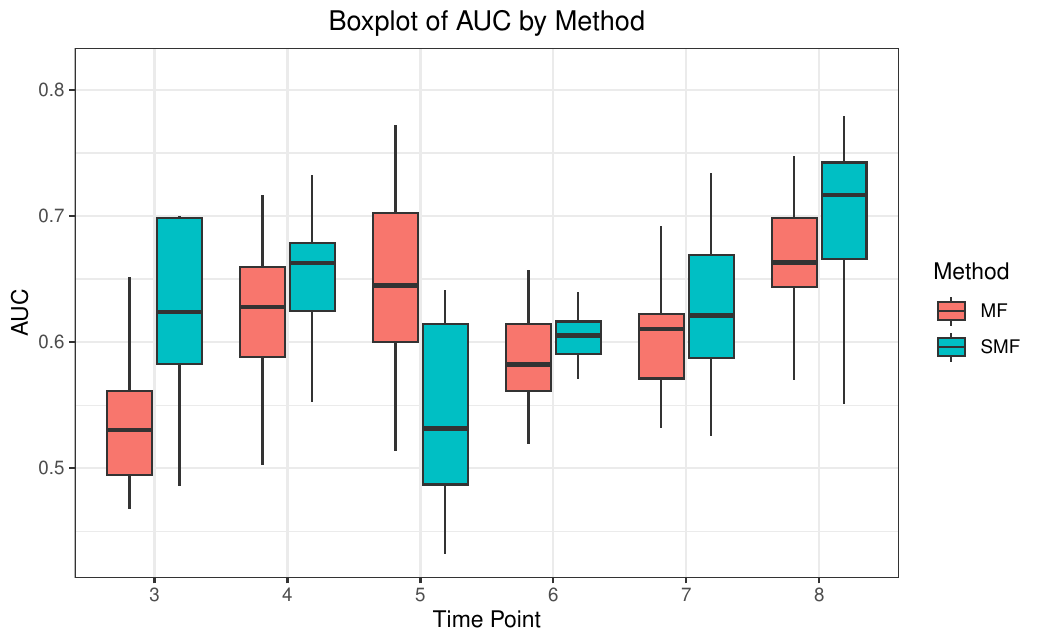}
		\caption{{ {Boxplot comparing AUC Predictions between SMF and MF for Time Points $t=3$ to $8$, using different algorithm initializations based on networks from previous time points. SMF outperforms MF consistently, except at time point $5$, where significant structural changes in the network may hinder the benefit of temporal dependencies across time points.}}}
		\label{fig:classroom_boxplot}
	\end{figure}

\begin{sidewaysfigure}
		\centering
		\includegraphics[width=0.9\linewidth]{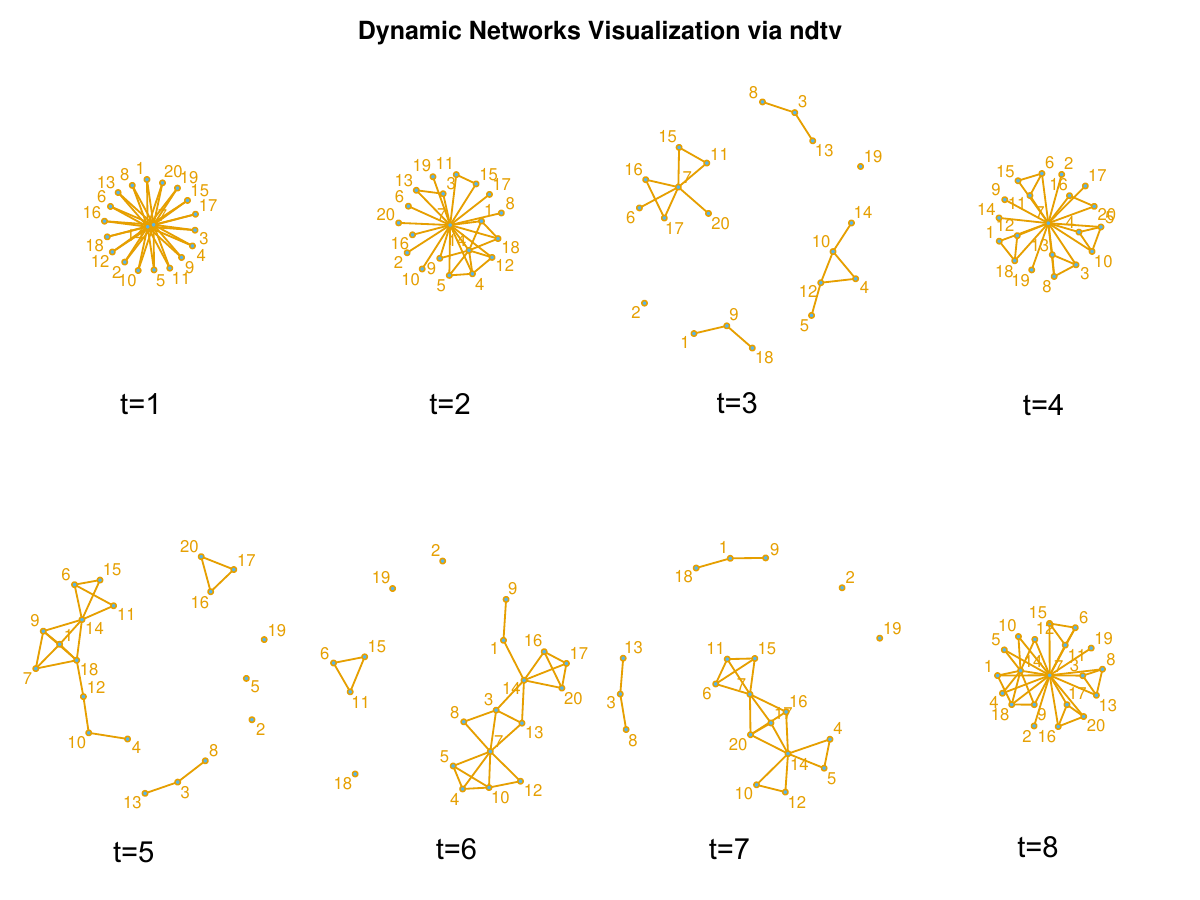}
		\caption{Visualization of the network dynamics from time point $1$  to $8$ for McFarland classroom data set using R package \texttt{ndtv}. From top left to top right, we have $t=1,...,4$; from bottom left to bottom right, we have $t=5,...,8$. Each number associated with the point is the index of the node.}
		\label{fig:class_2}
	\end{sidewaysfigure}

	    We divide the entire class time into $8$ equispaced time points.  The edges of each of the $8$ networks represent whether the two nodes interacted related to the study task during the entire time period. We chose $d=2$ for visualization purposes. A {$\m N(0,10)$}  prior is placed on the intercept and the  prior~\eqref{eq:prior_sd} is adopted for both SMF and MF. 	First, we compare SMF and MF in terms of prediction accuracy. For $t=3,4,...,8$,  the first $t-1$ networks are used as the training data, while the $t$-th network is used as the test data. The estimated latent positions at time point $t-1$ are used to predict the probabilities of edges between any two nodes at time $t$. Then the test AUC scores are obtained from the above-estimated probabilities vs. the true binary responses at time point $t$. We repeat the process for $25$ times with different initial values. The boxplots of the test AUC scores for MF and SMF are shown in Figure~\ref{fig:classroom_boxplot}. From the figure, we can see that except for time point $t=5$, where the network structure changed significantly and the dependence from previous time points may not be meaningful (see Figure~\ref{fig:class_2} for the change of the connections), SMF consistently performs better than MF, which again testified to the ability for SMF to capture the dependence across time better.

\begin{sidewaysfigure}
		\centering
		\includegraphics[width=0.9\linewidth]{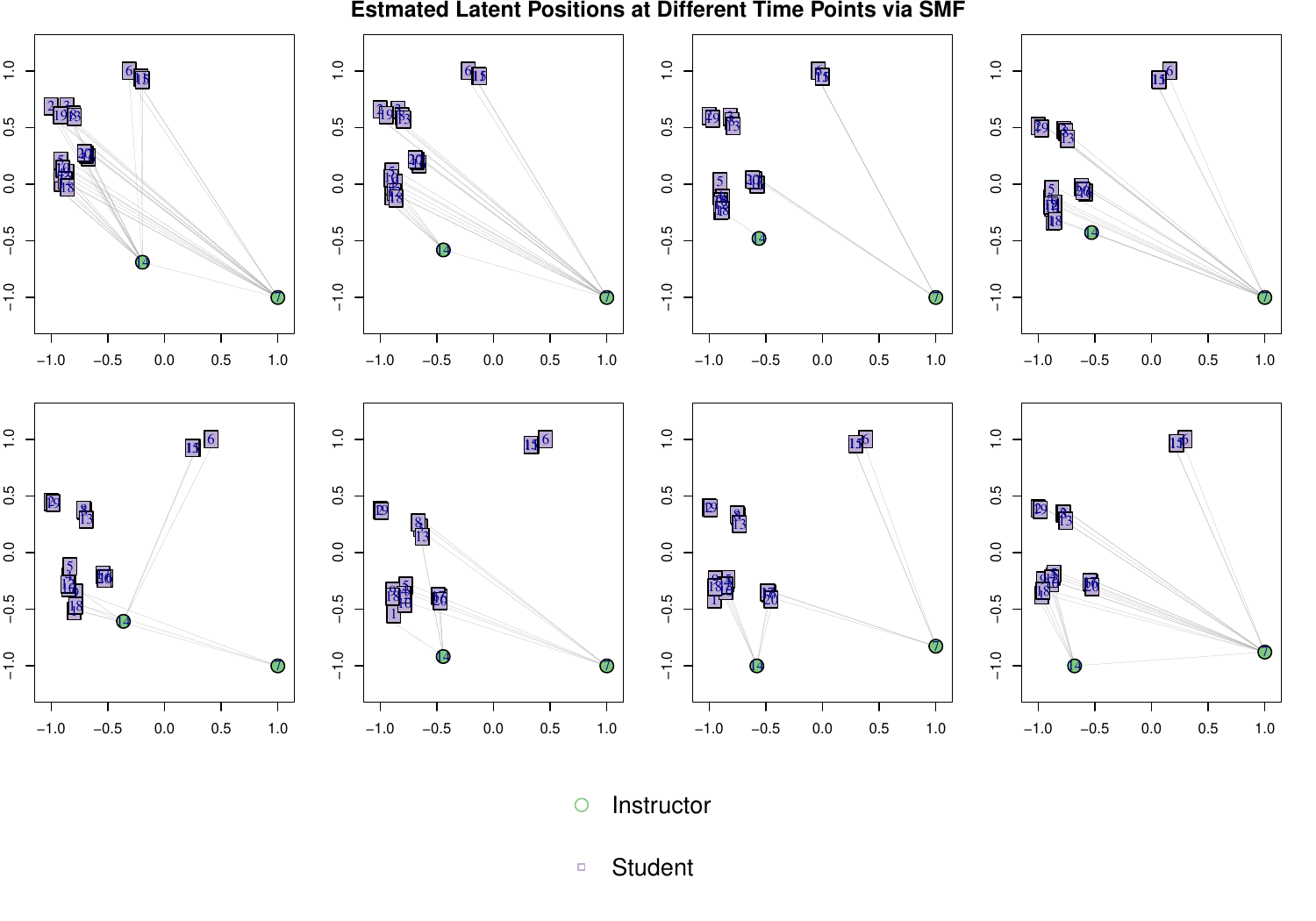}
		\caption{{ Dynamics of latent positions of all nodes from time point $1$  to $8$ for McFarland classroom data set via SMF. The locations of the nodes are the estimated latent positions. The top row of figures represents time points 1, 2, 3, and 4, while the bottom row represents time points 5, 6, 7, and 8. The edges between two nodes imply the interaction between the two observations within the corresponding time. Each number associated with the point is the index of the node.} }
		\label{fig:class}
	\end{sidewaysfigure}

\begin{sidewaysfigure}
		\centering
		\includegraphics[width=0.9\linewidth]{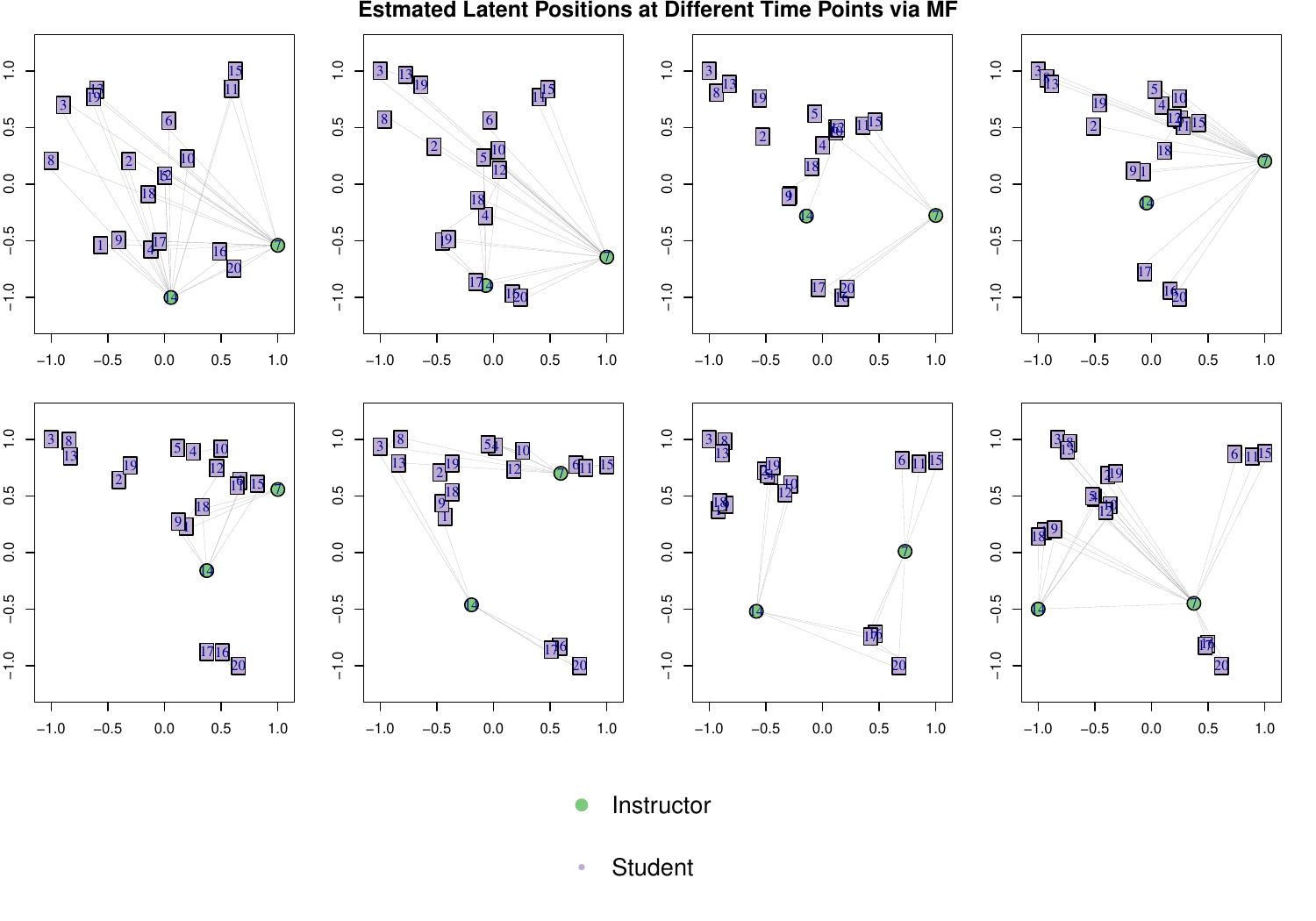}
		\caption{{Dynamics of latent positions of all nodes from time point $1$  to $8$ for McFarland classroom data set via MF. The locations of the nodes are the estimated latent positions. The top row of figures represents time points 1, 2, 3, and 4, while the bottom row represents time points 5, 6, 7, and 8. The edges between two nodes imply the interaction between the two observations within the corresponding time. Each number associated with the point is the index of the node. }}
		\label{fig:class_MF}
	\end{sidewaysfigure}

	Next, we implemented SMF with the networks at all 8 time points under the same hyperparameter specification to visualize the dynamic evolution of the latent positions.  Since the latent positions estimated directly from the algorithm are not identifiable, Procrustes rotation is performed \citep{hoff2002latent} where the latent positions of time $t=2,..,T$ are projected to the locations that are most close to its previous locations ($t=1,...,T-1$) through Procrustes rotation. Observe that the inner product is invariant to this transformation. Figure~\ref{fig:class} shows the dynamic evolution of the variational mean of the latent positions for both students and instructors (an animated version of Figure~\ref{fig:class} is provided in the supplementary material). 
 At time point $1$ (i.e., at the beginning of the class) the students indexed by $\{1,...,20\} \backslash\{7,14\}$ are approximately grouped into the following clusters $(6,11,15)$, $(3,8,13)$, $(10,12,4,5)$, $(1,18,9)$, $(2, 19)$ and $(20,17,16)$. The locations of the students remained the same until time point $4$.  From time point $4$ to $5$,  the inner-group distances between  $(1,9,18)$ and $(4,5,10,12)$ became smaller, which reflected the real scenario that the students were assigned into groups.   Then the group structure of the students remained similar for the remainder of the class. Overall, the evolution reflected the collaborative behavior between certain groups as they performed specific tasks during the class.
	As a point of comparison, we also obtained dynamic visualization of the networks via MF (Figure~\ref{fig:class_MF}) and the popular \texttt{ndtv} package (ndtv: Network Dynamic Temporal Visualization, \citealp{skye2021ndtv}). Although \texttt{ndtv} package is known for its dynamic networks visualizations through animations, static snapshots of the visualizations can also be created using \texttt{filmstrip} function (Figure~\ref{fig:class_2}). First, unlike Figure~\ref{fig:class}, the latent positions estimated via MF in Figure~\ref{fig:class_MF} did not have a smooth temporal evolution,  as the MF assumed independence across the time points.  
    In addition, compared to our visualization in Figure~\ref{fig:class}, results from the \texttt{ndtv} package in Figure~\ref{fig:class_2} lacked a clear pattern of the network evolution. 
	For example, the students indexed $\{1,18,9\}$ stayed close to each other at time points $t=5,6,7$ in Figure~\ref{fig:class}, while in Figure~\ref{fig:class_2}, $18$ is far away from $(1,9)$ at time $t=6$, while being connected to $1$ at the neighboring time points $t=5,7$.  
A similar phenomenon can be seen for student indexed $5$ at time $t=5$, where in Figure~\ref{fig:class} it is close to $(4,10,12)$ while in Figure~\ref{fig:class_2} it is not. The ability of our methodology to borrow information across time is specifically due to the Markovian structure \eqref{eq:prior} imposed on the evolution of the latent positions endowed with the Gamma prior~\eqref{eq:prior_sd} on the transition variance, allowing sufficient probability near the origin. 
	Thus our methodology revealed a more realistic pattern in the evolution in  Figure~\ref{fig:class} compared to MF and \texttt{ndtv} as most of the detected changes remained concentrated in time $t=4,5$ for the students (when the students formed groups) and $5,6,7$ for the instructors (after the instructors began assisting the students).

	\section{Discussion}\label{sec:5}
	There are a number of potential extensions of the proposed methodology and theory in this article. Properties of the Gaussian random walk prior is crucially exploited in our theory to obtain the optimal variational risk. It would be interesting to explore similar theoretical optimality results for Gaussian Process priors (e.g., \citealp{durante2017bayesian}). Moreover, the theoretical analysis of the lower bound can be extended to the case that the true latent positions evolve smoothly over time, like in \cite{pensky2019dynamic}.
	
	From a methodological point, it is of interest to explore how to perform community detection after estimating the latent positions. As the latent positions are characterized as vectors in Euclidean space, it is natural to consider some distance-based approaches like K-means for clustering. Adapting to the dimension $d$ of the embedding space is also a challenging problem. Finally, it is also interesting to explore dynamic latent space models with other complex data to fit real-world scenarios, such as continuous-time networks \citep{loyal2024fast} and dynamic networks with sudden structural changes \citep{zhao2022factorized}.

\acks{We are grateful to the action editor and four reviewers for their valuable suggestions and kind help, which significantly improved the quality of the paper.	The research is supported by the National Science Foundation CCF-1934904. Drs. Pati and Bhattacharya acknowledge support from NSF DMS 2210689.}

Reproducible implementations and experiments are publicly available at \url{https://github.com/pengzhaostat/SMF-structured-variational-inference}.

	\newpage
	\appendix
	\setcounter{equation}{0}
	\renewcommand\theequation{A.\arabic{equation}}
	\section{Appendix}

	\subsection{Proof of Theorem~\ref{thm:lower_minimax}}
	Within the networks, we adopt the hypotheses constructions for some low-rank matrices, while among the networks, we adopt the test constructions similar to the constructions in total variational literature~\citep{padilla2017dfs}. 
	
	For $\m U=\{\+U_t\}_{t=1}^T$ with $\+U_t =[\+u_{1t},...,\+u_{nt}]'$ and $\m V=\{\+V_t\}_{t=1}^T$ with $\+V_t =[\+v_{1t},...,\+v_{nt}]'$, let 
	\begin{equation*}
	d^2(\m U, \m V) = \sum_{t=1}^T\sum_{i\ne j=1}^n (\+u_{it}'\+u_{jt}-\+v_{it}'\+v_{jt})^2,
	\end{equation*}
	and
	\begin{equation*}
	d_0^2(\+U_t, \+V_t) = \sum_{i\ne j=1}^n (\+u_{it}'\+u_{jt}-\+v_{it}'\+v_{jt})^2.
	\end{equation*}

	\textbf{Hypothesis constructions for the low-rank part\\}
	First, we need the following lemma to obtain sparse Varshamov-Gilbert Bound under Hamming distance for the low-rank subset construction:
	\begin{lemma}[Lemma 4.10 in \citealp{massart2007concentration}]\label{lemma VG}
		Let $\Omega =\{0,1\}^n$ and $1 \leq s \leq n/4$. Then there exists a subset $\{w^{(1)},...,w^{(M)}\} \subset \Omega$ such that 
		\begin{enumerate}
			\item $\|w^{(i)}\|_0 =s$ for all $1 \leq i \leq M$;
			\item $\|w^{(i)}-w^{(j)}\|_0 \geq s/2$ for $0\leq i \ne j \leq M $;
			\item $\log M \geq cs \log(n/s)$ with $c \geq 0.233$.
		\end{enumerate}
	\end{lemma}
	
	Let $\Omega_M =\{\+w^{(1)},...,\+w^{(M)}\}  \subset \{0,1\}^{(n-d+1)/2} $ constructed based on the above Lemma (the construction holds under $n-d+1\geq 8$). For each $\+w$, we can construct a $n \times d$ matrix as follows:
	\begin{equation}\label{eq:construction}
	\+U^{w}=\begin{bmatrix}
	\+v^{w} &\+0 \\
	\+0  & \+I_{d-1}
	\end{bmatrix} \quad  \mbox{with} \quad \+v^w =\begin{bmatrix}
	1 \\...\\1 \\\epsilon \+w
	\end{bmatrix} \in \mathbb{R}^{n-d+1}, w \in \Omega_M
	\end{equation}
	where the first $(n-d+1)/2$ components for $\+v^w$ are all ones.

	The effect of this construction is that: for different $\+w_1,\+w_2 \in \Omega_M$,  since $s/2 \leq  \|\+w_1-\+w_2\|_0 \leq 2s$ and $\|\+U^w\|_F \leq \sqrt{n}$, we have
	\begin{align*}
	d_0(\+U^{w_1}, \+U^{w_2}) \leq \|\+U^{w_1}\+U^{'w_1}-\+U^{w_2}\+U^{'w_2} \|_F \leq \|\+U^{w_1}(\+U^{'w_1}-\+U^{'w_2} )\|_F+\|(\+U^{w_1}-\+U^{w_2})\+U^{'w_2} \|_F \\
	\leq 2\sqrt{n}\|\+U^{w_1}-\+U^{w_2}\|_F  = 2\sqrt{n}\|\+v^{w_1}-\+v^{w_2}\|_2\leq 2\sqrt{{2ns}}\epsilon.
	\end{align*} 
	In addition, consider $A:=\{i+(n-d+1)/2 : w_{1i} \ne 0\}$, $B:= \{j+(n-d+1)/2:w_{2j} \ne 0\}$, $C:=A \cap B$, where $w_{1i},w_{2j}$ are $i$ and $j$th component of $w_1$ and $w_2$. We have $|C| \leq s/2$, $|A-C|\geq s/2$ and $|B-C|\geq s/2$. By direct calculation, we have 
	
	$$d^2_0(\+U^{w_1}, \+U^{w_2}) = \sum_{i\ne j}^{n-d+1}( v^{w_1}_{i} v^{w_1}_{j}- v^{w_{2}}_i v^{w_2}_j  )^2.$$
	
	By only considering the sum for $i  \in \{1,...,(n-d+1)/2\}, j\in A-C$ where $v_i^{w_1}=v_i^{w_2} = 1$ $v^{w_1}_{j} =\epsilon $ and $v^{w_2}_j =0$ and $i \ne j$,  we have
	\begin{align*}
	d^2_0(\+U^{w_1}, \+U^{w_2}) \geq \sum_{i=1}^{(n-d+1)/2}\sum_{j \in A-C}(   \epsilon {w_1}_{j}-  \epsilon {w_2}_j  )^2\geq \frac{s(n-d+1)}{4} \epsilon^2.
	\end{align*}

	\textbf{Hypothesis constructions for the total variational denoising part\\}
	
	As in the total variation denoising literature, we partition the set $\{1,...,T\}$ into $m$ groups $S_1,S_2,...,S_m$ such that $S_1=\{1,...,k\}$, $S_2=\{k+1,...,2k\}$, ..., $S_m=\{(m-1)k+1,...,T\}$, where $k$ will be decided later. Then we have $k(m-1)+1 \leq T \leq km.$ For simplicity, we assume the partition is even $T=km$, otherwise we can consider $T' = km $, which has the same rate with $T$ since $km>T>km-k+1$.  As in the literature in nonparametric regression, we need to obtain the optimal order of $k$ or $m$.

	Let $\+w_0= [0,...,0]' \in \mathbb{R}^{(n-d-1)/2}$ and 
	\begin{equation*}
	\+U^{0}=\begin{bmatrix}
	\+v^{w_0} &\+0 \\
	\+0  & \+I_{d-1}
	\end{bmatrix} \quad  \mbox{with} \quad \+v^{w_0} =\begin{bmatrix}
	1 \\...\\1 \\ \+w_0
	\end{bmatrix} \in \mathbb{R}^{n-d+1}
	\end{equation*}
	and $\m U^0 = [\+U^0,...,\+U^0]$. We  need the Varshamov-Gilbert Bound~\ref{lemma VG} again to introduce another binary coding: let $\Omega_r =\{\+\phi^{(1)},...,\+\phi^{(M_0)}\}  \subset \{0,1\}^{m} $, such that $\|\+\phi^{(i)}\|_0=s_0$ with $s_0\leq m/4$ for all $1\leq i \leq M_0$ and and $\|\+\phi^{(i)}-\+\phi^{(j)}\|_0 \geq s_0/2$ for $0\leq i <j \leq M_0 $ and $\log M_0 \geq cs_0 \log(m/s_0)$ with $c \geq 0.233$. The construction holds under $m\geq 4$.
	
	Then the construction is based on a mixture of product space of $\Omega_M$ and group structure for $S_1,...S_m$: 
	\begin{equation}\label{eq:construct}
	\begin{aligned}
	\Theta_\epsilon =\{\m X^{(w,\phi)}_{...}:&\\
	&  \+X^{(w,\phi)}_t =\+U^{w^{(i)}}, \quad \forall t \in S_j \quad \mbox{if}\quad  \phi_j=1,\\
	&\+X^{(w,\phi)}_t =\+U^{0}, \quad \forall t \in S_j, \quad  \mbox{if}\quad  \phi_j=0, \\
	& \+w^{(i)} \in \Omega_M, \, \forall i=1,...,s_0, \, \+w^{(i)}\, \mbox{are chosen with replacement}, \+\phi \in \Omega_r\},
	\end{aligned}
	\end{equation}
	For example, when $\phi =(0,1,0,1,0,1,...)$,  $\+X^{(w,\phi)}_1$,..,$\+X^{(w,\phi)}_T$ is:
	$$
	\underbrace{\+U^{0},...,\+U^{0}}_{|S_1|},\underbrace{\+U^{w^{(1)}},...,\+U^{w^{(1)}}}_{|S_2|},\underbrace{\+U^{0},...,\+U^{0}}_{|S_3|},\underbrace{\+U^{w^{(2)}},...,\+U^{w^{(2)}}}_{|S_4|},...
	$$
	We have $|\Theta_\epsilon| = M_0 M^{s_0} $.  In addition, for $\m U, \m V \in \Theta_\epsilon$, we have 
	\begin{equation*}
	d^2(\m U,\m V) = \sum_{t=1}^T d_0^2(\+U_t, \+V_t) \geq k s_0 \frac{s(n-d+1)}{8} \epsilon^2 .
	\end{equation*}

	Besides, the KL divergence between any elements $\m U \in \Theta_\epsilon$ and $\m U^0$ can be upper bounded:
	\begin{equation}\label{eq:KL_1}
	D_{KL}(P_{\m U} \,||\, P_{\m U^0}) \leq C_0 d^2(\m U,\m V) \leq 16 C_0 kns_0s \epsilon^2,
	\end{equation}
	for some constant $C_0>0$ ($C_0=1$ for the binary case, $C_0=1/(2\sigma^2)$ for the Gaussian case).
	
	We use the following lemma to finally obtain the minimax lower bounds.
	
	\begin{lemma}[Theorem 2.5 in \citealp{tsybakov2008introduction}]
		Suppose $M \geq 2$ and $(\Theta,d)$ contains elements $\theta_0,...,\theta_M$ such that $d(\theta_i,\theta_j)\geq 2s >0$ for any $0\leq i\leq j \leq M$ and $\sum_{i=1}^{M}D_{KL}(P_{\theta_i},P_0)/M \leq \alpha \log M$ with $0<\alpha<1/8$. Then we have
		$$\inf_{\hat{\theta}} \sup_{\theta \in \Theta} P_\theta (d(\hat{\theta},\theta)\geq s) \geq \frac{\sqrt{M}}{1+\sqrt{M}} \left(1-2\alpha-\sqrt{\frac{2\alpha}{\log M}} \right).$$
	\end{lemma}

	To adopt the above Lemma, it suffices to show 
	\begin{equation*}
	16 C_0ks_0sn \epsilon^2 \leq \alpha \log  (M_0M^{s_0})  =  \alpha \log (|\Theta_\epsilon|),  
	\end{equation*}
	with $\alpha < 1/8$. Let $s= (n-d-1)/8$, $s_0= m/4$,  according to lemma~\ref{lemma VG}, it's enough to set 
	\begin{equation}\label{eq:KL_upper_by_cardinality}
	Tn(n-1) \epsilon^2 \leq  \frac{cm\log 4}{4}+\frac{c m (n-d-1)\log 4 }{16} = \frac{c m (n-d+7)\log 4 }{16},
	\end{equation}
	with $c=0.233/C_0$.

	\textbf{Minimax rate for point-wise dependence\\}
	Based on our construction, $  2(m-1)s  \epsilon \leq L$ should be satisfied, and we consider the following different cases:
	
	Case 1: If there exist constants $c_0, c_0'>0$ such that $ c_0' /\sqrt{nT}<L \leq c_0 (T-1) n^{1/2}$, which results in $16^3 T L^2/\{4 s c  \log 4\} < T(T-1)^2$, and $16^3 T  L^2/(4 c  \log 4 s)>36 = 4*(4-1)^2$.
	Therefore, since $n\geq 2d$, by assigning $\epsilon=L/\{2(m-1)s\}$  it is enough to let $m$  satisfy
	\begin{equation*}
	\frac{TL^2}{(m-1)^2 } \leq \frac{4 c m s \log 4 }{16^3},
	\end{equation*} which is
	\begin{equation}\label{eq:m}
	m(m-1)^2 \geq \frac{ 16^3  T  L^2}{4 c  \log 4 s},
	\end{equation}
	and $m$ can be chosen within $  4 \leq m \leq T$.
	Let $m$ be the least integer such that the above inequality hold, then there exists a constant $c_2$, such that  $m \leq c_2 T^{1/3} L^{2/3}n^{-1/3} $, which implies 
	$$ k s_0 \frac{s(n-d+1)}{8} \epsilon^2 \gtrsim  L^{\frac{2}{3}} n^{\frac{2}{3}} T^{\frac{1}{3}}.$$

	Case 2: If there exists a constant $c_0>0$ such that $ L > c_0  (T-1)\sqrt{n}$, which results in $ 16^3 T L^2/\{c  s\log 4\} > T(T-1)^2$, then we choose $m=T$, $\epsilon =c_0/\sqrt{n}$ such that $ (m-1) s \epsilon \leq c_0(T-1)\sqrt{n} \leq L/2$. Then 
	$$  k s_0 \frac{s(n-d+1)}{8} \gtrsim Tn.$$
	Case 3: If $L<c_0'/\sqrt{nT}$ for some constant $c_0'>0$ such that the least integer solution of inequality~\eqref{eq:m} satisfying $m< 4$. Then the above hypothesis construction in Equation~\eqref{eq:construct} doesn't hold. Instead of considering the construction in Equation~\eqref{eq:construct}, we consider $T$ copies of the same matrix, which implies the choice of $m$ is $1$. Note that the constraint on the norm of the difference of the matrix is automatically satisfied when all matrices are the same. By constructing the following subset
	\begin{equation}\label{eq:construct2}
	\begin{aligned}
	\Theta_\epsilon =\{\m X^{(w)}:  \+X^{(w)}_t =\+U^{w}, \quad \forall t =1,...,T, \,\+w \in \Omega_M\}.
	\end{aligned}
	\end{equation}
	the KL divergence between any elements $\m U\in \Theta_\epsilon$ and $\m U^0$ can be upper bounded:
	\begin{equation}\label{eq:KL_2}
	D_{KL}(P_{\m U} \,||\, P_{\m U^0}) \leq C_0d_0^2(\+U_t, \+V_t) \leq 16C_0 T s n\epsilon^2,
	\end{equation}
	for some constant $C_0>0$. 
	Then it suffices to let 
	\begin{equation*}
	16 T sn \epsilon^2 \leq  \frac{c (n-d-1) \log 4}{16} \leq \alpha \log  (M) .
	\end{equation*}
	Therefore, based on the above equation, we need to choose $\epsilon=\sqrt{1/(nT)}$.  Then we have
	$$ \frac{k s_0 s (n-d+1) \epsilon^2}{8} \gtrsim n.$$
	
	Finally, based on Markov's inequality, by combining the above three cases, we have
	
	\begin{equation*}
	\begin{aligned}
	\inf_{\hat{\m X}}\sup_{\m X \in \Theta_\epsilon} \E_{\m X} \left[\frac{1}{T n (n-1)} d^2(\hat{\m  X}, \m X) \right]   
	\gtrsim  \min \left\{\frac{L^{\frac{2}{3}}}{n^{\frac{4}{3}} T^{\frac{2}{3}}},\frac{1}{n}\right\}+\frac{1}{nT}.
	\end{aligned}
	\end{equation*}
	Therefore, the final conclusion holds.

	\subsection{Proof of Theorem~\ref{cor:NWD}}\label{proof:thm2a}
	\begin{proof}
		 As discussed in \cite{bhattacharya2019bayesian}, under the prior concentration condition that 
		\begin{equation*}
		\Pi(B_{n,T}(\m X^*;\epsilon_n)) \geq e^{-Tn(n-1)\epsilon_{n,T}^2}, 
		\end{equation*}
		we can obtain the convergence of the $\alpha$-divergence:
		\begin{equation*}
		D_\alpha(p_{\m X},p_{\m X^*}) = \frac{1}{\alpha-1} \log \int (p_{\m X^*})^\alpha (p_{\m X})^{1-\alpha} d\mu.
		\end{equation*}

		Based on calculation, for Gaussian likelihood, we have $\max\{D_{KL}(p_{\m X},p_{\m X^*}),V_2(p_{\m X},p_{\m X^*})\}\lesssim \sum_{i\ne j,t}(\+x_{it}'\+x_{jt}-\+x^{*'}_{it}\+x^{*}_{jt})^2$ where $V_2(p_{\m X},p_{\m X^*})$ is the second moment of KL ball. For the Bernoulli likelihood, by Lemma~\ref{lem:binary_KL}, we have
		\begin{equation*}
		D_{KL}(p_{\m X},p_{\m X^*}) =\int p_{\m X^*}\log(\frac{p_{\m X^*}}{p_{\m X}}) d\mu \leq \sum_{t=1}^{T}\sum_{i\ne j=1}^n(\+x'_{it}\+x_{jt}-\+x^{*'}_{it}\+x^{*}_{jt})^2.
		\end{equation*}
		Moreover, we have
		\begin{equation}\label{eq:V2}
		V_2(p_{\m X},p_{\m X^*}):=      \int p_{\m X^*}\log^2(\frac{p_{\m X^*}}{p_{\m X}}) d\mu \leq  \sum_{i\ne j=1}^{n}\sum_{t=1}^T2 p_{x^*_{it},x^*_{jt}}(\log \frac{p_{x^*_{it},x^*_{jt}}}{p_{x_{it},x_{jt}}})^2+2(1-p_{x^*_{it},x^*_{jt}})(\log \frac{1-p_{x^*_{it},x^*_{jt}}}{1-p_{x_{it},x_{jt}}})^2.
		\end{equation}
		Under the conditions that $p_{x^*_{it},x^*_{jt}}:= 1/\{1+\exp(-\+x^{*'}_{it}\+x^*_{jt})\}$ is bounded away from $0$ and $1$. The right hand side of Equation~\eqref{eq:V2} is bounded above by $\sum_{i \ne j,t}(\+x_{it}'\+x_{jt}-\+x^{*'}_{it}\+x^{*}_{jt})^2$ multiplied by some positive constant.
		Therefore, we also have $\max\{D_{KL}(p_{\m X},p_{\m X^*}),V_2(p_{\m X},p_{\m X^*})\}\lesssim \sum_{i\ne j,t}(\+x_{it}'\+x_{jt}-\+x^{*'}_{it}\+x^{*}_{jt})^2$ for the binary case.  Hence we only need to lower bound the prior probability of the set $\{\sum_{i \ne j,t}(\+x_{it}'\+x_{jt}-\+x^{*'}_{it}\+x^{*}_{jt})^2\leq n(n-1)T\epsilon^2\} \supset \{\max_t \max_{i \ne j}(\+x_{it}'\+x_{jt}-\+x^{*'}_{it}\+x^{*}_{jt})^2 \leq \epsilon^2\} $. Given $i\ne j, t$ we have
		\begin{equation*}
		\begin{aligned}
		|\+x_{it}'\+x_{jt}-\+x^{*'}_{it}\+x^{*}_{jt}| &\leq |(\+x'_{it}-\+x^{*'}_{it})\+x^{*}_{jt}|+|\+x'_{it}(\+x_{jt}-\+x^{*}_{jt})| \\
		&\leq \max_i \|\+x_{it}-\+x^{*}_{it}\|_2(\|\+x_{it}-\+x^*_{it}\|_2+2\|\+x^*_{it}\|_2) \leq \max_i \|\+x_{it}-\+x^{*}_{it}\|_2(\|\+x_{it}-\+x^{*}_{it}\|_2+2C).
		\end{aligned}
		\end{equation*}
		Then when $\max_i \|\+x_{it}-\+x^{*}_{it}\|_2 \leq \epsilon/\{(2+c_0)C\} \leq C$ for some constants $c_0>1$, we have
		\begin{equation*}
		\max_i\|\+x_{it}-\+x^{*}_{it}\|_2(2C+\|\+x_{it}-\+x^{*}_{it}\|_2) \leq \frac{\epsilon}{(2+c_0)C} 3C \leq \epsilon. 
		\end{equation*}
		Denote $E_0 = \max_i \|\+x_{it}-\+x^{*}_{it}\|_2 \leq \epsilon/\{(2+c_0)C\}$, $E_1 = \{\max_{i,j,t} |(X_{ijt}-X_{ij1})-(X^{*}_{ijt}-X^*_{ij1})| \leq \epsilon_0 \}$, $E_2 = \{\max_{i,j} |X_{ij1}-X^*_{ij1}| \leq \epsilon_0\}$ with $\epsilon_0 =\epsilon/((2+c_0)C\sqrt{d})$. Then we have
		\begin{equation*}
		\begin{aligned}
		\Pi(E_0) \geq \Pi\left(E_1\right) \Pi\left(E_2\right) 
		=  \prod_{i,j}\Pi\left(\sup_{t \geq 2}|\tilde X_{ijt}-\tilde X^{*}_{ijt}|\leq {\epsilon_0}\right)\prod_{i,j}\Pi\left(|X_{ij1}-X^{*}_{ij1}|\leq {\epsilon_0}\right),
		\end{aligned}
		\end{equation*}
		where $\tilde X_{ijt} = X_{ijt}-X_{ij1}$ for all $i,j,t$.
		
		Given $i,j$, $\xi_{ijt} \sim \mathcal{N}(0, \tau^2)$ for $t \geq 2$, we can denote $\tilde X_{ijt}=\sum_{s=1}^t \xi_{ijs}$ and $(\tilde X_{ij2},...,\tilde X_{ijt})' \sim \mathcal{N}(\+0,\+\Sigma_0)$. Denote $ \tilde{\+x}_{ij}^{*'}=(\tilde X^*_{ij2},...,\tilde X^*_{ijt})'$. Based on multivariate Gaussian concentration through Anderson’s
		inequality, we have 
		\begin{equation}\label{eq:E1}
		\begin{aligned}
		\Pi(E_1  ) &\geq \prod_{i,j}P\left(\sup_{t \geq 2}|\tilde X_{ijt}-\tilde X^{*}_{ijt}|\leq {\epsilon_0} \right) \\
		&\geq \prod_{i,j} \exp(-\frac{\tilde {\+x}_{ij}^{*'}\+\Sigma_0^{-1} \tilde {\+x}_{ij}^* }{2}) \Pi\left(\sup_{t}|\tilde X_{ijt}|\leq {\epsilon_0}  \right).\\
		\end{aligned}
		\end{equation}
		
		By the definition of $\+\Sigma_0$, we have
		$$
		-\frac{\tilde{\+x}_{ij}^{*'}\+\Sigma_0^{-1} \tilde{\+x_{ij}}^* }{2} = - \sum_{t=2}^T \frac{(\tilde X^*_{ijt}-\tilde X^*_{ij(t-1)})^2}{2\tau^2 } = - \sum_{t=2}^T \frac{( X^*_{ijt}- X^*_{ij(t-1)})^2}{2\tau^2 },
		$$
		where $\tilde X^*_{ij1} = 0$.
		For the second factor in Equation~\eqref{eq:E1} , given $i,j$, we consider a Gaussian process $\{\tilde X_{ij}(s),0\leq s\leq 1\}$ induced by $(\tilde X_{ij2},...,\tilde X_{ijt})$ such that $\tilde X_{ij}((s-2)/(T-2))=\tilde X_{ijt}$, $\tilde X_{ij}(0)=0$ and all other values are obtained through interpolations: $\tilde X_{ij}(s)=w_0 \tilde X_{ij{(t-1)}}+(1-w_0) \tilde X_{ij{t}}$  $\forall w_0 \in (0,1)$ with $s=w_0 (t-3)/(T-2)+(1-w_0)(t-2)/(T-2)$. Then clearly, we have 
		\begin{equation*}
		\Pi\left(\sup_{t =2,...,T}|\tilde X_{ijt}|\leq \delta   \right) \geq \Pi\left(\sup_{s \in [0,1]}|\tilde X_{ij}(s)|\leq \delta \right),
		\end{equation*}
		for any $\delta>0$. Denote $\sigma^2(h)=E(\tilde X(s+h)- \tilde X(s))^2 =  h T \tau^2$. Then $\sigma^2(h)$ is linear in $h$ hence concave. In addition, $\sigma(h)/(h^{1/2})= \sqrt{T \tau^2}$, which is non-decreasing in $(0,1)$. Based on Lemma~\ref{lem:normal}, we have
		\begin{equation*}
		\Pi(\sup_{0\leq s \leq 1}|\tilde X(s)| \leq  \delta  ) \geq C_4\exp(-C_3\frac{ T\tau^2}{\delta^2}  )
		\end{equation*}
		for $\delta>0$ with constants $C_3,C_4>0$.
		
		Therefore, for some constant $C_3,C_4>0$, we have
		\begin{equation} \label{eq:rate}
		\begin{aligned}
		\Pi(E_1 ) \geq  C_4 \exp{\left[-\sum_{t=2}^T\frac{\|\+X^*_{t}-\+X^*_{t-1}\|_F^2}{2 \tau^{2} }-C_3\frac{nT\tau^2 }{\epsilon^2}\right]} \\
		\geq  C_4 \exp{\left[-n\sum_{t=2}^T\max_{i}\frac{\|\+x^*_{it}-\+x^*_{i(t-1)}\|_2^2}{2 \tau^{2} }-C_3\frac{nT\tau^2 }{\epsilon^2}\right]}.
		\end{aligned}
		\end{equation}
		For PWD, with condition~\eqref{eq:PWD_condition}, we have
		\begin{equation*} 
		\begin{aligned}
		\Pi(E_1 ) \geq  C_4 \exp{\left[- \frac{-C_0^2 L^2}{2 nT\tau^{2} }-C_3\frac{Tn \tau^2  }{\epsilon^2}\right]}.
		\end{aligned}
		\end{equation*}
		
		Moreover, by taking that $$\tau^2 = \epsilon L/(nT)$$ we can obtain
		\begin{equation*}
		\begin{aligned}
		\log \Pi(E_1) \gtrsim -\frac{L}{\epsilon }   .
		\end{aligned}
		\end{equation*}
		
		For the initial error concentration $\Pi(E_2)$, by the mean-zero Gaussian of $X_{ij1}$ for all $i,j$, we have the concentration:
		\begin{align*}
		\Pi(E_2) = \prod_{i,j}\Pi\left(|X_{ij1}-X^{*}_{ij1}|\leq {\epsilon_0}\right) &\geq \frac{1}{(\sqrt{2\pi}\sigma_0)^{nd}}\exp(-\sum_{i,j} \frac{X^{*2}_{ij1}}{2\sigma_0^2}) (2\epsilon)^{nd} \\
		& \gtrsim \exp \left[-\frac{\|\+X^*_1\|_F^2}{2 \sigma_0^2}-nd-nd \log( \frac{1}{\epsilon}) \right].
		\end{align*}
		Note that $\sigma$ is a constant and $\|\+X^*_1\|_F^2 = O(n)$. We have $\log \Pi(E_2) \gtrsim -n \log( {1/\epsilon}) $. 
		
		Then the rate $\epsilon_{n,T} = L^{1/3}T^{-1/3}n^{-2/3}+\sqrt{\log(nT)/nT}$ can be obtained by letting the smallest possible $\epsilon_{n,T}$ such that $n(n-1)T\epsilon_{n,T}^2 \gtrsim  \max \{L/\epsilon_{n,T}, n \log( {1/\epsilon_{n,T}})\}$.
		
		Finally, this additive rate helps in the choice of the transition $\tau$. In particular, when $L<\log^{3/2}(nT)\sqrt{n/T}$ such that $L^{1/3}T^{-1/3}n^{-2/3} \lesssim \sqrt{\log(nT)/nT}$, the choice of $\tau$ can be relaxed as long as $-\log \Pi(E_1) \lesssim n\log(nT)$.  Therefore, let $\tau^2 = \log^2(nT)/(nT^2)$ in this case, we have
		\begin{align}
		Tn\tau^2/\epsilon_{n,T}^2 \lesssim n\log(nT), \quad \mbox{and}\quad L^2/(nT\tau^2)= n\log(nT). 
		\end{align}
		Therefore, the final choice of $\tau$ that guarantees the optimal convergence rate satisfies $\tau^2 = \log^2(nT)/(nT^2)+\epsilon_{n,T} L/(nT)$. 
		
		By Theorem 3.1 in \cite{bhattacharya2019bayesian}, the prior concentration $\Pi(B_{n,T}(\m X^*;\epsilon_{n,T})) \geq \exp(-Tn (n-1)\epsilon_{n,T}^2)$ implies that the posterior contraction of the averaged $\alpha$-divergence for any $0<\alpha<1$ is at the rate $\epsilon^2_{n,T}$. For the Gaussian case, by the direct calculation \citep{gil2013renyi}, we can obtain that the $\alpha$-divergence is lower bounded by the squared loss function up to some constant factor when the variance of the likelihood is fixed.  For binary case, based on the boundness of the truth and Lemma~\ref{lem:lower of divergence}, which indicates that the $1/2$ divergence is lower bounded by the squared loss function up to some constant factor, we can achieve the results in equation \eqref{result:NWD}.

	\end{proof}
	
	\subsection{Proof of Theorem~\ref{cor:NWD2}}\label{sec:proof_2b}
	\begin{proof}
		Let $\sigma_0^{*2}=1$  and $\tau^{*2}= \epsilon_{n,T}L/(nT)+\log^2(nT)/(nT^2)$. In the proof of Theorem~\ref{cor:NWD}, we show the prior concentration conditional on $\sigma_0=c_1\sigma_0^{*}$ and $\tau=c_2\tau^{*}$ for any constants $c_1,c_2>0$ is sufficient:
		\begin{equation*}
		- \log \{\Pi(B_{n,T}(\m X^*;\epsilon_{n,T}) \mid c_1\sigma_0^{*}, c_2\tau^*)\} \lesssim Tn(n-1)\epsilon_{n,T}^2. 
		\end{equation*}
		Therefore, by limiting on the subset $N(\sigma_0^*,\tau^*)=\{|\sigma_0^2-\sigma_0^{2*}|\leq \sigma_0^{2*}/2, \, |\tau^2-\tau^{2*}|\leq \tau^{2*}/2\}$       , we have $-\log \Pi(B_{n,T}(\m X^*;\epsilon_n) \mid \sigma_0, \tau) \lesssim Tn(n-1)\epsilon_{n,T}^2$. Then 
		\begin{align*}
		\int_{N(\sigma_0^*,\tau^*)} \Pi(B_{n,T}(\m X^*;\epsilon_n) \mid \sigma_0, \tau) p(\tau)p(\sigma_0) d\tau\sigma_0 \\
		\gtrsim P(|\tau^2-\tau^{*2}| \leq \tau^{*2}/2)P(|\sigma_0^2-\sigma_0^{*2}| \leq \sigma_0^{*2}/2)\exp(-Tn(n-1)c_0\epsilon_{n,T}^2),
		\end{align*}
		for some constant $c_0>0$.
		For $\sigma_0$, with the Inverse-gamma$(a_{\sigma_0},b_{\sigma_0})$ prior where $a_{\sigma_0},b_{\sigma_0}$ are constants, we have 
		$$P(|\sigma_0^2-\sigma_0^{*2}| \leq \sigma_0^{*2}/2) = P(1/2\leq \sigma_0^2 \leq 3/2),$$
		which is a fixed constant.
		For $\tau$, with the Gamma$(c_{\tau},d_\tau)$ prior where $c_{\tau},d_\tau$ are constants, we have
		\begin{equation*}
		\begin{aligned}
		P(|\tau^2-\tau^{*2}| \leq \tau^{*2}/2) = \int_{\epsilon_{n,T}L/(2nT)+\log^2(nT)/(2nT^2)}^{3\epsilon_{n,T}L/(2nT)+3\log^2(nT)/(2nT^2)}  f_{c_\tau,d_\tau}(\tau^2) d\tau^2 \\
		\geq  \min_{|\tau^2-\tau^{*2}| \leq \tau^{*2}/2} f_{c_\tau,d_\tau}(\tau^2)  \left\{\epsilon_{n,T}L/(2nT)+\log^2(nT)/(2nT^2)\right\},
		\end{aligned}
		\end{equation*}
		where $f_{c_\tau,d_\tau}(\tau^2)$ is the density function of Gamma$(c_\tau,d_\tau)$ prior. When $|\tau^2-\tau^{*2}| \leq \tau^{*2}/2$, we have
		\begin{align*}
		-\log \{\min_{|\tau^2-\tau^{*2}| \leq \tau^{*2}/2} f_{c_\tau,d_\tau}(\tau^2)\} \lesssim  \tau^{*2}- \log(\tau^{*2}) \lesssim \epsilon_{n,T}L/(nT)+\log^2(nT)/(4nT^2)+\log(nT).
		\end{align*} 
		Note that $\epsilon_{n,T}=o(1)$ due to $L=o(n^2T)$, we have
		$$\epsilon_{n,T}L/(nT) \lesssim  L/(nT) \lesssim n \lesssim n\log(nT) \lesssim n(n-1)T\epsilon_{n,T}^2. $$ In addition, $\log^2(nT)/(4nT^2) +\log(nT)\lesssim n(n-1)T\epsilon_{n,T}^2$ holds. 
		
		Moreover, we also have 
		$-\log(\epsilon_{n,T}L/(nT)+\log^2(nT)/(2nT^2)) \lesssim \log(nT) \lesssim n(n-1)T\epsilon_{n,T}^2$. Therefore, we showed that under the prior $\Pi$, it holds that $-\log \Pi(|\tau^2-\tau^{*2}|\leq \tau^{*2}/2) \lesssim Tn(n-1)\epsilon_{n,T}^2 $. Hence we have
		\begin{equation*}
		\Pi(B_{n,T}(\m X^*;\epsilon_n)) \geq  \exp(-Tn(n-1)M\epsilon_{n,T}^2)
		\end{equation*}
		for large enough constant $M>0$. With the choice $\epsilon_{n,T}^{new}=\sqrt{M}\epsilon_{n,T}$, we showed that the prior concentration is sufficient enough, and the rest of the proof is similar with Theorem~\ref{cor:NWD} by applying Theorem 3.1 in \cite{bhattacharya2019bayesian}. 
	\end{proof}

\subsection{Proof of Proposition~\ref{prop:message_updating}}
	Suppose $q(\beta)$, $q(\tau)$, $q(\sigma_0)$ and $q(\+x_{j\cdot})\,,j\ne i$ are given. By the definition of ELBO  and Equation~\eqref{eq:lsm}, we have
	\begin{align*}
	\mbox{ELBO} &=   \sum_{t=1}^{T-1}\int q_{it,i(t+1)}(\+x_{it},\+x_{i(t+1)})\log\psi_{it,i(t+1)}(\+x_{it},\+x_{i(t+1)})  \rd \m X  \\
	&+\int  q_{it}(\+x_{it})\log\phi_{it}(\+x_{it})  \rd \m X - \sum_{t=1}^T\int  q_{it}(\+x_{it})\log q_{it}(\+x_{it})  \rd \m X  \\
	&- \sum_{t=1}^{T-1} \int q_{it,i(t+1)}(\+x_{it},\+x_{i(t+1)})\{  \log q_{it,i(t+1)}(\+x_{it},\+x_{i(t+1)})\\&-\log q_{it}(\+x_{it}) -\log q_{i(t+1)}(\+x_{i(t+1)}) \}\rd \m X + \mbox{other term.}
	\end{align*}
	By introducing Lagrange multiplier $\lambda_{it,i(t+1)}(\+x_{i(t+1)})$ and $\lambda_{it,i(t-1)}(\+x_{i(t-1)})$ for the marginalization conditions, for the term related with $q_{it,i(t+1)}(\+x_{it},\+x_{i(t+1)})$, we have:
	\begin{align*}
	\log\psi_{it,i(t+1)}(\+x_{it},\+x_{i(t+1)}) -  \log \frac{q_{it,i(t+1)}(\+x_{it},\+x_{i(t+1)})}{q_{it}(\+x_{it})q_{i(t+1)}(\+x_{i(t+1)})} - \lambda_{it,i(t+1)}(\+x_{i(t+1)}) -\lambda_{i(t+1),it}(\+x_{it}) + \mbox{constant}=0.
	\end{align*}
	For the term related to $q_{it}(\+x_{it})$, we have:
	\begin{align*} 
	\log\phi_{it}(\+x_{it})- \log q_{it}(\+x_{it}) 
	+\lambda_{i(t+1),it}(\+x_{it}) +\lambda_{i(t-1),it}(\+x_{it})+  \mbox{constant}=0.
	\end{align*}  
	Then by combining the above result, we have:
	
	\begin{equation}\label{eq34}
	q_{it}(\+x_{it}) \propto \phi_{it}(\+x_{it})  \exp(\lambda_{i(t+1),it}(\+x_{it}) +\lambda_{i(t-1),it}(\+x_{it})).
	\end{equation} 
	Moreover, we have
	\begin{align}\label{eq35}
	\begin{aligned}
	q_{it,i(t+1)}(\+x_{it},\+x_{i(t+1)}) \propto \phi_{it}(\+x_{it}) \phi_{i(t+1)}(\+x_{i(t+1)}) \psi_{it,i(t+1)}(\+x_{it},\+x_{i(t+1)}) \\
	\cdot \exp(\lambda_{i(t+2),i(t+1)}(\+x_{i(t+1)}) +\lambda_{i(t-1),it}(\+x_{it})).
	\end{aligned}
	\end{align}
	Finally, based on the marginalization property of $q_{it,i(t+1)}(\+x_{it},\+x_{i(t+1)}) $, we have the backward updating:
	\begin{align*}
	\exp(\lambda_{i(t+1),it}(\+x_{it}) ) \propto \int \phi_{i(t+1)}(\+x_{i(t+1)}) \psi_{it,i(t+1)}(\+x_{it},\+x_{i(t+1)}) \exp(\lambda_{i(t+2),i(t+1)}(\+x_{i(t+1)}))  \rd \+x_{i(t+1)}
	\end{align*}
	and forward updating:
	\begin{align*}
	\exp(\lambda_{it,i(t+1)}(\+x_{i(t+1)}) ) \propto \int \phi_{it}(\+x_{it}) \psi_{it,i(t+1)}(\+x_{it},\+x_{i(t+1)}) \exp(\lambda_{i(t-1),it}(\+x_{it})) 
	\rd \+x_{it}.
	\end{align*}
	Let $m_{it,i(t+1)}(\+x_{i(t+1)}) = \exp(\lambda_{it,i(t+1)}(\+x_{i(t+1)}) )$ and  $m_{it,i(t-1)}(\+x_{i(t-1)}) = \exp(\lambda_{it,i(t-1)}(\+x_{i(t-1)}) )$.     The Equation~\eqref{eq:7} directly follows Equation~\eqref{eq34} and~\eqref{eq35} after plugging in the forward and backward messages and therefore the proposition is proved.

	\subsection{Proof of Theorem~\ref{thm:varia}}
	\begin{proof}
		The proof is based on Theorem 3.3 in \cite{yang2020alpha}, where we need to provide upper bounds for $$-\int \log \frac{P(\m Y\mid \m X)}{P(\m Y\mid \m X^*)}q(\m X) d\m X$$ and $$D_{KL}(q(\m X) \,||\, p(\m X)),  $$
		where $q(\m X)$ is a variational distribution in the SMF family and $p(\m X)$ is the prior.
		Based on the definition of $E_0$ in the proof in subsection~\ref{proof:thm2a}, we have
		$$ B_{n,T}(\m X^*;\epsilon) \supset E_0: =\{\max_{i,t} \|\+x_{it}-\+x_{it}^*\|_2 \leq  \epsilon_0\}, $$  with $ \epsilon_0= c_1\epsilon_{n,T}$, for constant $c_1>0$. The above constraint can be written in a separate form:
		$$E_0 = \cap_{i,t}  \{ \|\+x_{it}-\+x_{it}^*\|_2\leq \epsilon_0\}.  $$
		Then we can choose $q(\m X)$ in the following way:
		\begin{equation*}
		q(\m X) \propto \prod_{i=1}^n \prod_{t=2}^T p(\+x_{it} \mid \+x_{i(t-1)}) \ind\{\|\+x_{it}-\+x_{it}^*\|_2\leq \epsilon_0\}\prod_{i=1}^n  p(\+x_{i1} ) \ind\{\|\+x_{i1}-\+x_{i1}^*\|_2\leq \epsilon_0\},
		\end{equation*}
		where $p(\+x_{it} \mid \+x_{i(t-1})$ and $p(\+x_{i1} )$ are components of priors. Note that the above variational distribution belongs to the SMF distribution family. We prove the above two bounds based on the current construction of $q(\m X)$.
		First, by Fubini’s theorem and the definition of the prior, we have
		$$
		\begin{aligned}
		& \E_{\m X^*}\left[-\int_{\m X} q(\m X) \log \frac{P(\m Y \mid  \m X)}{P\left(\m Y \mid \m X^{*}\right)} d\m X\right] \\
		&=\int_{\m X} -\E_{\m X^*}\left[\log \frac{P(\m Y \mid \m X)}{P\left(\m Y \mid \m X^*\right)}\right] q(\m X) d \m X \\
		&\leq \int_{B_{n}\left(\m X^*, \epsilon\right)} D_{KL}\left[P\left(\m Y \mid \m X^*\right) || P\left(\m Y \mid \m X\right) \right] q(\m X) d \m X \leq n(n-1)T \epsilon^{2}.
		\end{aligned}
		$$
		Similarly, for the variance, by  Jensen's inequality and Fubini’s theorem, we have
		$$
		\begin{aligned}
		& \mbox{Var}_{\m X^*}\left[\int_{\m X} q(\m X) \log \frac{P(\m Y \mid  \m X)}{P\left(\m Y \mid \m X^{*}\right)} d\m X\right] \\
		&\leq \E_{\m X^*}\left[\int_{\m X} q(\m X) \log \frac{P(\m Y \mid  \m X)}{P\left(\m Y \mid \m X^{*}\right)} d\m X\right]^2\\
		&\leq \int_{B_{n}\left(\m X^*, \epsilon\right)} V_2\left[P\left(\m Y \mid \m X^*\right) || P\left(\m Y \mid \m X\right) \right] q(\m X) d \m X \leq n(n-1)T \epsilon^{2}.
		\end{aligned}
		$$
		Therefore, by Chebyshev’s inequality, for any $D>1$, based on the first and second moments of the above bounds, we have
		$$
		\begin{aligned}
		& P_{\m X^*}\left[\int_{\m X} q(\m X) \log \frac{P(\m Y \mid  \m X)}{P\left(\m Y \mid \m X^{*}\right)} d\m X \leq -Dn(n-1)T\epsilon^2\right] \\
		&\leq P_{\m X^*}\left[\int_{\m X} q(\m X) \log \frac{P(\m Y \mid  \m X)}{P\left(\m Y \mid \m X^{*}\right)} d\m X\right.\\
		&\left.-\E\left\{  \int_{\m X} q(\m X) \log \frac{P(\m Y \mid  \m X)}{P\left(\m Y \mid \m X^{*}\right)} d\m X \right\} \leq -(D-1)n(n-1)T\epsilon^2\right] \\
		&\leq \mbox{Var}_{\m X^*}\left[\int_{\m X} q(\m X) \log \frac{P(\m Y \mid  \m X)}{P\left(\m Y \mid \m X^{*}\right)} d\m X\right] /\left( (D-1)^2 n^2(n-1)^2T^2 \epsilon^4 \right) \\
		& \leq \frac{4}{(D-1)^2 n(n-1)T\epsilon^2}
		\end{aligned}
		$$
		holds with probability $1-1/\{(D-1)^2 n(n-1)T \epsilon^2\}$.
	
	This proves that when $n(n-1)T \epsilon \rightarrow \infty$, we have 
	$$-\int \log \frac{P(\m Y\mid \m X)}{P(\m Y\mid \m X^*)}q(\m X) d\m X \leq Dn(n-1)T\epsilon^2$$ with probability converging to one.
	
	In addition, based on the construction of the variational family, we have
	\begin{equation*}
	D_{KL}(q(\m X) || p(\m X)) = - \log(\Pi(E_0)),
	\end{equation*}
	since for any probability measure $\mu$ and measurable set $A$ with $\mu(A)>0$, we have $D_{KL}(\mu(\cdot \,\cap A)/\mu(A) \,||\, \mu) = -\log (\mu(A))$.  
	By the proof in subsection~\ref{proof:thm2a}, we have $- \log(\Pi(E_0)) \lesssim -\log(\Pi(E_1 \cap E_2))  \lesssim \max \{L/\epsilon, n \log( {1/\epsilon})\}$ for PWD($L$) with Lipschitz condition. 
	Therefore, the convergence of the $\alpha$-divergence follows by Theorem 3.3 in \cite{yang2020alpha}. Finally, the $\alpha$-divergence is lower bounded by the loss according to the final part of the proof of Theorem~\ref{cor:NWD}.
	
	\end{proof}
	
	\subsection{Proof of Theorem~\ref{thm:varia2} }\label{sec:proof_3b}
	
	\begin{proof}
		Note that the prior now satisfies $p(\m X,\tau,\sigma_0) = p(\m X \mid \tau, \sigma_0) p(\tau)p(\sigma_0)$ and the variational distribution instead satisfies $q(\m X,\tau,\sigma_0) = \prod_{i=1}^n q_i(\+x_{i\cdot}) q(\tau)q(\sigma_0)$. Let $\sigma_0^{*2}=1$ and $\tau^{*2}= \epsilon_{n,T}L/(nT)+\log^2(nT)/(nT^2)$, we consider the following variational distribution:
		\begin{align}\label{eq:variational_candidate}
		\begin{aligned}
		q(\m X,\tau,\sigma_0) \propto \prod_{i=1}^n \prod_{t=2}^T p(\+x_{it} \mid \+x_{i(t-1)},\tau^*) \ind\{\|\+x_{it}-\+x_{it}^*\|_2\leq c_1\epsilon_{n,T}\} \\
		\times\prod_{i=1}^n p(\+x_{i1} \mid \sigma_0^*) \ind\{\|\+x_{i1}-\+x_{i1}^*\|_2\leq c_1 \epsilon_{n,T}\} \\
		\times p(\tau) \ind\{ \tau^{*2}<\tau^2<\tau^{*2}e^{\epsilon_{n,T}^2}\}p(\sigma_0)\ind\{ \sigma_0^{*2}<\sigma_0^2<\sigma_0^{*2}e^{\epsilon_{n,T}^2}\},
		\end{aligned}
		\end{align}
		where $c_1$ is the constant used in the proof of Theorem~\ref{thm:varia}.
		Given the prior, we still check the conditions 
		\begin{equation}\label{eq:cond_1}
		-\int \log \frac{P(\m Y\mid \m X)}{P(\m Y\mid \m X^*)}q(\m X,\tau,\sigma_0) d\m X d\tau d\sigma_0 \lesssim  Tn(n-1)\epsilon_{n,T}^2
		\end{equation}
		\begin{equation}\label{eq:cond_2}
		D_{KL}(q(\m X,\tau,\sigma_0) \,||\, p(\m X,\tau,\sigma_0)) \lesssim Tn(n-1)\epsilon_{n,T}^2
		\end{equation}
		First, the condition~\eqref{eq:cond_1} directly follows the proof of Theorem~\ref{thm:varia} given the MF structure $q(\m X,\tau,\sigma_0)=q(\m X)q(\tau)q(\sigma_0)$. 
		
		Then  by the chain rule of KL divergence, we have
		\begin{align}\label{eq:KL_chain}
		\begin{aligned}
		D_{KL}(q(\m X,\tau,\sigma_0) \,||\, p(\m X,\tau,\sigma_0)) = D_{KL}(q(\tau) || p(\tau)) + D_{KL}(q(\sigma_0) || p(\sigma_0))\\
		+ \int q(\tau)q(\sigma_0) \int q(\m X) \log \frac{q(\m X)}{p(\m X |\tau, \sigma_0)} d\m X d\tau d\sigma_0.
		\end{aligned}
		\end{align}
		With the Gamma$(c_{\tau},d_\tau)$ prior and $\epsilon_{n,T}<1$, we have
		\begin{align}\label{eq:A17}
		\begin{aligned}
		D_{KL}(q(\tau) || p(\tau))&= - \log(P(\tau^{*2}<\tau^2<\tau^{*2}e^{\epsilon_{n,T}^2})) \\
		&\leq -\log( \min_{\tau^{*2}<\tau^2<\tau^{*2}e^{\epsilon_{n,T}^2}} f_{c_\tau,d_\tau}(\tau^2) (e^{\epsilon_{n,T}^2}-1)) \\
		&\stackrel{(i)}{\leq} -\log(\epsilon_{n,T}^{2}) -\log( \min_{\tau^{*2}<\tau^2<\tau^{*2}e^{\epsilon_{n,T}^2}} f_{c_\tau,d_\tau}(\tau^2)) \\
		&\stackrel{(ii)}{\lesssim}   Tn(n-1)\epsilon_{n,T}^2-\log( \min_{\tau^{*2}<\tau^2<\tau^{*2}e^{\epsilon_{n,T}^2}} f_{c_\tau,d_\tau}(\tau^2)),
		\end{aligned}
		\end{align}
		where in $(i)$ we use $e^{x}-1 \geq x$ for any $x$ and $(ii)$ is because $\epsilon_{n,T}^2\geq \log(nT)/(nT)$. In addition, by a similar approach with proof in Theorem~\ref{cor:NWD2}, we have
		\begin{align*}
		-\log (\min_{\tau^{*2}<\tau^2<\tau^{*2}e^{\epsilon_{n,T}^2}} f_{c_\tau,d_\tau}(\tau^2))\lesssim  \tau^{*2}- \log(\tau^{*2}) \lesssim n(n-1)T\epsilon_{n,T}^2.
		\end{align*}

		With $\epsilon_{n,T}<1$, we have $1<\sigma_0^2<e$ in the constrained region, where the density of Inverse-Gamma$(a_{\sigma_0},b_{\sigma_0})$ is lower bounded by a constant. Hence,
		\begin{align}\label{eq:A18}
		D_{KL}(q(\sigma_0) || p(\sigma_0))= - \log(P(\sigma_0^{*2}<\sigma_0^2<\sigma_0^{*2}e^{\epsilon_{n,T}^2})) \lesssim -\log(\epsilon_{n,T}^2)  \stackrel{(i)}{\lesssim} Tn(n-1)\epsilon_{n,T}^2,
		\end{align}
		where $(i)$ is due to $\epsilon_{n,T}^2 \geq \log(nT)/(nT)$.
		For the third term of the KL divergence, we have
		\begin{align*}
		\int q(\m X) \log \frac{q(\m X)}{p(\m X |\tau, \sigma_0)} d\m X = \int_{E_0} q(\m X) \log \frac{p(\m X |\tau^*, \sigma^*_0)}{p(\m X |\tau, \sigma_0)}d \m X -\log(\Pi(E_0 \mid \tau^{*},\sigma_0^{*})).
		\end{align*}
		Note that we already have $-\log(\Pi(E_0 \mid \tau^{*},\sigma_0^{*})) \lesssim Tn(n-1)\epsilon_{n,T}^2$ by the proof of the prior concentration in subsection~\ref{proof:thm2a}.
		
		Moreover, we have the density,
		\begin{align*}
		p(\m X\mid \tau ,\sigma_0) =\frac{1}{(\sqrt{2\pi})^{nTd}} \exp\left\{-\frac{n(T-1)d}{2}\log(\tau^{2})-\frac{nd}{2}\log(\sigma_0^{2})- \frac{\|\+X_1\|_F^2}{2\sigma_0^2}-\frac{\sum_{t=2}^T\|\+X_{t}-\+X_{t-1}\|_F^2}{2\tau^2} \right\},
		\end{align*}
		which implies that 
		\begin{align*}
		\log \frac{p(\m X |\tau^*, \sigma^*_0)}{p(\m X |\tau, \sigma_0)} = \frac{n(T-1)d}{2}\log(\tau^{2})-\frac{n(T-1)d}{2}\log(\tau^{*2})+\frac{nd}{2}\log(\sigma_0^{2})-\frac{nd}{2}\log(\sigma_0^{*2})\\
		+\frac{\|\+X_1\|_F^2}{2\sigma_0^2}-\frac{\|\+X_1\|_F^2}{2\sigma_0^{*2}}+\frac{\sum_{t=2}^T\|\+X_{t}-\+X_{t-1}\|_F^2}{2\tau^2}-\frac{\sum_{t=2}^T\|\+X_{t}-\+X_{t-1}\|_F^2}{2\tau^{*2}}.
		\end{align*}
		
		With the constrained region $\tau^{*2}<\tau^2<\tau^{*2}e^{\epsilon_{n,T}^2}$ and $\sigma_0^{*2}<\sigma_0^2<\sigma_0^{*2}e^{\epsilon_{n,T}^2}$, we have,
		
		\begin{equation*}
		\log \frac{p(\m X |\tau^*, \sigma^*_0)}{p(\m X |\tau, \sigma_0)} \leq \frac{n(T-1)d}{2}\epsilon_{n,T}^2+\frac{nd}{2}\epsilon_{n,T}^2 \lesssim Tn(n-1)\epsilon_{n,T}^2,
		\end{equation*}
		which implies that the third term of the KL divergence~\eqref{eq:KL_chain} is also bounded by $Tn(n-1)\epsilon_{n,T}^2$. Therefore, we proved that condition~\eqref{eq:cond_2} is satisfied. 
		
		Finally, the conclusion holds by applying similar arguments in the final part of the proof of Theorem~\ref{thm:varia}.
		
	\end{proof}

	\subsection{Nodewise Adaptive Priors}\label{sec:node}
	In this section, we consider the likelihood~\eqref{eq:latent_space} with nodewise adaptive priors:
	\begin{align}\label{eq:prior_nodewise_app}
	\begin{aligned}
	\+x_{i1} &\sim \m N(\+0,\sigma_{0i}^2 \mb I_d), \quad &\+x_{i(t+1)}\mid \+x_{it} \sim \m N( \+x_{it},\tau_i^2 \mb I_d),   \\
	\sigma_{0i}^2 &\sim \mbox{Inverse-Gamma}\left(a_{\sigma_0},b_{\sigma_0}\right), \quad &\tau_i^2 \sim \mbox{Gamma}\left(c_\tau,d_\tau\right),
	\end{aligned}
	\end{align}
	for $i=1,...,n; t=1,...,T-1$ to capture the nodewise level differences. The SMF are now in the following form:
	\begin{equation}\label{eq:block_mf_node}
	q(\m X,\+\tau,\+\sigma_0,\beta) = \prod_{i=1}^{n}q_i(\+x_{i\cdot})q(\tau_i)q(\sigma_{0i})q(\beta).
	\end{equation}
	First, there are only minimal changes in the computational framework. First, for the $q_i(\+x_{i\cdot})$ updatings, we have the graph potentials $\+x_{i\cdot}$ as follows:
	\begin{align*}
	\begin{aligned}
	&\phi_{i1}(\+x_{i1}) = \exp\{-\mu_{1/\tau_i^2}\|\+x_{i1}\|_2^2/2-\mu_{1/\sigma_{0i}^2}\|\+x_{i1}\|_2^2/2\}\prod_{j\ne i}\exp [\E_{q(\beta)q(\+x_{j1})} \{ \log P_\alpha(Y_{ij1}\mid \+x_{i1},\+x_{j1},\beta)\} ], \\
	&\phi_{it}(\+x_{it}) =\exp\{-\mu_{1/\tau_i^2}\|\+x_{it}\|_2^2/2\}\prod_{j\ne i}\exp [\E_{q(\beta)q(\+x_{jt})} \{ \log P_\alpha(Y_{ij1}\mid \+x_{it},\+x_{jt},\beta)\} ],   \,\forall t\in \{2,...,T\}\\
	&\psi_{it,i(t+1)}(\+x_{it},\+x_{i(t+1)}) =\exp(\mu_{1/\tau_i^2} \+x_{i(t+1)}' \+x_{it}), \,  \forall t\in \{1,...,T-1\},
	\end{aligned}
	\end{align*}
	where $\mu_{1/\tau_i^2} = \E_{q(\tau_i)}(1/\tau_i^2)$ and $\mu_{1/\sigma_{0i}^2} = \E_{q(\sigma_{0i})}(1/\sigma_{0i}^2)$.
	Then the updating of $q_i(\+x_{i\cdot})$ follows the same MP framework under the above revised potentials. In addition, for the updating of scales, we have
	
	\begin{align}\label{eq:scale_update_node}
	\begin{aligned}
	q^{(new)}(\tau_i^2) &\propto \exp\left[\E_{q_i(\+x_{i\cdot})} \left\{-\sum_{t=2}^{T} \frac{\|\+x_{it}-\+x_{i(t-1)}\|_2^2}{2\tau_i^2}\right\}-\frac{(T-1)d+c_\tau-1}{2} \log(\tau_i^2)-d_\tau\tau_i^2\right]. \\
	q^{(new)}(\sigma_{0i}^2) &\propto \exp\left[\E_{q(\+x_{i1})} \left(- \frac{\|\+x_{i1}\|_2^2}{2\sigma_{0i}^2}\right)-\left(\frac{d}{2}+a_{\sigma_{0}}+1 \right) \log(\sigma_{0i}^2)-\frac{b_{\sigma_{0}}}{\sigma_{0i}^2}\right].
	\end{aligned}
	\end{align}
	Therefore, we can obtain the that new update of  $q(\tau_i^2)$ follows a Generalized inverse Gaussian distribution with parameter $a=2d_\tau,b=\E_{q_i(\+x_{i\cdot})} \{\sum_{t=2}^{T} \|\+x_{it}-\+x_{i(t-1)}\|_2^2/2\},p=1/2-(T-1)d/2-c_\tau/2$. Then the moment required in updating $ \+x_{it}$  can be obtained:  $\E_{q(\tau_i)}(1/\tau_i^2)=K_{p+1}(\sqrt{b})/\left\{\sqrt{b} K_p(\sqrt{b})\right\}-2p/b$, where $K_p(\cdot)$ is the modified Bessel function of the second kind.  In addition, the new update of $\sigma_{0i}^{(new)2}\sim$ Inverse-Gamma$((d+a_{\sigma_0})/2,\{\E_{q(\+x_{i1})}(\|\+x_{i1}\|_2^2)+2b_{\sigma_{0i}}\}/2)$, which implies $\mu_{1/\sigma_{0i}^2} = \E_{q(\sigma_{0i})}(1/\sigma_{0i}^2)= (d+a_{\sigma_0})/\{\E_{q( \+x_{i1})}(\|\+x_{i1}\|_2^2)+2b_{\sigma_0}\}$.

	 To capture a smooth evolution of the latent coordinates over time for each node, we assume the following parameter space for the latent positions:
	\begin{equation}\label{truth:PAWD}
	\mbox{PWAD}(\+L) :=\left\{\m X: \|\+x_{it}-\+x_{i(t-1)}\|_2 \leq \frac{L_i}{nT}, \, \+L=[L_1,...,L_n], L:=\|\+L\|_\infty\right\},
	\end{equation}
	where PWAD denotes point-wise adaptive dependence.
	The theoretical results can also be obtained similarly:

     	\begin{theorem}[Fractional posterior convergence rate for nodewise adaptive priors]\label{cor:NWD_node}
		Suppose the true data generating process satisfies Equation~\eqref{eq:dynamic}, $\m X^* \in \mbox{PAWD}(\+L)$  with $0 \leq  L=o(Tn^2)$ and condition~\eqref{bounded} holds. Suppose $d$ is a known fixed constant. Let $\epsilon_{n,T}=\|\+L\|_2^{1/3}/(T^{1/3}n^{1/2})+\sqrt{\log(nT)/(nT)} $. Then if we apply the priors defined in Equation~\eqref{eq:prior} and adopt priors~\eqref{eq:prior_nodewise_app} for $\sigma_{0i}$ and $\tau_i$,
		we have for $n,T \rightarrow \infty$,
		\begin{equation}\label{result:NWD_node}
		\E\left[\Pi_{\alpha}\left\{ \frac{1}{Tn(n-1)} \sum_{t=1}^{T}\sum_{i \ne j=1}^n \left( \hat{\+x}'_{it}\hat{\+x}_{jt} -\+x_{it}^{*'}\+x_{jt}^{*} \right)^2 \geq M\epsilon^2_{n,T} \mid \m Y\right\} \right] \rightarrow 0,
		\end{equation}
		with $P_{\m X^*}$ probability converging to one, where $M>0$ is a large enough constant.
	\end{theorem}
	
	\begin{proof}
		The proof is similar to the proof of Theorem~\ref{cor:NWD2} in Section~\ref{sec:proof_2b}. It suffices to show that the prior concentration for the set $N(\sigma_0^*,\tau^*)=\{|\sigma_{0i}^2-\sigma_0^{*2}|\leq \sigma_0^{*2}/2, \, |\tau_{i}^2-\tau^{*2}|\leq \tau^{*2}/2,\, i=1,..,n\}$ is sufficiently large. Due to the independence of the prior, we have 
		\begin{equation*}
		-\log	P(|\sigma_{0i}^2-\sigma_0^{*2}|\leq \sigma_0^{*2}/2,i=1,...,n) = -\log \{P(|\sigma_{01}^2-\sigma_0^{*2}|)^n\} \lesssim n \lesssim n(n-1)T \epsilon_{n,T}^2.
		\end{equation*}
		Similarly,
		\begin{equation*}
		\begin{aligned}
		-\log	P(|\tau_{i}^2-\tau^{*2}|\leq \tau^{*2}/2,i=1,...,n) = -n\log \{P(|\tau_{1}^2-\tau^{*2}|\leq \tau^{*2}/2)\} \\
		\lesssim -n\log(\epsilon_{n,T}L/(nT)+\log^2(nT)/(2nT^2)) +n  \tau^{*2}- n\log(\tau^{*2}).
		\end{aligned}
		\end{equation*}	
		Since $\log^2(nT)/(nT^2) \leq\tau^{*2} \leq L/(nT)+\log^2(nT)/(nT^2)$, we have $ -n\log(\log^2(nT)/(2nT^2))  \lesssim n\log(nT)$; $n  \tau^{*2} \leq L/T+ \log^2(nT)/T^2 \lesssim n\log(nT)$ and $- n\log(\tau^{*2}) \lesssim n\log(nT)$.
		Therefore, we show that $-\log \Pi(N(\sigma_0^*,\tau^*)) \lesssim n(n-1)T \epsilon_{n,T}^2$, then the rest of the proof follows the same with Section~\ref{sec:proof_2b} by the improved bound $\|\+X_t^*-\+X_{t-1}^*\|_F^2 \leq \|\+L\|_2^2/(n^2T^2)$ in equation~\ref{eq:rate}.
		
	\end{proof}

	\begin{theorem}[Variational risk bound for nodewise adaptive SMF]\label{thm:varia_node}
		Suppose the true data generating process satisfies Equation~\eqref{eq:dynamic}, $\m X^* \in \mbox{PAWD}(\+L)$  with $0 \leq  L=o(Tn^2)$ and condition~\eqref{bounded} holds. Suppose $d$ is a known fixed constant. Let $\epsilon_{n,T}=\|\+L\|_2^{1/3}/(T^{1/3}n^{1/2})+\sqrt{\log(nT)/(nT)} $. Then if we apply the priors defined in Equation~\eqref{eq:prior} and adopt  priors~\eqref{eq:prior_nodewise_app} for $\sigma_{oi}$ and $\tau_i$ for $i=1,...,n$  and obtaining the optimal variational distribution $\hat{q}(\m X)$ under nodewise adaptive SMF  family~\eqref{eq:block_mf_node},
		we have with $P_{\m X^*}$ probability tending to one as $n,T \rightarrow \infty$,
		\begin{equation}\label{result:variational_risk_node}
		\E_{\hat{q}(\m X)} \left[ \frac{1}{Tn(n-1)} \sum_{t=1}^{T}\sum_{i \ne j=1}^n \left( \hat{\+x}'_{it}\hat{\+x}_{jt} -\+x_{it}^{*'}\+x_{jt}^{*} \right)^2\right]  \lesssim \epsilon^2_{n,T} .
		\end{equation}
	\end{theorem}
	
	\begin{proof}
		We consider the following variational distribution:
		\begin{align}
		\begin{aligned}
		q(\m X,\tau,\sigma_0) \propto \prod_{i=1}^n \prod_{t=2}^T p(\+x_{it} \mid \+x_{i(t-1)},\tau^*) \ind\{\|\+x_{it}-\+x_{it}^*\|_2\leq c\epsilon_{n,T}\} \\
		\times\prod_{i=1}^n p(\+x_{i1} \mid \sigma_0^*) \ind\{\|\+x_{i1}-\+x_{i1}^*\|_2\leq c\epsilon_{n,T}\} \\
		\times \prod_{i=1}^n p(\tau_i) \ind\{ \tau^{*2}<\tau_i^2<\tau^{*2}e^{\epsilon_{n,T}^2}\} \prod_{i=1}^n p(\sigma_{0i})\ind\{ \sigma_0^{*2}<\sigma_{0i}^2<\sigma_0^{*2}e^{\epsilon_{n,T}^2}\}.
		\end{aligned}
		\end{align}
	\end{proof}
	
	After the change  of the priors and variational family, first by Equation~\eqref{eq:A18} and $-n\log(\epsilon_{n,T}^2) \lesssim n\log(nT)$, we have
	
	\begin{align*}
	D_{KL}(q(\+\sigma_0) || p(\+\sigma_0))= - n\log(P(\sigma_0^{*2}<\sigma_{01}^2<\sigma_0^{*2}e^{\epsilon_{n,T}^2})) \lesssim -n\log(\epsilon_{n,T}^2) {\lesssim} Tn(n-1)\epsilon_{n,T}^2.
	\end{align*}
	Similarly, by Equation~\eqref{eq:A17}, we also have
	
	\begin{align*}
	\begin{aligned}
	D_{KL}(q(\+\tau) || p(\+\tau))&= - n\log(P(\tau^{*2}<\tau_1^2<\tau^{*2}e^{\epsilon_{n,T}^2})) \\
	&\lesssim -n\log(\epsilon_{n,T}^{2}) -n\log( \min_{\tau^{*2}<\tau^2<\tau^{*2}e^{\epsilon_{n,T}^2}} f_{c_\tau,d_\tau}(\tau^2)) \\
	&\lesssim Tn(n-1)\epsilon_{n,T}^2.
	\end{aligned}
	\end{align*}

	Moreover, we have the density,
	\begin{align*}
	p(\m X\mid \+\tau ,\+\sigma_0) =\frac{1}{(\sqrt{2\pi})^{nTd}} \exp\left\{-\sum_{i=1}^{n}\frac{(T-1)d}{2}\log(\tau_i^{2})-\sum_{i=1}^{n}\frac{d}{2}\log(\sigma_{0i}^{2})\right.\\
	\left.	- \sum_{i=1}^{n}\frac{\|\+x_{i1}\|_2^2}{2\sigma_{0i}^2}-\frac{\sum_{i=1}^{n}\sum_{t=2}^T\|\+x_{it}-\+x_{i(t-1)}\|_2^2}{2\tau_i^2} \right\},
	\end{align*}
	which implies that 
	\begin{align*}
	\log \frac{p(\m X |\+\tau^*, \+\sigma^*_0)}{p(\m X |\+\tau, \+\sigma_0)} =\sum_{i=1}^{n}\frac{(T-1)d}{2}\log(\tau_i^{2})-\frac{n(T-1)d}{2}\log(\tau^{*2})+\sum_{i=1}^{n}\frac{d}{2}\log(\sigma_{0i}^{2})-\frac{nd}{2}\log(\sigma_0^{*2})\\
	+\sum_{i=1}^{n}\frac{\|\+x_{i1}\|_2^2}{2\sigma_{0i}^2}-\frac{\|\+X_1\|_F^2}{2\sigma_0^{*2}}+\frac{\sum_{i=1}^{n}\sum_{t=2}^T\|\+x_{it}-\+x_{i(t-1)}\|_2^2}{2\tau_i^2}-\frac{\sum_{t=2}^T\|\+X_{t}-\+X_{t-1}\|_F^2}{2\tau^{*2}},
	\end{align*}
	where $\+\tau^*=(\tau^*,\tau^*,...,\tau^*)'$ and $\+\sigma_{0}^* = (\sigma_{0}^*,\sigma_{0}^*,...,\sigma_{0}^*)'$.
	With the constrained region $\tau^{*2}<\tau^2<\tau^{*2}e^{\epsilon_{n,T}^2}$ and $\sigma_0^{*2}<\sigma_0^2<\sigma_0^{*2}e^{\epsilon_{n,T}^2}$, we have,
	
	\begin{equation*}
	\log \frac{p(\m X |\+\tau^*, \+\sigma^*_0)}{p(\m X |\+\tau, \+\sigma_0)} \leq \frac{n(T-1)d}{2}\epsilon_{n,T}^2+\frac{nd}{2}\epsilon_{n,T}^2 \lesssim Tn(n-1)\epsilon_{n,T}^2.
	\end{equation*}
	Then the rest of the proofs follow the same with proof of Theorem~\eqref{thm:varia} in Section~\ref{sec:proof_3b}.
	
	\subsection{Auxiliary Lemmas}
	
	%
	
	\begin{lemma}[Small ball probability, Theorem 1.1 in \citealp{shao1993note}]\label{lem:normal}       
		Let $\{X(t),0\leq t\leq 1\}$ be a real-valued Gaussian process with mean zero, $X(0)=0$ and stationary increments. Denote $\sigma^2(h)=E(X(t+h)-X(t))^2$ for $0\leq t\leq t+h\leq 1$. If $\sigma^2(h)$ is concave and $\sigma(h)/h^\alpha$ is non-decreasing in $(0,1)$ for some $\alpha>0$, then we have
		\begin{equation*}
		P(\sup_{0\leq t \leq 1}|X(t)| \leq C_\alpha \sigma(x)) \geq \exp(-2/x),
		\end{equation*}
		where $C_\alpha =1+3e\sqrt{\pi/\alpha}$.
	\end{lemma}

	{
	\begin{lemma}[Upper bound for binary KL divergence]\label{lem:binary_KL}
		Let $p_a=1/(1+\exp(-a))$ and $p_b=1/(1+\exp (-b))$. Define $P_a$ and $P_b$ as the Bernoulli measures with probability $p_a$ and $p_b$. Then we have
		\begin{equation*}
		D_{KL}(P_{a} \,||\, P_{b}) + D_{KL}(P_{b} \,||\, P_{a}) \leq (p_a \vee p_b)(a-b)^2.
		\end{equation*}
	\end{lemma}
	}
	\begin{proof}
		\begin{equation*}
		\begin{aligned}
		D_{KL}(P_{a} \,||\, P_{b}) + D_{KL}(P_{b} \,||\, P_{a}) &=(p_a-p_b) \log \frac{p_a}{p_b}+(p_b-p_a) \log \frac{1-p_a}{1-p_b}\\
		=(p_a-p_b) \log\left(\frac{p_a}{1-p_a} \frac{1-p_b}{p_b}\right) 
		&= \left\{\frac{1}{1+\exp(-a)}-\frac{1}{1+\exp(-b)}\right\} \log(e^{a} e^{-b}) \\
		& = (a-b)\left\{\frac{1}{1+\exp(-a)}-\frac{1}{1+\exp(-b)}\right\}.
		\end{aligned}
		\end{equation*}
		Without loss of generality, we can assume $a>b$, then by $\exp(x)\geq 1+x$, we have
		\begin{align*}
		\frac{1}{1+\exp(-a)}-\frac{1}{1+\exp(-b)} = \frac{e^{-b}-e^{-a}}{(1+\exp(-a))(1+\exp(-b))} \\
		\leq  \frac{1-e^{b-a}}{(1+e^{-a})(1+e^b)} \leq p_a(1-e^{b-a}) \leq p_a(a-b).
		\end{align*}
	\end{proof}

{
	\begin{lemma}[Upper bound of second order KL moment]\label{lem:second_KL}
		Let $p_a=1/(1+\exp(-a))$ and $p_b=1/(1+\exp (-b))$. Define $P_a$ and $P_b$ as the Bernoulli measures with probability $p_a$ and $p_b$. Then we have
		\begin{equation*}
		\int P_a \log^2 \left( \frac{P_a}{P_b}\right) d\mu \leq  \left[\frac{p_a}{(p_a \wedge  p_b)^2}+\frac{1-p_a}{(1-p_a\vee p_b)^2}\right] (p_a \vee p_b)^2(a-b)^2.
		\end{equation*}
	\end{lemma}
 }
	
	\begin{proof}
 Note that 
		\begin{equation*}
		\begin{aligned}
\int P_a \log^2 \left( \frac{P_a}{P_b}\right) d\mu = p_a\log^2 \left( \frac{p_a}{p_b}\right) +(1-p_a) \log^2\left( \frac{1-p_a}{1-p_b}\right).
		\end{aligned}
		\end{equation*}
  We have
\begin{align*}
    \log^2 \left( \frac{p_a}{p_b}\right)  = \log^2\left( \frac{p_a \vee p_b}{p_a \wedge p_b}-1+1\right) \leq \left(\frac{p_a \vee p_b - p_a \wedge p_b}{p_a \wedge p_b}\right)^2 = \left(\frac{p_a - p_b}{p_a \wedge p_b}\right)^2.
\end{align*}
  Similarly,
\begin{align*}
    \log^2 \left( \frac{1-p_a}{1-p_b}\right)  = \log^2\left( \frac{(1-p_a) \vee (1-p_b)}{(1-p_a) \wedge (1-p_b)}-1+1\right) \leq \left(\frac{p_a - p_b}{1-p_a \vee p_b}\right)^2.
\end{align*}
  
	For the $(p_a-p_b)^2$ term,  by $\exp(x)\geq 1+x$, we have
		\begin{align*}
		\frac{1}{1+\exp(-a \vee b)}-\frac{1}{1+\exp(-a \wedge b)} = \frac{e^{-a \wedge b}-e^{-a \vee b}}{(1+\exp(-a \vee b))(1+\exp(-a \wedge b))} \\
		\leq  \frac{1-e^{a \vee b-a \wedge b}}{(1+e^{a \wedge b})(1+e^{-a \vee b})} \leq (p_a \vee p_b)(1-e^{a \vee b-a \wedge b}) \leq (p_a \vee p_b) (a \wedge b-a \vee b).
		\end{align*}
	\end{proof}

{	\begin{lemma}[Lower bound of the $1/2$ divergence] \label{lem:lower of divergence}Let $p_a=1/(1+\exp(-a))$ and $p_b=1/(1+\exp (-b))$. Define $P_a$ and $P_b$ as the Bernoulli measures with probability $p_a$ and $p_b$.
 \begin{enumerate}
     \item  Suppose that there exist constants $c,C>0$ such that $c<a,b<C$, then we have
		\begin{equation*}
		D_{\frac{1}{2}}(P_a,P_b) \gtrsim  (b-a)^2.
		\end{equation*}
\item   Suppose that $a,b \rightarrow -\infty$ such that $p_a,p_b \rightarrow 0$, then we have
		\begin{equation*}
		D_{\frac{1}{2}}(P_a,P_b) \gtrsim  \exp \left\{a \wedge b \right\}(b-a)^2.
		\end{equation*}
  \end{enumerate}
  	\end{lemma}
   }

	\begin{proof}
 		\begin{align*}
		D_{\frac{1}{2}}(p_a,p_b)=-2\log(1-h^2(p_a,p_b)) \geq 2h^2(p_a,p_b) =  \left[(\sqrt{p_a}-\sqrt{p_b})^2+(\sqrt{1-p_a}-\sqrt{1-p_b})^2\right].
		\end{align*}
	For the first conclusion,  since $a,b$ are bounded, $p_a,p_b$ are bounded away form $0$ and $1$, and $(\sqrt{p_a}+\sqrt{p_b}),(\sqrt{1-p_a}+\sqrt{1-p_b})$ are bounded from $0$ as well. Hence,
		\begin{align*}
		D_{\frac{1}{2}}(p_a,p_b) \gtrsim \left[(\sqrt{p_a}-\sqrt{p_b})^2(\sqrt{p_a}+\sqrt{p_b})^2+(\sqrt{1-p_a}-\sqrt{1-p_b})^2(\sqrt{1-p_a}+\sqrt{1-p_b})^2\right] \\
		\gtrsim (p_a-p_b)^2 \stackrel{(i)}{=} \left\{\frac{\exp(x)}{(1+\exp(x))^2} \right\}^2(a-b)^2 \gtrsim (a-b)^2.
		\end{align*}
		where $(i)$ is because the mean value theorem and $a<x<b$ is bounded.

  For the second conclusion, when the probabilities $p_a,p_b$ are converging to zeros, for the term $(\sqrt{p_a}-\sqrt{p_b})^2$, by the mean value theorem of function $\sqrt{p_x}$ with respect to $x$, we have
  \begin{equation*}
     (\sqrt{p_a}-\sqrt{p_b})^2 \geq  \left( \frac{\sqrt{\exp(x)} \sqrt{1+\exp(x)}}{2 (1+\exp(x))^2} \right)^2 (a-b)^2 \stackrel{(i)}{\gtrsim} e^{a \wedge b }(a-b)^2,
  \end{equation*}
  where $(i)$ is because $\exp(x)$ is the order of $\exp \{a \wedge b \}$ for $a\wedge b<x<a \vee b$. For the term $(\sqrt{1-p_a}-\sqrt{1-p_b})^2$, note that $\sqrt{1-p_a}+\sqrt{1-p_b}$ is still bound away from $0$, we have 
  \begin{equation*}
      (\sqrt{1-p_a}-\sqrt{1-p_b})^2 \gtrsim (p_a-p_b)^2 = \left\{\frac{\exp(x)}{(1+\exp(x))^2} \right\}^2(a-b)^2 \gtrsim e^{(2a) \wedge (2b) } (a-b)^2,
  \end{equation*}
  for $a<x<b$. Finally, $\exp(a \wedge b )(a-b)^2$ dominates when the sum of the two lower bounds is taken into account.
  
	\end{proof}

	\begin{lemma}[Probability bound for maximal of sub-Gaussian random variables] \label{lem:sub_gaussian} Let $X_1,...,X_n$ be independent sub-Gaussian random variables with mean zero and sub-Gaussian norm upper bounded by $\sigma$. Then we have for every $t>0$,
		$$ P\left\{\max_{i=1,...,n} |X_i| \geq \sqrt{2\sigma^2(\log n +t)} \right\}\leq 2e^{-t}.$$
	\end{lemma}
	\begin{proof}
		By union bound and the sub-Gaussianity, we have
		\begin{equation*}
		P\left\{\max_{i=1,...,n} |X_i |\geq u\right\}  \leq \sum_{i=1}^n P\{|X_i|\geq u\} \leq 2ne^{-\frac{u^2}{2\sigma^2}},
		\end{equation*}
		by choosing $u=\sqrt{2 \sigma^2(\log n +t)}$, the conclusion is proved.
	\end{proof}

\subsection{Additional Simulation Examples}\label{sec:additional_simu}

\textbf{Gaussian Networks:} 	25 replicated data sets are generated from
	$Y_{ijt} \sim \mathcal{N}(0.1+\+x_{it}'\+x_{jt},0.1^2)$ for $i \ne j=1,...,n$ and $t=1,...,T$ where $n=100, \,T=100, \,d=2$. Let $\+x_{i1} \sim \mathcal{N}((0,0)',\tau^2 \indm)$, and transitions $\+x_{it} = \+x_{i(t-1)} + \+\epsilon_{t-1}$, where given any coordinate $j$ for a fixed node $i$, let $[\epsilon_{ij1},...,\epsilon_{ijT}]' \sim \mathcal{N}(\+0,\tau^2 \indm)$.  The iterations are stopped when the difference between predictive RMSEs in two consecutive cycles is less than $10^{-3}$.  In the algorithm, both the initial and transition variances are learned adaptively with prior~\eqref{eq:prior_sd}. The prior for the intercept is set to {$\mathcal{N}(0,10)$}. Table~\ref{tab:1} shows the mean of the $25$ replicated simulations. Clearly, SMF performs much better than MF in parameter recovery when the transition is small. The result again reinforces that when the dependence among latent positions is significant, SMF should be adopted.

 \textbf{ {Reused simulation case:}}
 For the many simulation cases in this subsection, we use 25 replicated data sets generated from the following case:
	$Y_{ijt} \sim \mathcal{N}(\+x_{it}'\+x_{jt},0.1^2)$ for $i \ne j=1,...,n$ and $t=1,...,T$ where $n=20, \,T=20, \,d=2$. Let $\+x_{i1} \sim \mathcal{N}((0,0)',\tau^2 \indm)$, and transitions $\+x_{it} = \+x_{i(t-1)} + \+\epsilon_{t-1}$, where given any coordinate $j$ for a fixed node $i$, let $[\epsilon_{ij1},...,\epsilon_{ijT}]' \sim \mathcal{N}(\+0,\tau^2 \indm)$. The prior is set in the same way as the above.

	\begin{table}
		
		\centering
		\begin{tabular}{r|lllllllllllll}
			\hline
			$\tau$  &0.001 & 0.005 & 0.01 & 0.05 & 0.1 & 0.5 \\
			\hline
			
			{MF}           & 0.0101 & 0.0105 & 0.0129 & 0.0219 & 0.0205 & 0.0211 \\

			{SMF}            & 2.34 $\times 10^{-3}$ & 5.97 $\times 10^{-3}$ & 9.53 $\times 10^{-3}$ & 0.0200 & 0.0206 & 0.0210 \\
			\hline
		\end{tabular}
  \caption{Performance comparison for Gaussian networks between SMF and MF. The measure is the median of root mean square error for estimation of the latent distances of the repeated simulations.}\label{tab:1}
	\end{table}

\textbf{Recovery of inner products:}
{We consider a simulation case regarding visualization of the recovery of $\+x_{it}'\+x_{jt}$, by showing the comparison of estimated $\hat{\+x}_{it}'\hat{\+x}_{jt}$   vs the truth $\+x^{*'}_{it}\+x^{*}_{jt}$. The data is generated for one realization of the `Reused simulation case' with $\beta^*=0$ and the two following scenarios: 1. Using our algorithm with estimating $\hat{\beta}$ as unknown beta case;
 2. Specify $\hat{\beta}=0$ as known beta case. We then compare the true inner products $\+x_{it}^{*'}\+x_{jt}^*$ and the estimated inner products $\hat{\+x}_{it}'\hat{\+x}_{jt}$ without adding the estimated intercept for both cases. The simulation is provided in Figure~\ref{fig:true_vs_est_inner}: 
\begin{figure}
    \centering
    \includegraphics[width=0.9\linewidth]{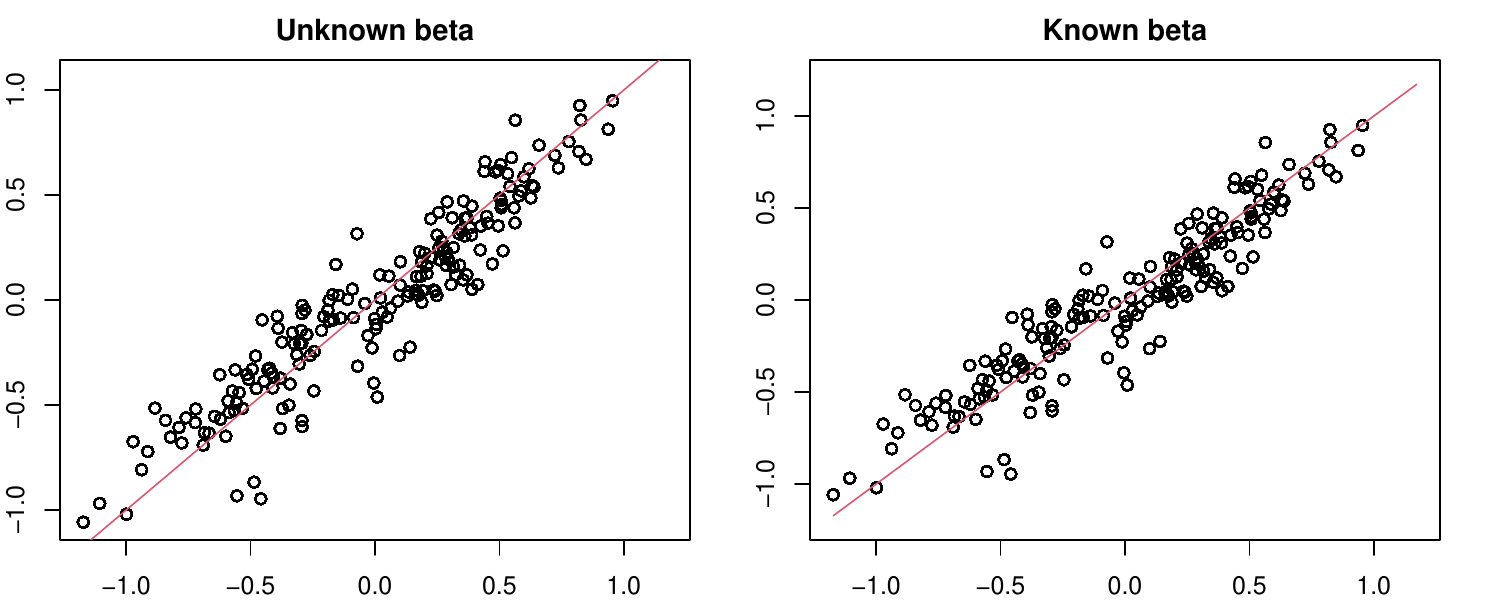}
    \caption{ Comparison of recovery of the inner product $\mathbf{x}_{i t}^{*\prime} \mathbf{x}_{i t}^*$ for  estimating $\hat{\beta}$ using the algorithm in the left vs correctly specifying $\hat{\beta}=0$ in the right. The algorithm estimates $\hat{\beta} = 0.0148$ when $\beta$ is assumed unknown. The x-axis is the true inner products $\+x_{it}^{*'}\+x_{jt}^*$ and the y-axis is the estimated inner products $\hat{\+x}_{it}'\hat{\+x}_{jt}$.}
    \label{fig:true_vs_est_inner}
\end{figure}
In the figure above, it is evident that the estimation of inner products remains accurate even when the value of $\beta$ is estimated in the algorithm instead of being correctly specified at $\beta^*=0$. This is due to the good estimate of $\hat{\beta}=0.0148$. The theoretical explanations behind this intriguing phenomenon are reserved for future research.}

\begin{figure}
	\centering
		\includegraphics[width=0.8\linewidth]{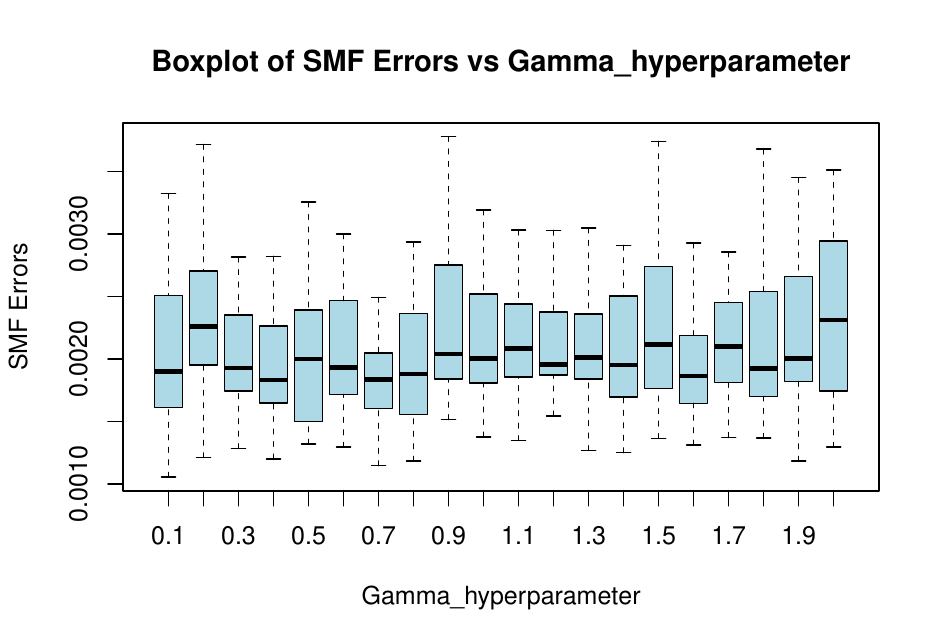}
		\caption{ {Performance comparison for Gaussian networks of estimation of SMF and MF across different hyperparameters of $d_\tau$ for Gamma distribution with $c_{\tau} = 1$.  }}
		\label{fig:Esti_vs_IG_hyper}
	\end{figure}

\begin{figure}
	\centering
		\includegraphics[width=0.8\linewidth]{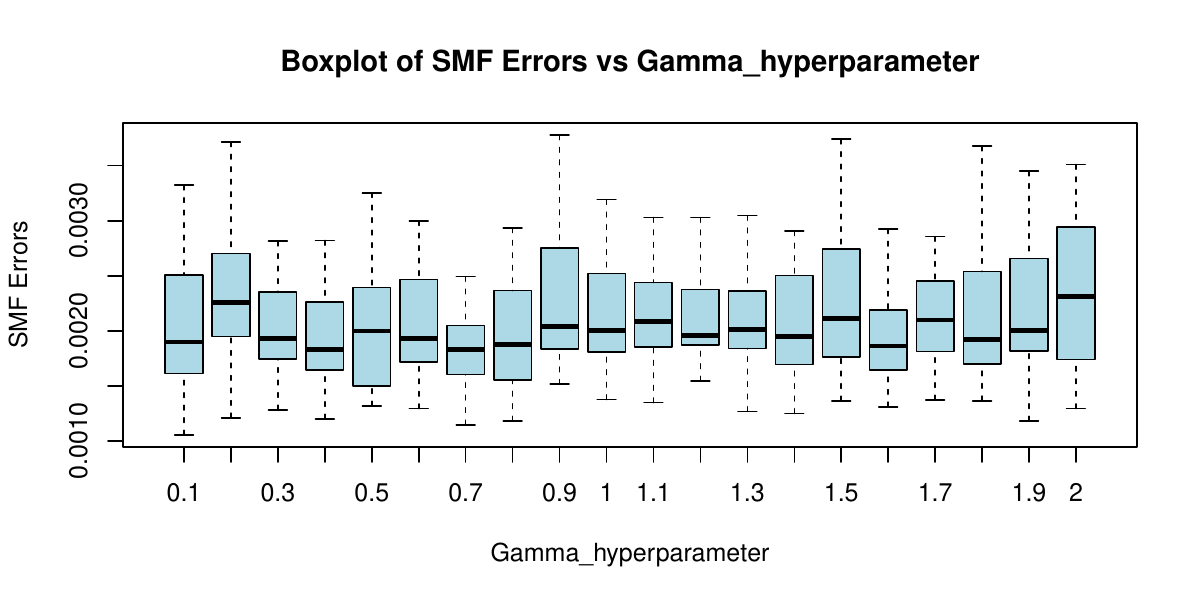}
		\caption{ {Performance comparison for Gaussian networks of estimation of SMF and MF across different hyperparameters of $c_\tau$ for Gamma distribution with $d_{\tau} = 1/2$. }}
		\label{fig:Esti_vs_IG_hyper_2}
	\end{figure}

\textbf{Sensitivity analysis with respect to the choice of Hyperparameters:}
For the Gamma prior, we test the sensitivity of the hyperparameters $c_\tau$ and $d_\tau$.  We fix $c_\tau =1$ and change $d_\tau$ from $0.1$ to $2$ with 20 grids and repeat 25 simulations of the {`Reused simulation case'} and the result is shown in Figure~\ref{fig:Esti_vs_IG_hyper}.
On the other hand, we also fix $d_\tau =1/2$ and change $c_\tau$ from $0.1$ to $2$ with 20 grids and repeat 25 simulations of the {`Reused simulation case'}, and the result is shown in Figure~\ref{fig:Esti_vs_IG_hyper_2}.

\begin{figure}[H]
	\centering
		\includegraphics[width=0.6\linewidth]{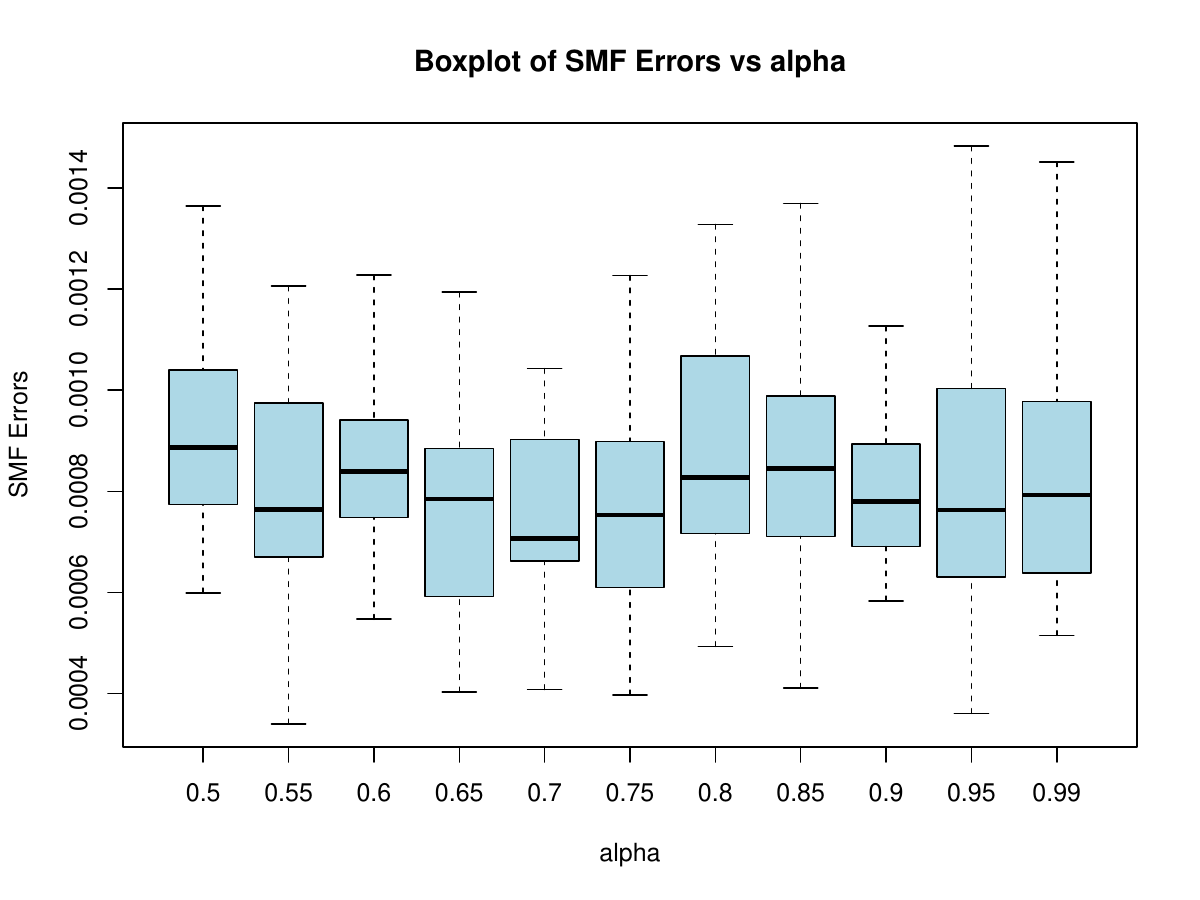}
		\caption{ {Performance comparison for Gaussian networks of estimation of SMF and MF across different $\alpha$. }}
		\label{fig:error_vs_alpha}
	\end{figure}

\textbf{Sensitivity analysis with respect to the choice of $\alpha$:}
We compared the effect of different $\alpha$ values in our model. {We use $11$ grids for $\alpha$\\=0.5, \, 0.55, \, 0.6, \, 0.65, \, 0.7, \, 0.75, \, 0.8, \, 0.85,\, 0.9,\, 0.95,\,0.99.} We repeat the simulation of the {`Reused simulation case'} for 25 times. The simulation result is shown in Figure~\ref{fig:error_vs_alpha}.  As can be seen from the figure, the results are consistent and suggest that the choice of $\alpha$ does not affect the outcome.

\begin{figure}[ht]
    \centering
    \includegraphics[width=0.8\textwidth]{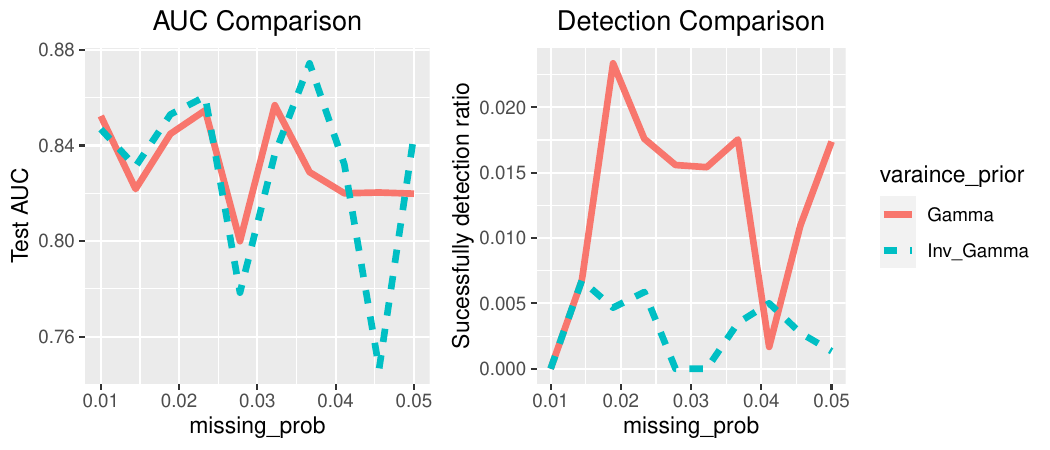}
     \includegraphics[width=0.8\textwidth]{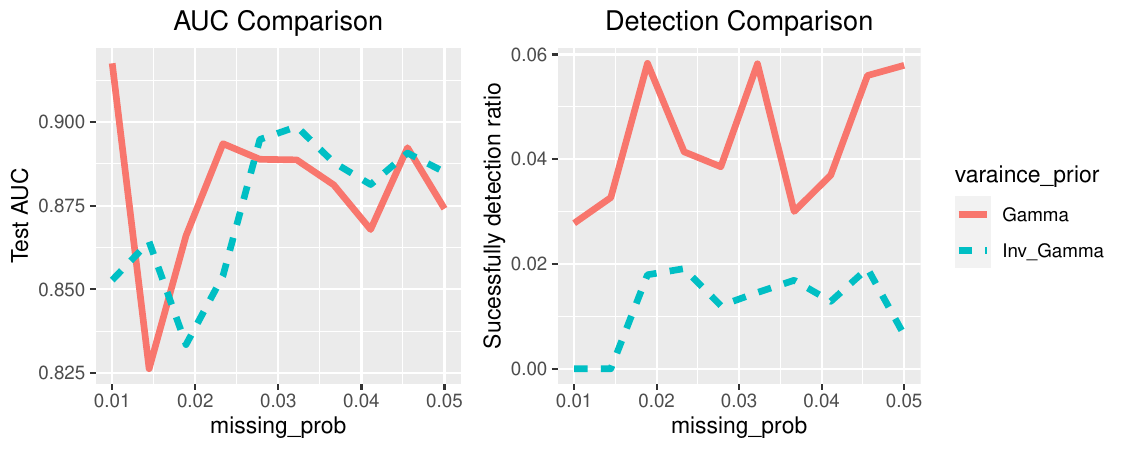}
      \includegraphics[width=0.8\textwidth]{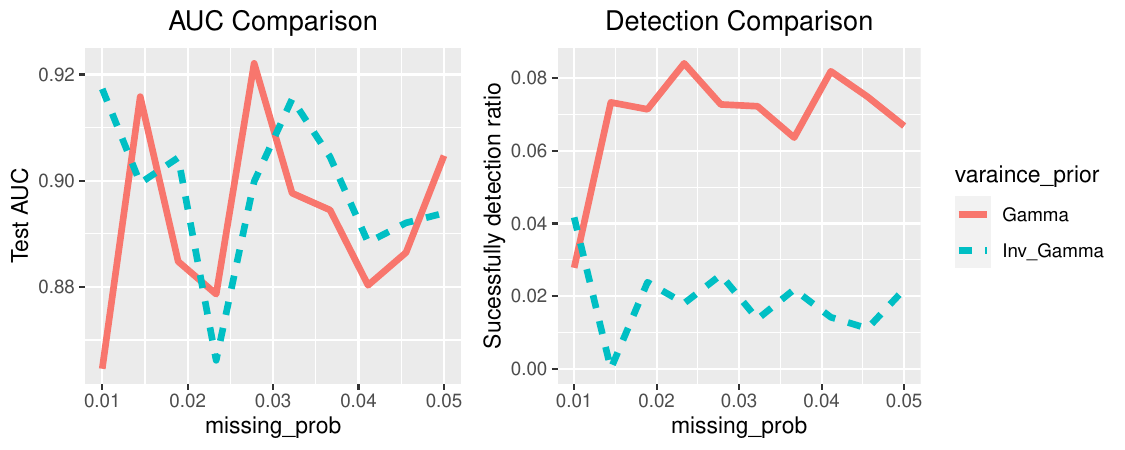}
    \caption{Comparison using Enron's email data set between Gamma prior and Inverse Gamma prior on the transition variance, with cases for $d=2, 3, 4$ displayed from top to bottom.}
    \label{fig:d234}
\end{figure}

\textbf{Sensitivity analysis with respect to the choice of latent dimensions for Enron's email data:}
In the Enron email data set, we found that the $d=5$ case provides a good comparison between our method and using an inverse Gamma prior. Our method consistently performed better than using the inverse Gamma prior. Here we have also shown the comparison results for $d=2,3,4$ in Figure~\ref{fig:d234}. We observed that $d=2,3,4,5$ showed similar behavior, where using the Gamma prior consistently outperformed the inverse Gamma prior for the transition variance. 

\begin{figure}
	\centering
		\includegraphics[width=0.8\linewidth]{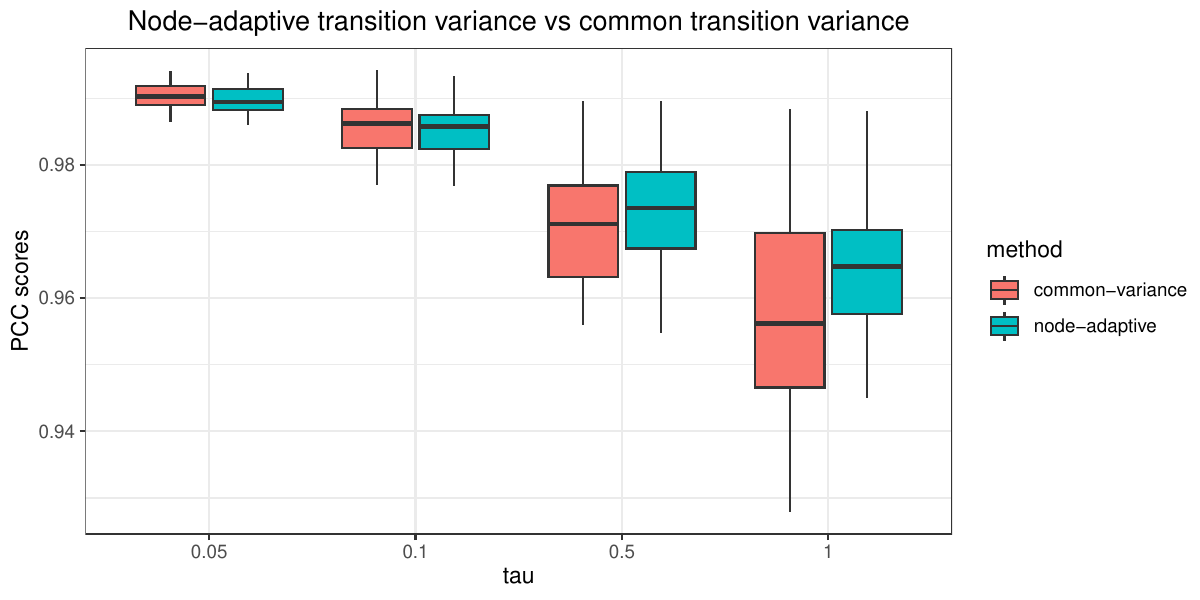}
		\caption{ Performance comparison for nodewise adaptive prior on variance vs. common transition variance across different $\tau$ as the transition standard derivation $10\%$ of changing nodes, while the rest $90\%$ of nodes stay static across all time points.}
		\label{fig:nodewise}
	\end{figure}

 \textbf{Nodewise adaptive priors:}
Our new simulations demonstrate that node adaptivity can improve estimation accuracy when different nodes have different transition scales. We consider a scenario where $90\%$ nodes remain static across all time (so that their corresponding $L_i$ can be seen as $0$), while only the rest $10\%$ of nodes change over time. For the changing node, we use $\tau$ as the true transition standard derivation to control the magnitude of changes: $\+x_{it} \mid \+x_{i(t-1)} \sim \m N(\+x_{i(t-1)},\tau^2 \+I_d) $. The other settings are similar: we use 25 replicated data sets are generated from
	$Y_{ijt} \sim \mbox{Bernoulli}(\+x_{it}'\+x_{jt})$ for $i \ne j=1,...,n$ and $t=1,...,T$ where $n=20, \,T=50, \,d=2$. Let $\+x_{i1} \sim \mathcal{N}((0,0)', 0.1\indm)$,  where given any coordinate $j$ for a fixed node $i$.  We compare the estimation accuracy between the nodewise adaptive priors~\eqref{eq:prior_nodewise} and common variance priors~\eqref{eq:prior_sd} across different values of $\tau$ such as $\tau=0.05,0.1,0.5,1$. Figure~\ref{fig:nodewise} illustrates the performance between nodewise adaptive priors and common priors. When $\tau=0.05,0.1$ is small, the common variance prior performs most the same or slightly better than nodewise adaptive priors because differences in the magnitude of change for different nodes may not be large enough and both methods may converge at the same rate. However, when $\tau=0.5,1$ are not close to zero, nodewise adaptive priors perform much better than the common variance prior. This is reasonable as the differences in the change of scale between different nodes are large in these cases and introducing nodewise adaptivity can capture the true data-generating process more precisely.

	\subsection{MF Updatings for \texorpdfstring{$\beta$}{beta}}
	Suppose the prior for $\beta$ is $\mathcal{N}(\mu_\beta,\sigma_\beta^2)$. For Gaussian likelihood, the updating for $\beta$ can be obtained       
	\begin{align*}
	\hat{q}(\beta) &\propto \exp [\E_{-\beta}\{\log p_\alpha (\m X,\beta,\m Y )\}] \propto \exp [\E_{-\beta}\{\log P_\alpha(\m Y \mid \m X,\beta )\}+ \log p(\beta)] \\
	& \propto \exp \left[\E_{-\beta}\left\{ \alpha \sum_{t=1}^T\sum_{i \neq j}  -\frac{(Y_{ijt}-\beta- \+x_{it}'\+x_{jt})^2}{2\sigma^2}  \right\}  -\frac{(\beta-\mu_{\beta})^2}{2\sigma_\beta^2} \right] \\
	& \propto \exp \left[ \sum_{t=1}^T\sum_{i \neq j}  -\alpha \frac{\beta^2-2\beta (Y_{ijt}-\+\mu_{it}'\+\mu_{jt}) }{2\sigma^2}    -\frac{(\beta-\mu_{\beta})^2}{2\sigma_\beta^2} \right].
	\end{align*}  
	Therefore, $q^{(\mbox{new})}(\beta)$ is the density of $\m N(\mu_{\beta}^{(new)},\sigma_{\beta}^{(new)})$, with
	\begin{align*}
	&\sigma_{\beta}^{(new)2} = \left\{\sigma_{\beta}^{-2}+ \alpha Tn(n-1)\sigma^{-2}\right\}^{-1}, & \mu_{\beta}^{(new)}=\sigma_{\beta}^{(new)^2}\left\{\sigma_{\beta}^{-2} \mu_{\beta} +\sum_{i \neq j}\sum_{t} \alpha \sigma^{-2} (Y_{ijt}-\+\mu_{it}'\+\mu_{jt}) \right\}.
	\end{align*}
	
	For the binary case, the updating for $\beta$ after tangent transformation can also be obtained    
	\begin{align*}
	q^{(\mbox{new})}(\beta; \Xi) \propto \exp [\E_{\m X}\{\log  \underline{P}_\alpha (\m Y \mid \m X,\+\beta;\Xi )\} + \log p(\beta)] \\
	\propto \exp \left [\sum_{i \neq j}\sum_{t} \alpha \left\{A(\xi_{ijt})\right\}\beta^2+\sum_{i \neq j}\sum_{t}\alpha  \left\{Y_{ijt}-\frac{1}{2}+2A(\xi_{ijt}) \+\mu_{it}'\+\mu_{jt}  \right\}\beta-\frac{1}{2} \sigma^{-2}_{\beta} \beta^2+ \mu_{\beta} \sigma^{-2}_{\beta} \beta \right].
	\end{align*}
	Therefore, $q^{(\mbox{new})}(\beta; \Xi)$ is the density of $\m N(\mu_{\beta}^{(new)},\sigma_{\beta}^{(new)})$, with
	\begin{align*}
	&\sigma_{\beta}^{(new)2} = \left\{\sigma_{\beta}^{-2}-2 \alpha \sum_{i \neq j}\sum_{t}  A(\xi_{ijt}) \right\}^{-1}, & \mu_{\beta}^{(new)}=\sigma_{\beta}^{(new)^2}\left [\sigma_{\beta}^{-2} \mu_{\beta} +\sum_{i \neq j}\sum_{t} \alpha\left\{Y_{ijt}-\frac{1}{2}+2  A(\xi_{ijt}) \+\mu_{it}'\+\mu_{jt} \right\}\right].
	\end{align*}

	\subsection{MF Updatings for \texorpdfstring{$\m X$}{X}}

	The updating for $\beta$ in MF is the same with SMF. For updating $\m X$ in the Gaussian case, we have
	\begin{align*}
	\hat{q}(\+x_{it}) &\propto \exp [\E_{-\+x_{it}}\{\log p_\alpha (\m X,\beta,\m Y )] \propto \exp [\E_{-\+x_{it}}\{\log P_\alpha(\m Y \mid \m X,\beta )\}+ \log p(\m X)\}] \\
	& \propto \exp \left[\E_{-\+x_{it}}\left\{ \sum_{i \neq j}  -\alpha\frac{(Y_{ijt}-\beta -\+x_{it}'\+x_{jt})^2}{2\sigma^2}    -\frac{\|\+x_{it}-\+x_{i(t-1)}\|^2}{2\tau^2} -\frac{\|\+x_{it}-\+x_{i(t+1)}\|^2}{2\tau^2} \right\}\right] \\ 
	&\propto \exp \left[\left\{ \sum_{i \neq j}  -\alpha\frac{-2 ( Y_{ijt}-\mu^{(new)}_{\beta}) \+x_{it}'\+\mu_{jt}+  \+x_{it}'(\+\mu_{jt}\+\mu_{jt}'+\+\Sigma_{jt})\+x_{it}}{2\sigma^2}   -\frac{\|\+x_{it}\|^2-2 \+x_{it}\+\mu_{i(t-1)}}{2\tau^2}\right.\right. \\
	&\left.\left. -\frac{\|\+x_{it}\|^2-2 \+x_{it}\+\mu_{i(t+1)}}{2\tau^2} \right\}\right].
	\end{align*}
	Therefore, $q^{(\mbox{new})}(\+x_{it})$ is the density of $\m N(\+\mu_{it}^{(new)},\+\Sigma_{it}^{(new)})$, with
	\begin{align*}
	\+\Sigma_{it}^{(new)} = \left\{ 2\tau^{-2}  \indm  + \alpha\sigma^{-2} \sum_{i \neq j} (\+\mu_{jt}\+\mu_{jt}'+\+\Sigma_{jt}) \right\}^{-1},\\  \+\mu_{it}^{(new)}=\+\Sigma_{it}^{(new)}\left(\tau^{-2} \+\mu_{i(t-1)}+\tau^{-2}\+\mu_{i(t+1)} +\sum_{i \neq j} \alpha\sigma^{-2} (Y_{ijt}-\mu^{(new)}_{\beta}) \+\mu_{jt} \right).
	\end{align*}

	For the binary case, here we derive the updating formula under the mean-filed approximation for $\m X$ after performing the tangent approximation. For the mean-field updating for $\+x_{it}$, we have:
	\begin{align*}
	\hat{q}(\+x_{it}) &\propto \exp [\E_{-\+x_{it}}\{\log \underline p_\alpha(\m X,\beta,\m Y )] \propto \exp [\E_{-\+x_{it}}\{\log \underline P_\alpha(\m Y \mid \m X,\beta )\}+ \log p(\m X)\}] \\
	& \propto \exp \left[\E_{-\+x_{it}}\left\{ \sum_{i \neq j} \alpha \left (A(\xi_{ijt}) (\+x_{it}'\+x_{jt})^2+(2 A(\xi_{ijt})\beta +Y_{ijt}-\frac{1}{2})\+x_{it}'\+x_{jt} \right)  \right. \right.\\
	& \left. \left.-\frac{\|\+x_{it}-\+x_{i(t-1)}\|^2}{2\tau^2} -\frac{\|\+x_{it}-\+x_{i(t+1)}\|^2}{2\tau^2} \right\}\right] \\ 
	&\propto \exp \left[\left\{ \sum_{i \neq j}\alpha \left( A(\xi_{ijt})\+x_{it}'(\+\mu_{jt}\+\mu_{jt}'+\+\Sigma_{jt})\+x_{it} +(2 A(\xi_{ijt})\mu^{(new)}_\beta +Y_{ijt}-\frac{1}{2})\+x_{it}'\+\mu_{jt} \right)  \right. \right.\\
	& \left. \left. -\frac{\|\+x_{it}\|^2-2 \+x_{it}\+\mu_{i(t-1)}}{2\tau^2} -\frac{\|\+x_{it}\|^2-2 \+x_{it}\+\mu_{i(t+1)}}{2\tau^2} \right\}\right].
	\end{align*}
	Therefore, $q^{(\mbox{new})}(\+x_{it})$ is the density of $\m N(\+\mu_{it}^{(new)},\+\Sigma_{it}^{(new)})$, with
	\begin{align*}
	\+\Sigma_{it}^{(new)} = \left\{ 2\tau^{-2}\indm - 2\alpha \sum_{i \neq j} \left(A(\xi_{ijt})(\+\mu_{jt}\+\mu_{jt}'+\+\Sigma_{jt}) \right) \right\}^{-1},\\  \+\mu_{it}^{(new)}=\+\Sigma_{it}^{(new)}\left(\tau^{-2} \+\mu_{i(t-1)}+\tau^{-2} \+\mu_{i(t+1)} +\sum_{i \neq j} \alpha(2 A(\xi_{ijt}) \mu_{\beta}^{(new)}+Y_{ijt}-\frac{1}{2}) \+\mu_{jt} \right).
	\end{align*}

\vskip 0.2in
	\bibliography{DNAMP}

\end{document}